\documentclass{article}

\usepackage{microtype}
\usepackage{graphicx}
\usepackage{subcaption}
\usepackage{booktabs} %

\usepackage[breaklinks=true]{hyperref}

\usepackage[accepted]{icml2026}

\usepackage{amsmath}
\usepackage{amssymb}
\usepackage{mathtools}
\usepackage{amsthm}

\usepackage[capitalize,noabbrev]{cleveref}

\usepackage[dvipsnames]{xcolor}  %
\usepackage[acronym]{glossaries} %
\glsdisablehyper
\newcommand{\glsf}[1]{\glsreset{#1}\gls{#1}}  %
  
\usepackage{multirow} %
\usepackage{makecell}  %
\usepackage[most]{tcolorbox}
\usepackage{siunitx}
\usepackage{tikz}
\usepackage{pgfplots}
\usetikzlibrary{patterns}
\usetikzlibrary{patterns.meta}
\usetikzlibrary{pgfplots.fillbetween}
\usepgfplotslibrary{groupplots}

\usetikzlibrary{external}
\tcbset{shield externalize}  %

\definecolor{darkgray142}{RGB}{144,144,144}
\definecolor{darkgray176}{RGB}{176,176,176}
\definecolor{darkgray208}{RGB}{208,208,208}
\definecolor{lightgray204}{RGB}{204,204,204}

\definecolor{ApproxPointCount}{RGB}{46,204,113}
\definecolor{WithNoise}{RGB}{46,204,113}

\definecolor{PointPatchFF}{HTML}{5B9BD5}
\definecolor{PointPatch}{HTML}{1F4E79}

\definecolor{PointPatchAttnFF}{HTML}{4ECDC4}
\definecolor{PointPatchAttn}{HTML}{1A6560}

\definecolor{PCMFF}{HTML}{70AD47}
\definecolor{PCM}{HTML}{385723}

\definecolor{DP3FF}{HTML}{AF7AC5}
\definecolor{DP3}{HTML}{6C3483}

\definecolor{PointTransformerFF}{HTML}{F1948A}
\definecolor{PointTransformer}{HTML}{922B21}

\definecolor{PPRGBFF}{HTML}{ED7D31}
\definecolor{PPRGB}{HTML}{A2460B}

\definecolor{PPRGBPretrainedFF}{HTML}{F4D03F}
\definecolor{PPRGBPretrained}{HTML}{B7950B}

\definecolor{PointMap}{HTML}{8B5A2B} %
\definecolor{RGBOnly}{HTML}{4A2711} %

\newacronym{pmp}{PMP}{PointMapPolicy}
\newacronym{il}{IL}{Imitation Learning}
\newacronym{bc}{BC}{Behavioral Cloning}
\newacronym{rl}{RL}{Reinforcement Learning}
\newacronym{knn}{KNN}{K-Nearest Neighors}
\newacronym{fps}{FPS}{Farthest Point Sampling}
\newacronym{ssm}{SSM}{State Space Model}
\newacronym{ode}{ODE}{Ordinary Differential Equation}
\newacronym{pde}{PDE}{Partial Differential Equation}
\newacronym{sde}{SDE}{Stochastic Differential Equation}
\newacronym{gnn}{GNN}{Graph Neural Network}
\newacronym{mlp}{MLP}{Multi-Layer Perceptron}
\newacronym{rff}{RFFs}{random Fourier features}
\newacronym{spe}{SPE}{Sinusoidal Positional Encoding}
\newacronym{nerf}{NeRF}{Neural Radiance Field}
\newacronym{fft}{FFT}{Fast Fourier Transform}
\newacronym{gft}{GFT}{Graph Fourier Transform}

\usepackage{amsmath,amsfonts,bm}

\def\eqref#1{equation~\ref{#1}}

\def\1{\bm{1}}

\DeclareMathAlphabet{\mathsfit}{\encodingdefault}{\sfdefault}{m}{sl}
\SetMathAlphabet{\mathsfit}{bold}{\encodingdefault}{\sfdefault}{bx}{n}

\newcommand{\traj}{\tau}
\newcommand{\goal}{\mathbf{g}}
\newcommand{\obs}{o}
\newcommand{\act}{a}
\newcommand{\acts}{\mathbf{a}}

\newcommand*\diff{\mathop{}\!\mathrm{d}}

\theoremstyle{plain}

\theoremstyle{definition}

\theoremstyle{remark}

\usepackage[textsize=tiny]{todonotes}

\icmltitlerunning{Fourier Features for Point Cloud Imitation Learning}

\begin{document}

\twocolumn[
  \icmltitle{Fourier Features Let Agents Learn High \\
    Precision Policies with Imitation Learning}

  \icmlsetsymbol{equal}{*}

  \begin{icmlauthorlist}
    \icmlauthor{Balázs Gyenes}{kit,h4h}
    \icmlauthor{Emiliyan Gospodinov}{kit}
    \icmlauthor{Jan Frieling}{kit}
    \icmlauthor{Enrico Krohmer}{kit}
    \icmlauthor{Nicolas Schreiber}{kit}
    \icmlauthor{Xiaogang Jia}{kit}
    \icmlauthor{Niklas Freymuth}{kit}
    \icmlauthor{Gerhard Neumann}{kit}
  \end{icmlauthorlist}

  \icmlaffiliation{kit}{Autonomous Learning Robots, Karlsruhe Institute of Technology, Germany}
  \icmlaffiliation{h4h}{HIDSS4Health - Helmholtz Information and Data Science School for Health, Karlsruhe/Heidelberg, Germany}

  \icmlcorrespondingauthor{Balázs Gyenes}{balazs@gyenes.ca}
  \icmlcorrespondingauthor{Gerhard Neumann}{gerhard.neumann@kit.edu}

  \icmlkeywords{imitation learning, robotics, point clouds, fourier, precision, high-precision}

  \vskip 0.3in
]

\printAffiliationsAndNotice{}  %

\begin{abstract}

High-precision robotic manipulation requires fine-grained spatial reasoning that is often difficult to achieve with RGB-only policies due to depth ambiguity and perspective scale issues.
Policies that leverage 3D information directly, such as those based on point clouds, offer a stronger geometric prior over purely image-based ones, yet their performance remains highly task-dependent.
We hypothesize that this discrepancy may be due to the spectral bias of neural networks towards learning low frequency functions, which especially affects architectures conditioned on slow-moving Cartesian features.
We thus propose to map point clouds from Cartesian space into high-dimensional Fourier space, effectively equipping the point cloud encoder with direct access to high-frequency features. 
We experimentally validate the use of Fourier features on challenging manipulation tasks from the RoboCasa and ManiSkill3 benchmarks and on a real robot setup.
Despite their simplicity, we find that Fourier features provide significant benefits across diverse encoder architectures and benchmarks and are robust across hyperparameters.
Our results indicate that Fourier features let policies leverage geometric details more effectively than Cartesian features, showing their potential as a general-purpose tool for point cloud-based imitation learning.
We provide source code and videos on our project page: \href{https://fourier-il.github.io/fourier-il/}{https://fourier-il.github.io/fourier-il}.

\end{abstract}

\section{Introduction}

Diffusion-based \gls{il} has emerged as a powerful framework for robotic visuomotor control~\citep{chi2023diffusionpolicy, reuss2023goal, wu2025diffusing, pi_0.5}. 
By treating action generation as a denoising process~\citep{ho2020denoising}, diffusion policies naturally capture the multi-modal action distributions of human expert demonstrations. %
This capability has made diffusion policies the state-of-the-art on long-horizon and multi-task manipulation benchmarks.
Yet, the success of diffusion policies depends on the observation encoder's ability to extract subtle positional cues from the observed scene that inform the policy's next action.
Policies that cannot respond to the geometric information hidden in observations are unable to imitate expert demonstrations that condition on this information.

RGB images remain the most common type of observation due to their semantic richness and the widespread availability of pretrained vision encoders. 
However, because they lack an explicit representation of 3D geometry, they require the policy to implicitly infer a 2D-to-3D mapping and are also sensitive to viewpoint and image artifacts, such as lighting variations~\citep{ke20243d, adapt3r, donat2025towards}.
In contrast, 3D modalities such as depth maps, point clouds, and point maps directly represent shape, distance, and spatial relationships, helping the policy reason about geometry and occlusions and execute complex motions accurately.
While this explicit structuring provides a strong geometric prior, the success of policies that rely purely on 3D modalities varies greatly between tasks.
As a result, a number of hybrid 2D/3D architectures have been proposed~\citep{ke20243d, adapt3r, goyal2023rvt}, which often use foundation models to extract features from the 2D RGB stream and combine this with 3D information in a variety of ways.

In this paper, we argue that the key shortcoming of current 3D modalities lies in the \textit{spectral bias} of the architectures that are used to process them.
For high-precision tasks, such as inserting a peg into a socket, the policy needs to learn a sharp decision boundary for e.g. whether to insert the peg or reposition it.
Since the observations that should result in these respective actions differ only slightly, a high frequency function is best suited to represent this desired policy faithfully.
Although neural networks are universal function approximators~\cite{HORNIK1989359}, \glspl{mlp} and fully-connected layers have a spectral bias towards learning low-frequency components first, while high-frequency components converge slowly or may not be learned at all~\cite{pmlr-v97-rahaman19a, fourier_features}.
Although this phenomenon is well-known, these \glspl{mlp} underpin the majority of point cloud architectures, which use them to encode Cartesian coordinates into latent features.
In contrast, the convolutional layers underpinning most image-based architectures are inherently more biased towards high-frequency signals, with evidence that they may be lacking sensitivity to lower frequencies~\cite{abello2021dissecting, jo2017measuring, wang2020high}.

\begin{figure}[t!]
    \centering
    \includegraphics[width=\linewidth]{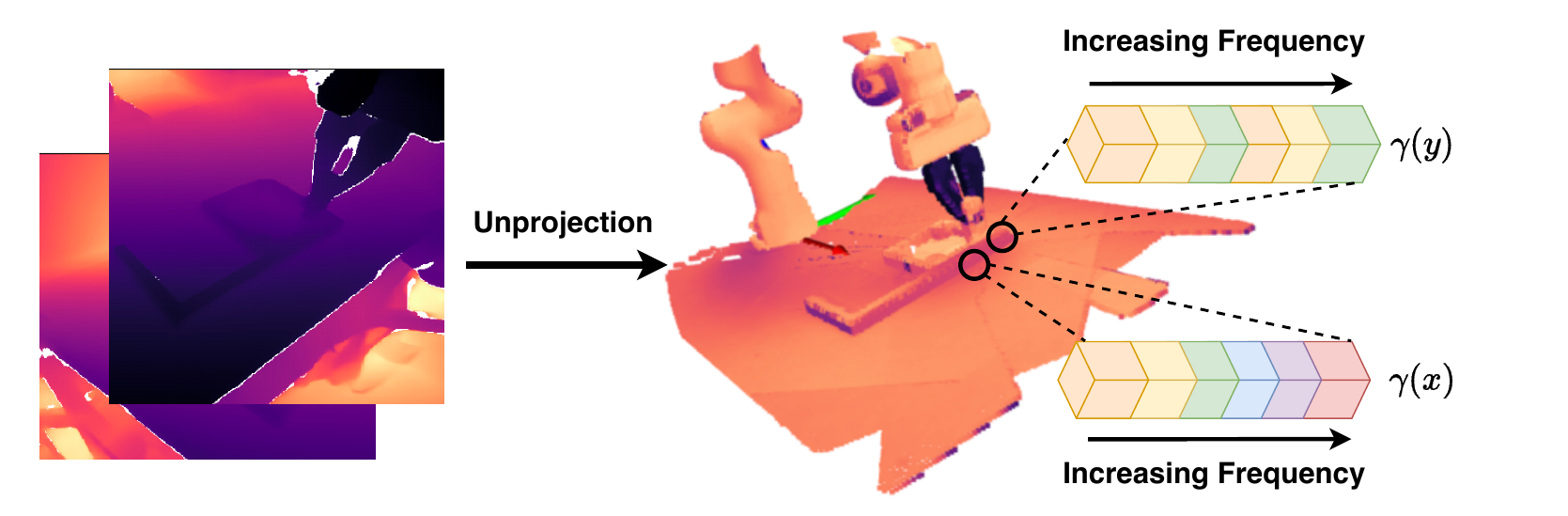}
    \caption{
        \textbf{Method Overview.}
        Adding a Fourier feature mapping from Cartesian coordinates into a higher-dimensional feature space improves performance for any point cloud encoder used for diffusion imitation learning.
        For high-precision policies, the network must learn to condition on fine details in the scene geometry e.g. to decide whether to insert the leg into the slot or reposition it, yet neural networks learn the high frequency components of the target function only slowly, if at all.
        While neighbouring points in the scene have very similar Cartesian features, the high-dimensional Fourier features allow them to easily be distinguished.
    }
    \label{fig:fourier_features}
    \vspace{-0.2cm}
\end{figure}

In fields such as novel view synthesis, the spectral bias of \glspl{mlp} is remedied using a Fourier feature mapping of the input coordinates~\citep{mildenhall2021nerf, fourier_features}.
However, recent foundation models for robotic control from point clouds~\cite{sugar, zhou2024uni3d, lift3d, pointvla} do not leverage this insight, and its use within \gls{il} architectures has so far been limited to isolated cases~\citep{adapt3r}.
This work therefore systematically evaluates Fourier feature mappings and their use for diffusion-based \gls{il} on point clouds across a broad range of policy architectures.
By projecting points into a higher dimensional space where subtle geometric differences are amplified, we counteract spectral bias.
Experimentally, we show that using Fourier-encoded input representations leads to consistent improvements across different point cloud architectures and benchmarks. 
These input representations improve success rate by up to $20\%$ and $7\%$ on RoboCasa~\citep{robocasa2024} and ManiSkill3~\citep{tao2025maniskill}, respectively.
Further, they improve normalized score on $4$ challenging real world tasks from $14.8\%$ to $40.2\%$.
Qualitatively, policies trained with Fourier mappings exhibit smoother and more precise motions, particularly on robotic control tasks where fine-grained manipulation matters.

Our contributions are as follows: 
\textbf{1)} we incorporating Fourier feature mappings into various point cloud encoders and show how this resolves their spectral bias; 
\textbf{2)} through experiments on a real robot and on the RoboCasa and ManiSkill task suites, we demonstrate consistent and robust improvements over baselines without Fourier feature mappings; and
\textbf{3)} through extensive analysis and parameter studies, we provide a number of insights.
Fourier features are most beneficial when point clouds contain rich geometric detail, but provide a boost even in the absence of fine geometry, perhaps by improving learning dynamics.
Furthermore, Fourier features do not require additional regularization and are robust to choice of hyperparameters.

\section{Related Work}

\textbf{Imitation Learning in Robotics.} 
Recent progress in \gls{il} has been driven by incorporating diffusion \citep{chi2023diffusionpolicy, reuss2023goal} and flow matching~\citep{lipman2023flow} objectives, which enable policies to learn multi-modal action distributions, and by training policies on large-scale datasets \citep{black2024pi_0, pi_0.5, brohan2022rt, zitkovich2023rt, zhu2025scaling} to significantly increase generalization and performance.
However, these approaches are primarily conditioned on RGB images. 
This choice allows leveraging powerful pretrained visual encoders and provides strong semantic features, but lacks explicit representation of 3D geometry, causing issues with depth and scale ambiguity.
Similarly, image-based representations are sensitive to viewpoint and lighting variations \citep{Ze2024DP3, zhu2024point, adapt3r}. 

\textbf{3D Visual Representations for Imitation Learning.}
To address these shortcomings, 3D information can be leveraged in stand-alone modalities, such as point clouds or point maps, or in combination with RGB.
On a number of challenging tasks, lightweight point cloud-based policies outperform RGB and RGB-D modalities while requiring significantly less data~\cite{Ze2024DP3, zhu2024point, idp3}.
Point maps~\cite{dust3r}, an alternative to point clouds defined on a grid-like structure, have been proposed for \gls{il}~\cite{pointmappolicy}, but their performance can be inconsistent~\cite{zhu2024point}.
In contrast, a variety of hybrid 2D/3D approaches augment 2D features from pre-trained image encoders with 3D position information reconstructed from the original depth maps, either at the patch level~\citep{ke20243d}, the point level~\citep{adapt3r}, or in voxels~\citep{gervet2023act3d}. 
This combines the benefits of large-scale image datasets with those of explicit geometric representations, emphasizing the benefits of complex architectural design. %
We argue that the recent popularity of such approaches is ultimately due to the spectral frequency bias of most point cloud encoders, which prevents them from accurately learning high-precision tasks when employed in isolation.
We show that simple architectures become significantly more effective when equipped with non-parametric Fourier input mappings.

\textbf{Deep Learning on Point Clouds}
Since the advent of PointNet~\cite{qi2017pointnet}, many architectures have been proposed for processing unordered sets of points~\cite{qi2017pointnetplusplus, point_transformer, stratified_transformer, qian2022pointnext}.
A common paradigm inspired by vision transformers is to group the point clouds into patches which are then tokenized using a lightweight PointNet-style network~\cite{pang2022masked, pointBERT, chen2023pointgpt}.
Patches are computed by taking the \gls{knn} around each patch center point, which are sampled randomly using \gls{fps}.
To reduce the dimensionality of the embedding, the patch tokens can optionally be aggregated using max pooling~\cite{zhu2024point} or attention~\cite{gyenes2024pointpatchrl}.
The flexibility of this paradigm has allowed it to be used to train multi-modal foundation models on point clouds~\cite{sugar, zhou2024uni3d, lift3d, pointvla}, but these models all operate on slow-changing Cartesian features.
Because of the ubiquity of PointPatch architectures in the literature, we focus our experiments on several of its variants.

\textbf{Deep Learning with Fourier Features.}  
Neural networks have a spectral bias, i.e., a tendency to learn low-frequency components faster than high-frequency ones~\citep{pmlr-v97-rahaman19a}.
As confirmed in more recent empirical work~\citep{lippe2023pde, wurth2026diffusion}, this is partly a characteristic of the mean-squared error (MSE) loss, which tends to focus on low-frequency signals in the data.
Fourier features \citep{mildenhall2021nerf, fourier_features} mitigate this issue by projecting low-dimensional Cartesian coordinates into high-dimensional sinusoidal embeddings with multiple frequencies.
This technique was key for enabling \glspl{nerf} to learn detailed 3D scenes with high fidelity and avoid blurry, oversmoothed reconstructions~\citep{mildenhall2021nerf, fourier_features, barron2022mip}.
The frequencies can be fixed or learned end-to-end~\citep{gao2023adaptive, sun2024learning}.
Adapt3R~\citep{adapt3r} leverages Fourier features to outperform other architectures on unseen viewpoints during inference, but they do not investigate their effect in other contexts.
In contrast, we apply Fourier mappings systematically across point cloud architectures in diffusion-based \gls{il}, comparing them from a frequency-domain perspective.

\section{Method}

\begin{figure}[ht]
    \centering
    \includegraphics[width=\linewidth]{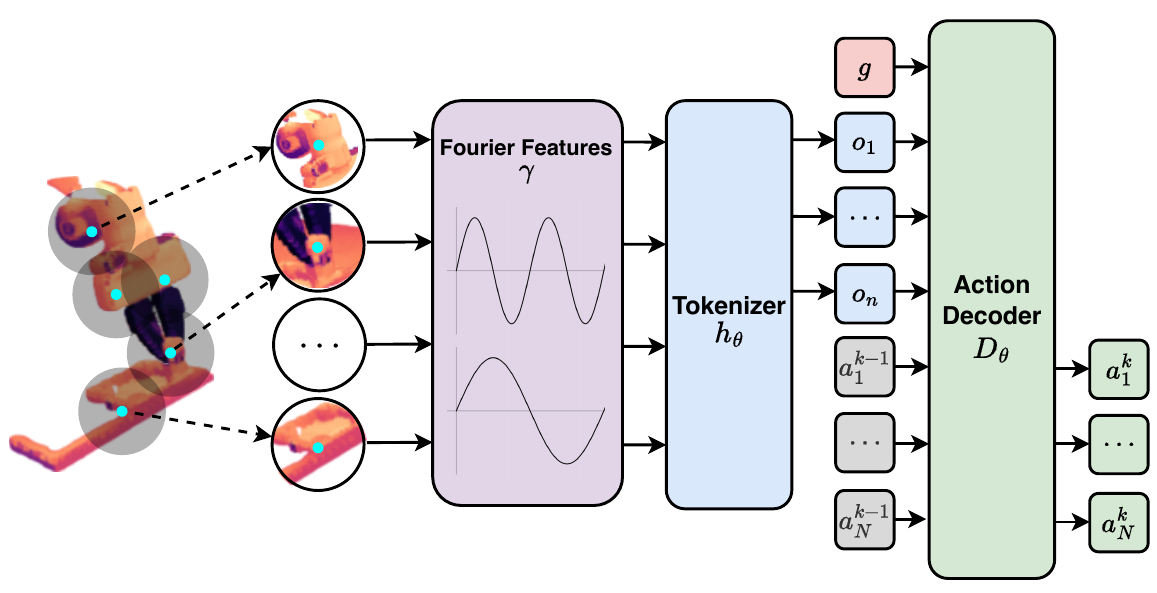}
    \caption{
        \textbf{Overview of PointPatch encoder family.}
        We group the input point cloud into neighborhoods $\mathcal{N}(i)$ (patch centers indicated in blue on the left).
        We map point coordinates into Fourier feature space to amplify subtle geometric differences between similar observations.
        The tokenizer extracts and aggregates features for each neighborhood to produce a set of tokens which are then forwarded to a goal-conditioned diffusion policy $D_\theta$ to denoise the next action chunk.
    }
    \label{fig:architecture}
    \vspace{-0.2cm}
\end{figure}

\begin{figure*}[htb!]

    \centering
    \tcbset{
        col_style/.style={
            enhanced,
            colback=gray!8,      %
            colframe=gray!20,     %
            arc=6pt,              %
            boxrule=0.5pt,        %
            left=2pt, right=2pt, top=4pt, bottom=4pt, %
            nobeforeafter,        %
            valign=top
        }
    }

    \begin{tcolorbox}[col_style, width=0.32\linewidth]
        \centering
        \textbf{RoboCasa}\\[0.5em]
        \begin{minipage}{0.48\linewidth}
            \includegraphics[width=\linewidth, height=0.95\linewidth, keepaspectratio=false, clip]{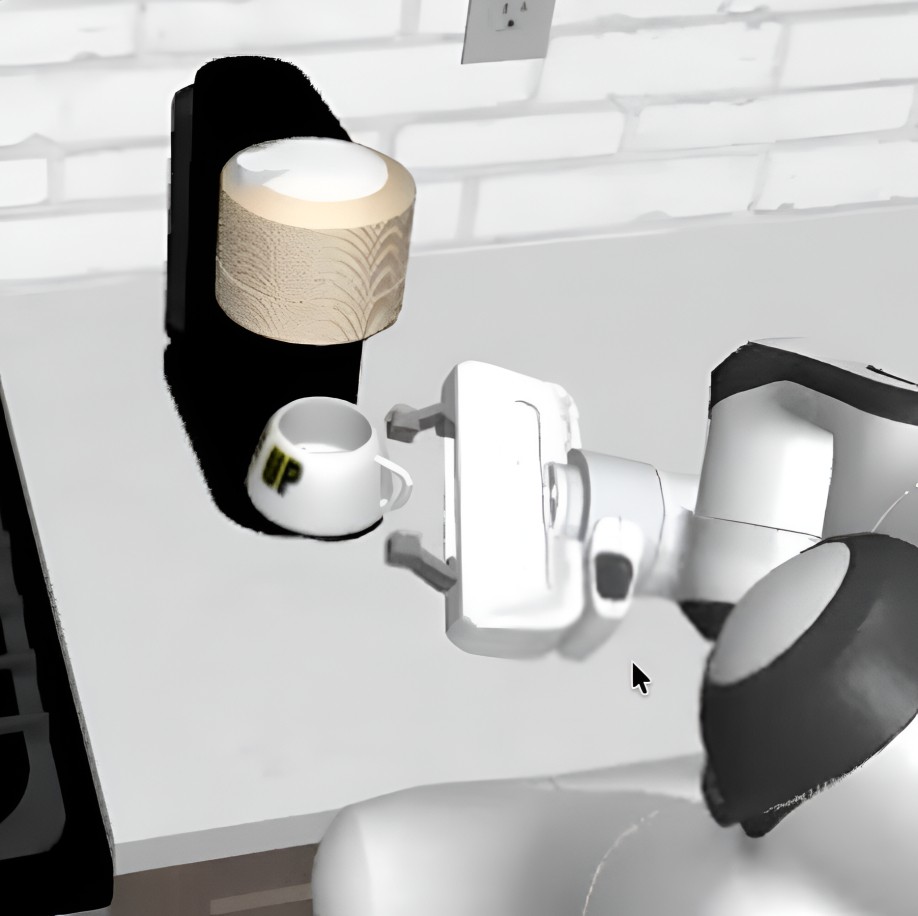}
            \scriptsize \centering \mbox{\strut CoffeeServeMug}
        \end{minipage}
        \hfill
        \begin{minipage}{0.48\linewidth}
            \includegraphics[width=\linewidth, height=0.95\linewidth, keepaspectratio=false, clip]{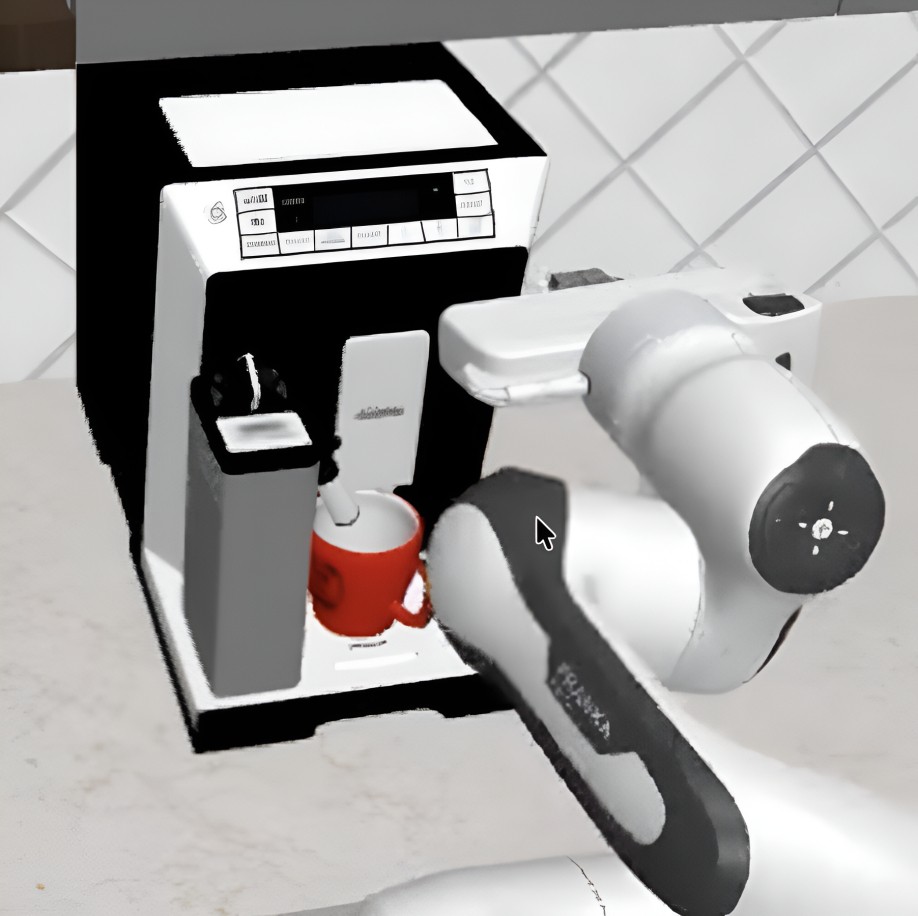}
            \scriptsize \centering \mbox{\strut CoffeePressButton}
        \end{minipage}
        \\[0.5em] %
        \begin{minipage}{0.48\linewidth}
            \includegraphics[width=\linewidth, height=0.95\linewidth, keepaspectratio=false, clip]{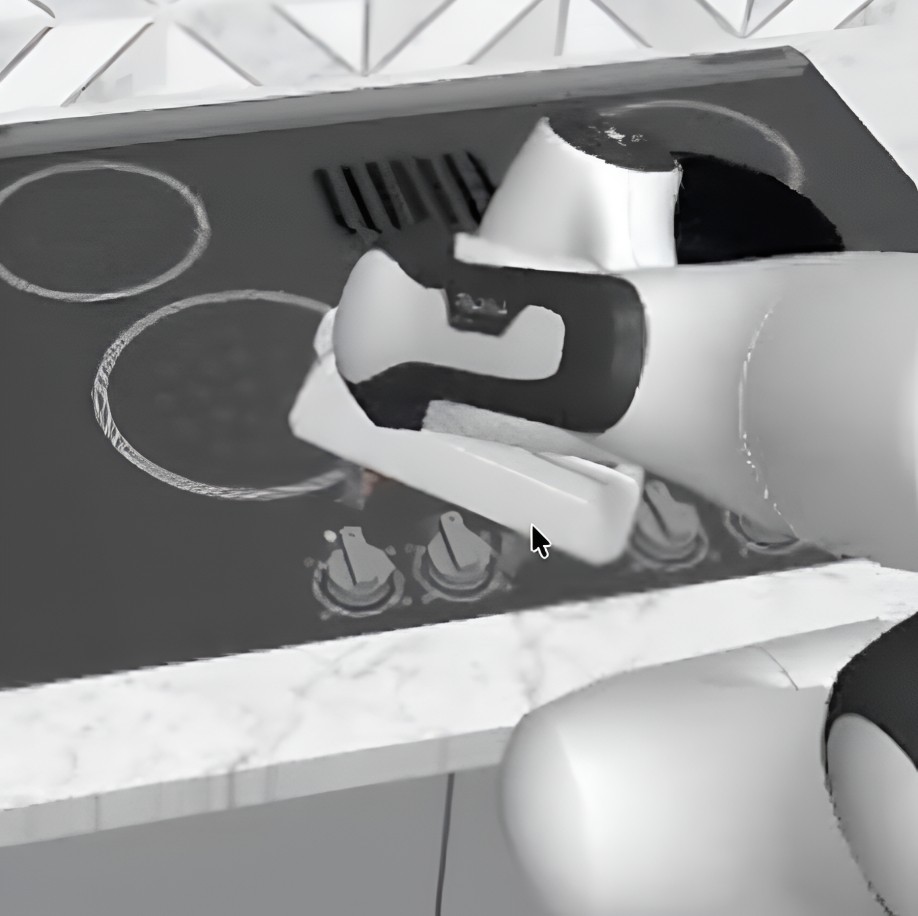}
            \scriptsize \centering \mbox{\strut TurnOnStove}
        \end{minipage}
        \hfill
        \begin{minipage}{0.48\linewidth}
            \includegraphics[width=\linewidth, height=0.95\linewidth, keepaspectratio=false, clip]{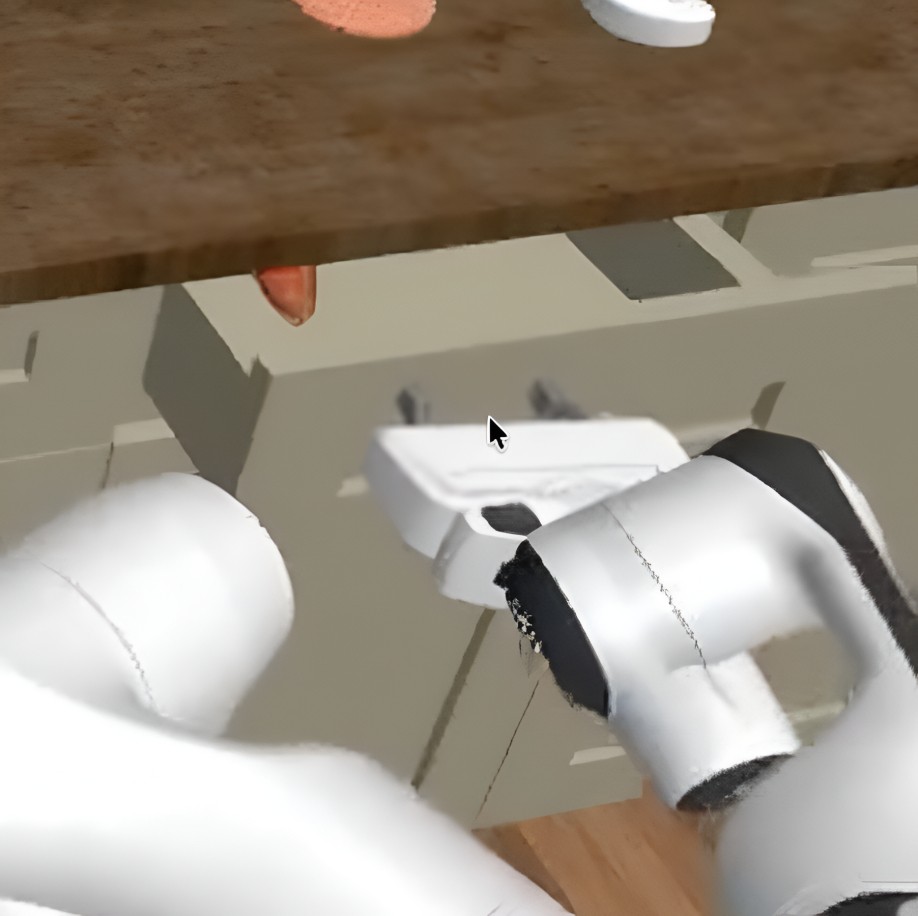}
            \scriptsize \centering \mbox{\strut CloseDrawer}
        \end{minipage}
    \end{tcolorbox}
    \hfill
    \begin{tcolorbox}[col_style, width=0.32\linewidth]
        \centering
        \textbf{ManiSkill3}\\[0.5em]
        \begin{minipage}{0.48\linewidth}
            \includegraphics[width=\linewidth, height=0.95\linewidth, keepaspectratio=false, clip]{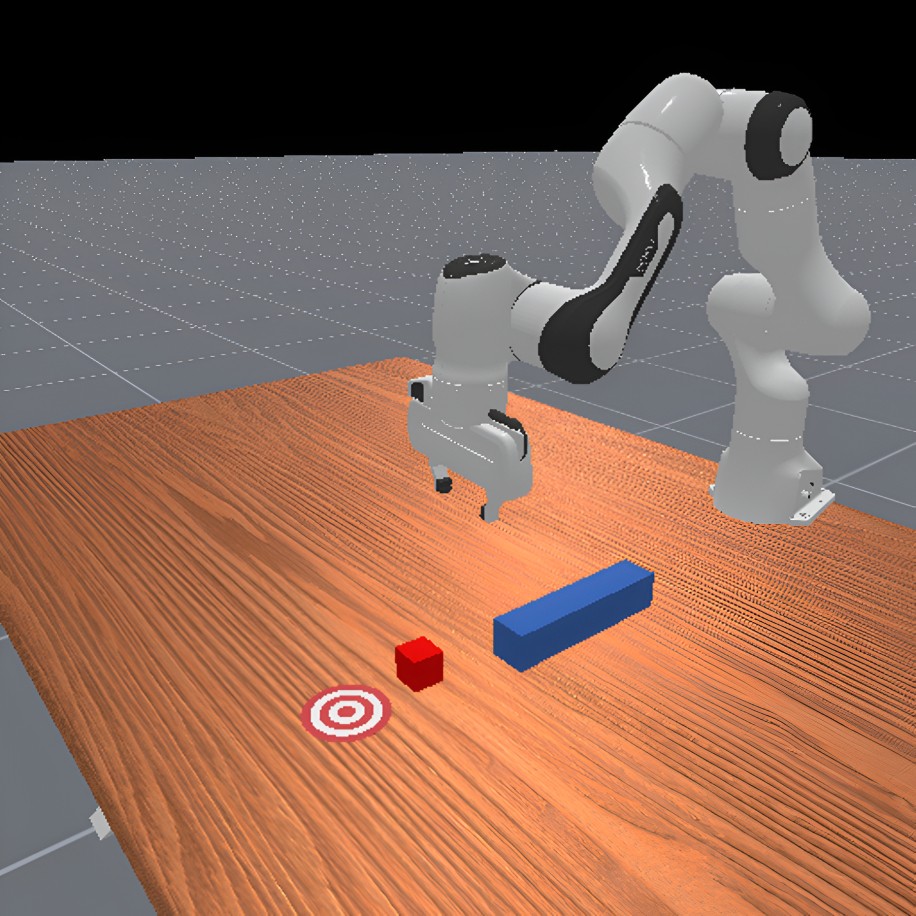}
            \scriptsize \centering \mbox{\strut PokeCube-v1}
        \end{minipage}
        \hfill
        \begin{minipage}{0.48\linewidth}
            \includegraphics[width=\linewidth, height=0.95\linewidth, keepaspectratio=false, clip]{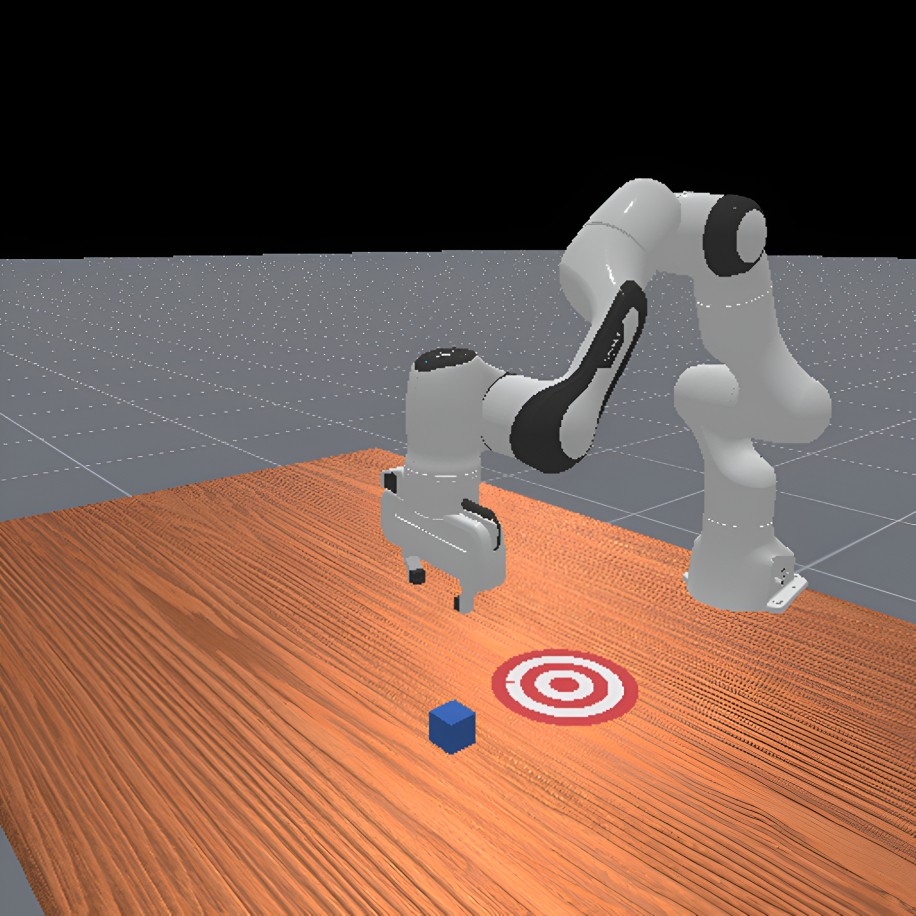}
            \scriptsize \centering \mbox{\strut PullCube-v1}
        \end{minipage}
        \\[0.5em]
        \begin{minipage}{0.48\linewidth}
            \includegraphics[width=\linewidth, height=0.95\linewidth, keepaspectratio=false, clip]{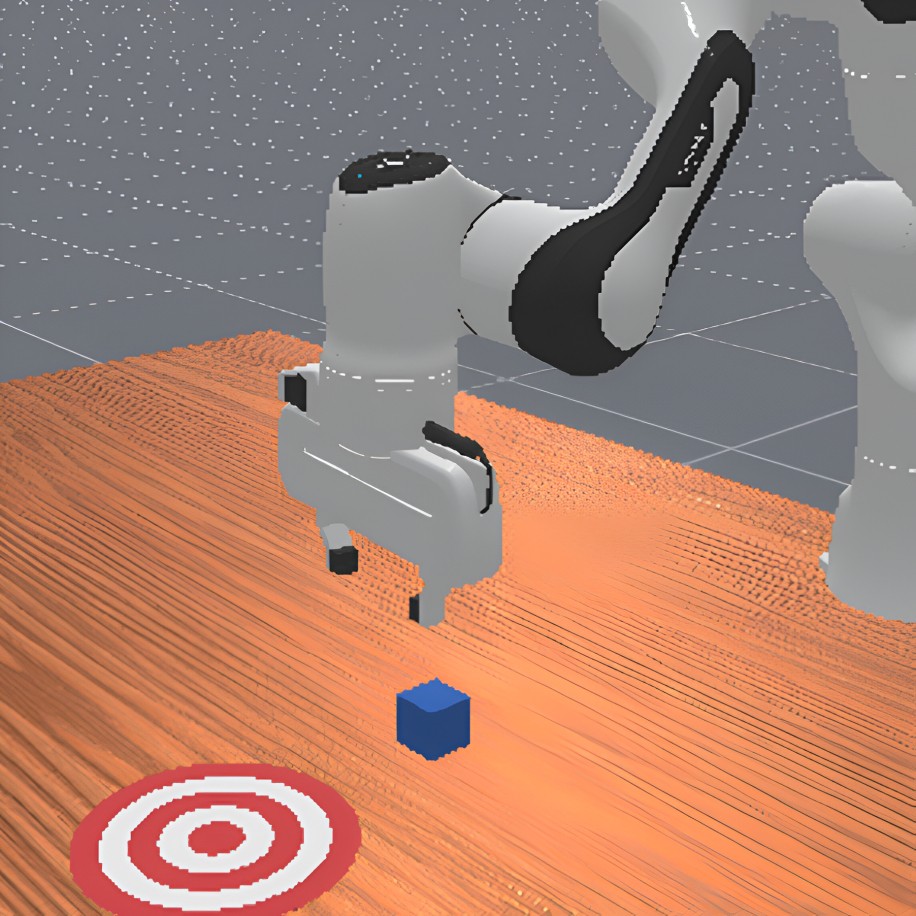}
            \scriptsize \centering \mbox{\strut PushCube-v1}
        \end{minipage}
        \hfill
        \begin{minipage}{0.48\linewidth}
            \includegraphics[width=\linewidth, height=0.95\linewidth, keepaspectratio=false, clip]{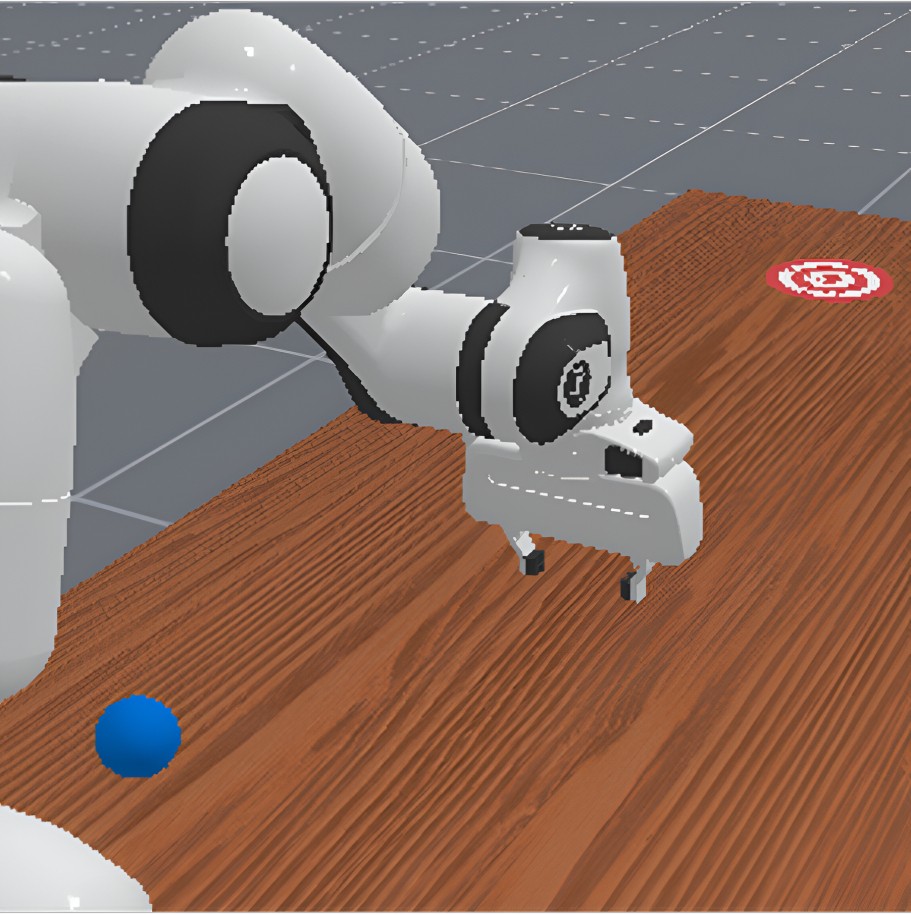}
            \scriptsize \centering \mbox{\strut RollBall-v1}
        \end{minipage}
    \end{tcolorbox}
    \hfill
    \begin{tcolorbox}[col_style, width=0.32\linewidth]
        \centering
        \textbf{Real World}\\[0.5em]
        \begin{minipage}{0.48\linewidth}
            \includegraphics[width=\linewidth, height=0.95\linewidth, keepaspectratio=false, clip]{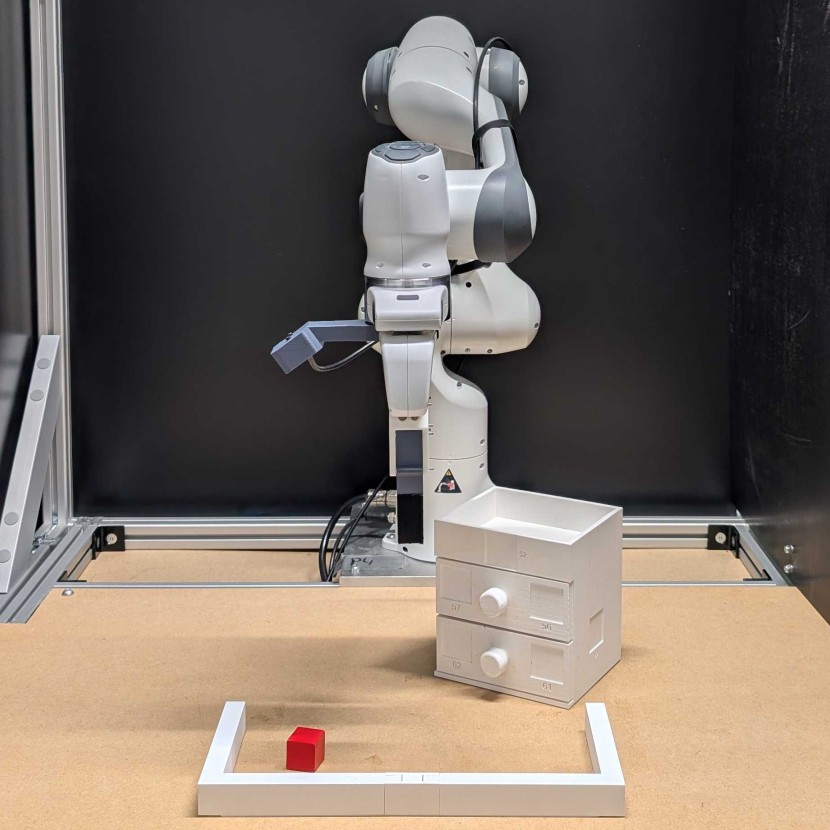}
            \scriptsize \centering \mbox{\strut Drawer}
        \end{minipage}
        \hfill
        \begin{minipage}{0.48\linewidth}
            \includegraphics[width=\linewidth, height=0.95\linewidth, keepaspectratio=false, clip]{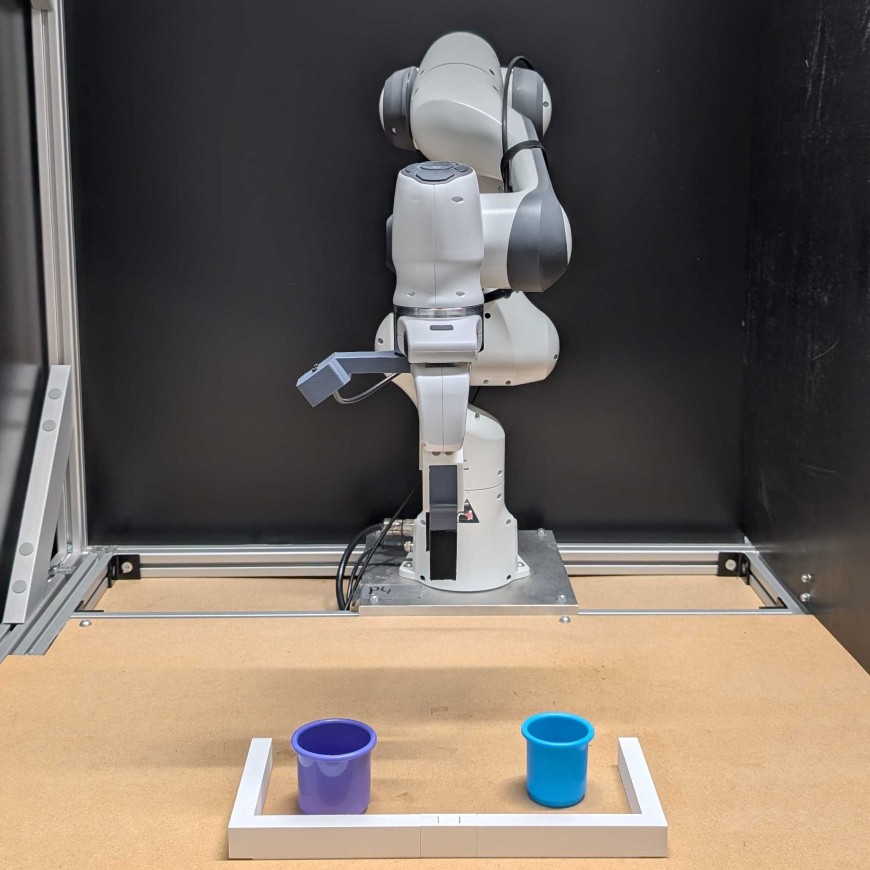}
            \scriptsize \centering \mbox{\strut Cup-Stacking}
        \end{minipage}
        \\[0.5em]
        \begin{minipage}{0.48\linewidth}
            \includegraphics[width=\linewidth, height=0.95\linewidth, keepaspectratio=false, clip]{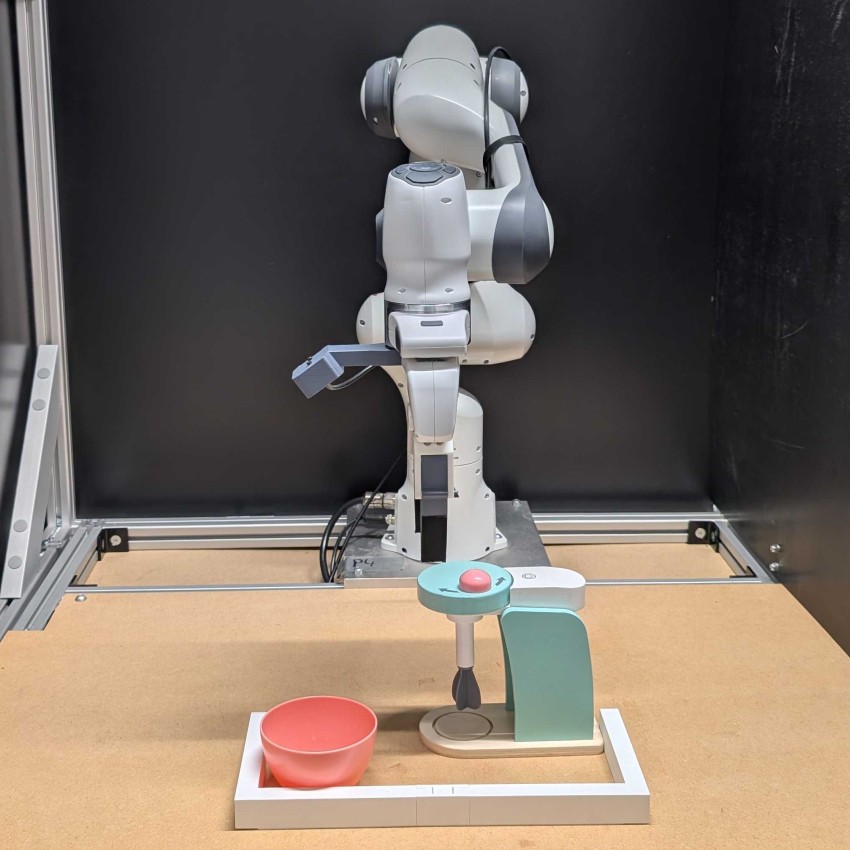}
            \scriptsize \centering \mbox{\strut Arranging}
        \end{minipage}
        \hfill
        \begin{minipage}{0.48\linewidth}
            \includegraphics[width=\linewidth, height=0.95\linewidth, keepaspectratio=false, clip]{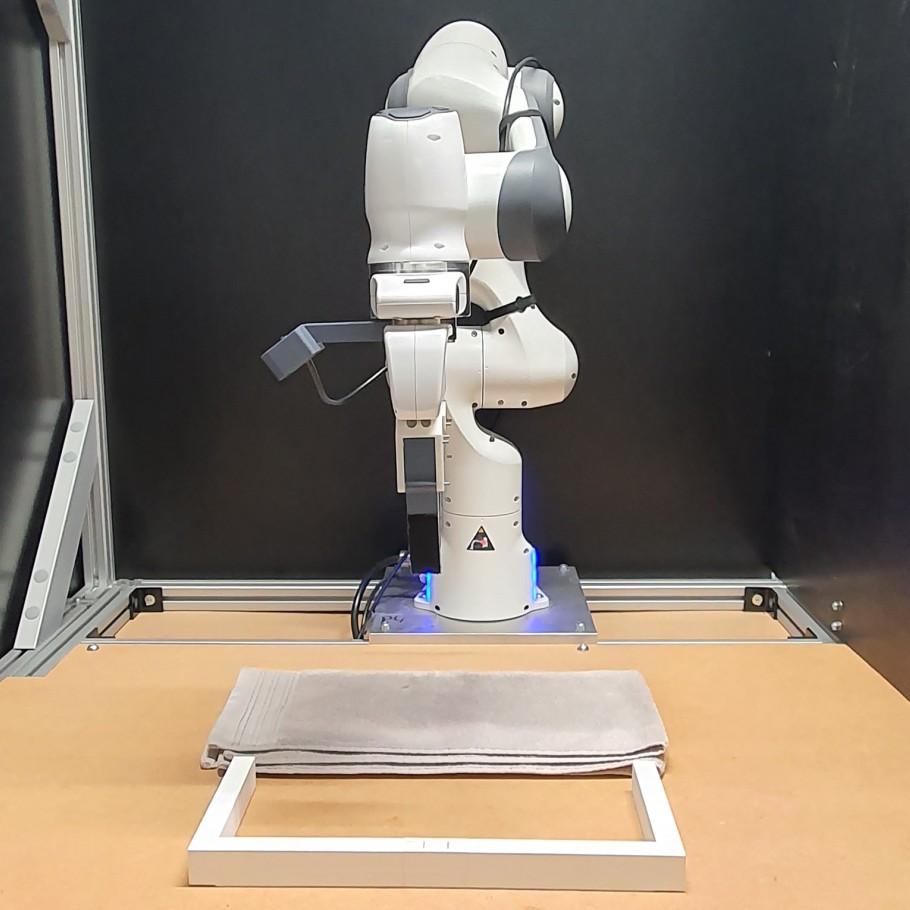}
            \scriptsize \centering \mbox{\strut Folding}
        \end{minipage}%
    \end{tcolorbox}
     
    \caption{
        Overview of all evalution tasks from RoboCasa, ManiSkill3, and Real World benchmarks.
        \textbf{Left}: $4$ of $16$ RoboCasa tasks used for evaluation.
        \textbf{Middle}: all evaluated ManiSkill3 tasks.
        \textbf{Right}: starting configurations for all real-world tasks.
    }
    \label{fig:task_suites}
\end{figure*}

\subsection{Problem Formulation}

\gls{il} aims to learn a policy from expert demonstrations.
We are given a dataset of $N$ expert trajectories \mbox{$\traj_i = \left(\goal_i, (\obs_1, \act_1), (\obs_2, \act_2), \dots, (\obs_{\ell_i}, \act_{\ell_i})\right)$}, where $\ell_i$ is the trajectory length and $\goal_i$ is the language description for the trajectory.
We aim to learn a policy $\pi(\acts|\obs,\goal)$ that maps observations $\obs$ and embedded goal $\goal$ to actions $\acts$.
Predicting sequences of actions, i.e. action chunking, where $\acts = (\act_k, \act_{k+1}, \dots, \act_{k+H})$ with current time step $k$ and horizon $H$, results in more temporally correlated trajectories than predicting individual actions~\citep{zhao2023learning}.
Each observation $\obs$ contains depth images from $M$ cameras.
In combination with the camera intrinsic and extrinsic parameters from calibration, we can construct any desired 3D observation representation from these depth images.

\subsection{Score-Based Diffusion}

To learn policies from expert demonstrations, we use the typical Elucidated Diffusion Models (EDM) framework~\citep{karras2022elucidating, reuss2023goal} for score-based action diffusion conditioned on scene observations.
Diffusion models learn to generate new samples by iteratively reversing a Gaussian perturbation process.
Under this framework, the policy $\pi_{\theta}(\acts | \obs)$ successively denoises actions generated from Gaussian noise back to the data manifold.
The noising and its inverse process can be expressed with the following \gls{sde}
\begin{equation}
\diff \acts_{\pm}  =  \big( \pm \beta_t \sigma_t - \dot{\sigma}_t  \big) \sigma_t \nabla_a \log p_t(\acts | \obs, \goal) dt + \sqrt{2 \beta_t} \sigma_t d\omega_t,
\label{eq: conditional diffusion SDE}
\end{equation}
where $\diff \acts_{+}$ and $\diff \acts_{-}$ describe the forward and reverse process, respectively, $\beta_t$ determines the noise injection rate at diffusion time step $t$, $d\omega_t$ represents infinitesimal Gaussian noise, and $\nabla_a \log p_t(\acts | \obs, \goal)$ denotes the score function of the diffusion process.
To learn that score function, we train a neural network $D_{\theta}(\acts + \boldsymbol{\epsilon}, \obs, \goal, \sigma_t)$ via score matching~\citep{6795935}:
\begin{equation}
\label{eq: SM objective}
\mathcal{L}_{\text{SM}} = \mathbb{E}_{\mathbf{\sigma}, \acts, \boldsymbol{\epsilon}} \big[ \alpha (\sigma_t) \newline  | D_{\theta}(\acts + \boldsymbol{\epsilon}, \obs, \goal, \sigma_t)  - \acts  |_2^2 \big]\text{.}
\end{equation}

During policy sampling, i.e. during the reverse process, action samples are guided towards high-density regions of the data distribution by following the score function $\nabla_a \log p_t(\acts | \obs, \goal)$.
Therefore we can generate new action sequences beginning with Gaussian noise by iteratively denoising the action sequence with a numerical probability flow \gls{ode} solver.
We utilize the DDIM-solver to enable efficient action denoising in few steps~\citep{song2021denoising}.

\subsection{Point Clouds}

Given a set of depth images from $M$ cameras $D^{(0)}, \ldots, D^{(M)} \in \mathbb{R}^{W \times H}$ as well as the camera intrinsics $K^{(0)}, \ldots, K^{(M)} \in \mathbb{R}^{3\times 3}$, we first construct point clouds $X^{(j)} \in \mathbb{R}^{W H \times 3}$ in each camera's local coordinate frame via unprojection, i.e.,
\begin{equation}
    X^{(m)}_{i W + j} = D^{(m)}_{i,j} \cdot (K^{(m)})^{-1} \left(i, j, 1\right)^\textbf{T}\text{.}
\end{equation}
By multiplying each point cloud with its corresponding extrinsic matrix, we can transform it from the camera coordinate frame to the world frame.
The final point cloud $X$ is obtained by concatenating point clouds from all $M$ views.

We treat point clouds as graphs, where the coordinates XYZ are the node features $\mathbf{x}^0$.
This allows us to formulate the point cloud encoder as a message-passing \acrfull{gnn}~\citep{gnn}, a flexible framework that encompasses numerous well-known architectures.
The point cloud encoder returns a tokenized embedding of the observed point cloud $\{\mathbf{T}_i\} \in \mathbb{R}^{n \times d}$.
For PointPatch architectures, a positional encoding based on the patch center is also added to each token.

\subsection{Fourier Feature Mapping}
\label{subsec:fourier_feature_mapping}

Neural networks are biased toward learning low-frequency components first, while high-frequency components converge slowly or may not be learned at all~\citep{pmlr-v97-rahaman19a, fourier_features}. 
However, an \gls{il} policy parametrized by a neural network may need to learn a high frequency function to represent a sharp decision boundary, such as whether to reposition a grasped object or insert it.
In 3D point clouds, a Fourier feature mapping allows the network to better distinguish points that are otherwise similar in Cartesian space.
For a denoising diffusion model, this would allow the network to represent a score function that is a high-frequency function of the scene geometry, though not necessarily of the actions.

In contrast to previous work that adds Fourier features to specific, novel architectures~\citep{adapt3r}, we hypothesize that applying a Fourier feature mapping to Cartesian points feature benefits essentially \textit{any} point cloud-based policy.
We adopt a \gls{nerf}-style, axis-aligned Fourier feature mapping \citep{mildenhall2021nerf}.
Let \mbox{$\mathbf{p} = (x,y,z) \in \mathbb{R}^3$} define a Cartesian point. %
The encoding function \mbox{$\gamma:\mathbb{R}\to\mathbb{R}^{2L}$} applies sinusoids of different wavelengths $\lambda_k$ separately to the three coordinate values in $\mathbf{p}$ via the transformation 
\begin{equation}
    \begin{split}
        \gamma_k(x) &\;=\; \biggl[ \sin \Bigl( \frac{2 \pi x}{\lambda_k } \Bigr), \, \cos \Bigl( \frac{2 \pi x}{\lambda_k} \Bigr) \biggr]^\textbf{T}, \\
    \lambda_k &\;=\; \lambda_\text{max} \,
    \left( \frac{\lambda_\text{min}}{\lambda_\text{max}} \right)^{\tfrac{k-1}{L-1}}, \qquad    k = 1, \dots, L\text{,}
    \end{split}
    \label{eq:ff_scalar}
\end{equation}
where $L$ is the number of frequency bands.

As the Fourier feature mapping is periodic, the point cloud must be bounded by the interval $[-\lambda_\text{max}/2,\lambda_\text{max}/2]$ to ensure unique features.
If this is not possible, the input Cartesian coordinates can be concatenated with the Fourier features, which always yields a unique mapping.

\subsection{Data Augmentation}
\label{subsec:data_augmentation}

The choice of wavelengths $\lambda_k$ is essential, as too short wavelengths may cause the network to overfit on the data, while too long wavelengths may not resolve the spectral bias~\citep{fourier_features}. %
Previous work using Fourier features for consistency models has even observed training instability with some hyperparameters~\cite{song2024improved}.
Instead of carefully tuning the wavelengths to each task, we choose a consistent set of wavelengths and use data augmentation to train the network to ignore frequencies that do not contain useful information.
To achieve this, we apply VariableJitter~\citep{kirurc}, which samples the noise scale $\sigma \sim \mathcal{U}(0, \sigma_\text{max})$ from a uniform distribution for each point cloud.
Compared to uniform jitter, which applies noise $\epsilon \sim \mathcal{U}(-\sigma, \sigma)$ drawn from a fixed distribution to each point, this process avoids the difficulty of tuning the amplitude of typical uniform jitter.
It ensures a trade-off between augmenting the data to reduce overfitting and ensuring there is no gap between training and testing data.

\section{Experiments}
\label{sec:experiments}

\begin{figure}[tb!]
    \centering
    \begin{minipage}[t]{0.51\linewidth}
        \centering
        \vspace{0pt}
        \includegraphics[width=\textwidth]{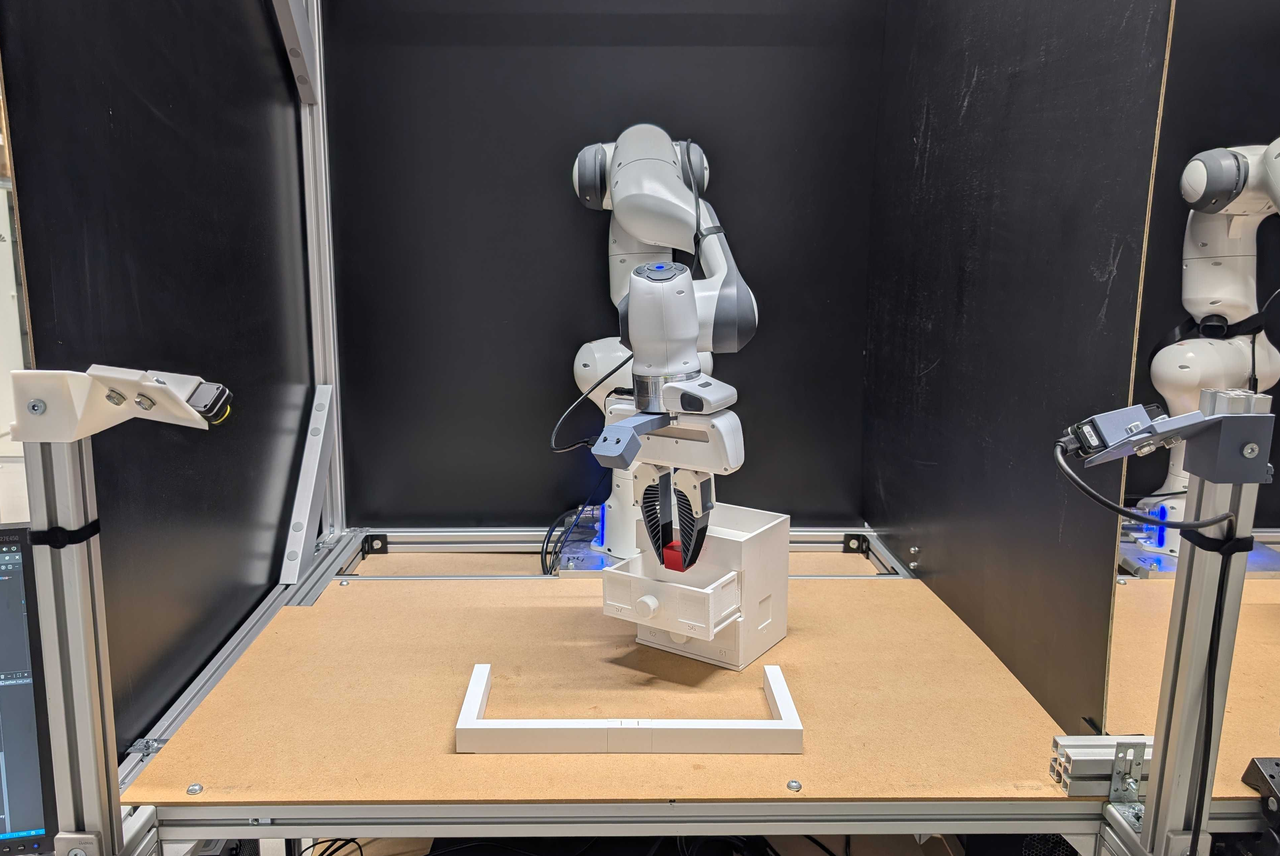}
        \label{fig:drawer_overview}
    \end{minipage}%
    \hfill
    \begin{minipage}[t]{0.454\linewidth}
        \centering
        \vspace{0pt}
        \begin{minipage}[b]{0.48\linewidth}
            \centering
            \includegraphics[width=\linewidth]{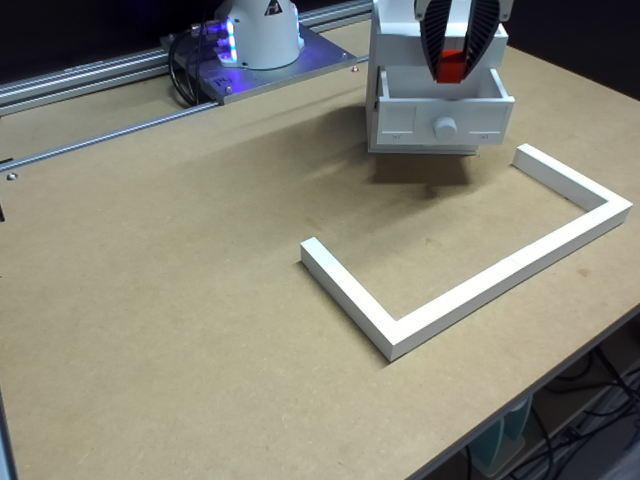}\\[2pt]
            \includegraphics[width=\linewidth]{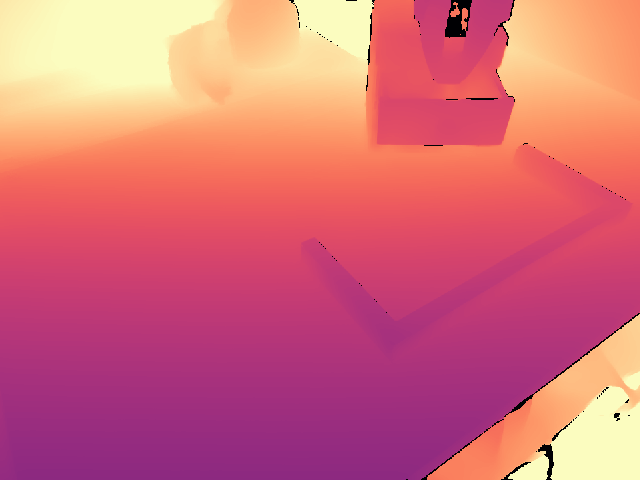}
        \end{minipage}\hfill%
        \begin{minipage}[b]{0.48\linewidth}
            \centering
            \includegraphics[width=\linewidth]{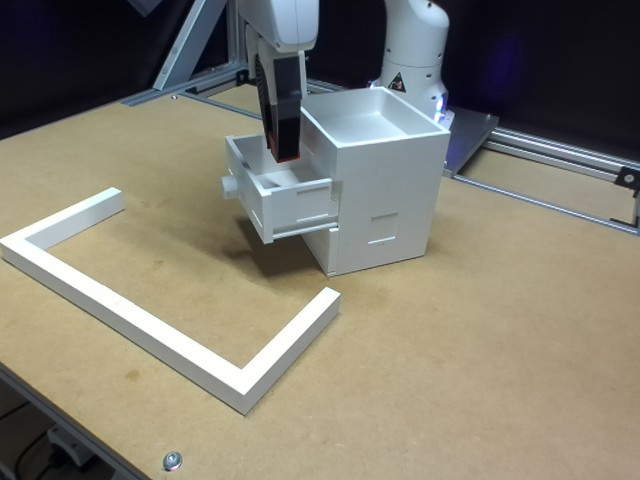}\\[2pt]
            \includegraphics[width=\linewidth]{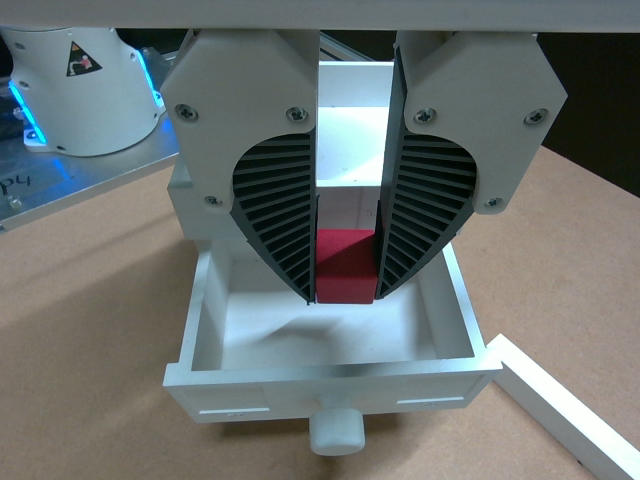}
        \end{minipage}\hfill%
    \end{minipage}\hfill%
    \vspace{-0.2cm}
    \caption{
        \textbf{Left}: Setup for the real-world drawer experiments. 
        \textbf{Right}: RGB views from left, right and gripper cameras and depth view from the left camera.
    }
    \label{fig:results_real_setup}
    \vspace{-0.2cm}
\end{figure}

\subsection{Benchmarks and Datasets}
We evaluate our approach on two widely used simulation benchmarks, RoboCasa~\citep{robocasa2024} and ManiSkill3~\citep{tao2025maniskill}, as well as on four challenging real-world tasks~\citep{pointmappolicy}.
Figure~\ref{fig:task_suites} visualizes exemplary tasks from all task suites.
All models are trained in a multi-task setting, where the policy is provided a goal description in text form.
RoboCasa and real world tasks include two static cameras and an in-hand camera, while ManiSkill3 tasks use only a static camera.
In order to highlight the effect of Fourier features on 3D representations, we do not provide color features in simulation.
Appendix~\ref{app:tasks} provides additional details on all tasks.

\textbf{RoboCasa.}
RoboCasa~\citep{robocasa2024} includes high-precision, long-horizon manipulation tasks in visually rich kitchen scenes.
We focus on $16$ atomic tasks that stress fine geometric alignment and contact, which is where spectral bias is most detrimental.
For each task we use $50$ human-collected demonstrations provided by RoboCasa.

\textbf{ManiSkill3.}
We further test on ManiSkill3~\citep{tao2025maniskill}, where we evaluate on four tasks covering grasping and tool usage.
Since the majority of tasks use color information to indicate some aspect of the target, we map the target's Cartesian coordinates to Fourier features and pass this as an additional goal token.
We train on $500$ demonstrations per task generated by an expert policies.

\textbf{Real World.}
Finally, we adapt four challenging real-world tasks~\citep{pointmappolicy} that feature long horizons, multiple phases, and precise manipulation.
Figure \ref{fig:results_real_setup} displays our setup, which consists of two static Zed Mini cameras and a RealSense D405 as an in-hand camera.
Each task is comprised of distinct sub-tasks with different goal descriptions, such as ``stack the red cup in the blue cup" for the ``Cup-Stacking" task.
For these complex tasks involving color information, we adopt a multi-modal approach utilizing RGB images and point clouds.
Early experiments with unimodal methods were not successful.
We use between $75$ and $102$ human demonstrations, depending on the task.

\textbf{Preprocessing.}
Points beyond a maximum depth ($2$m in ManiSkill3 and the real world and $10$m in RoboCasa) are removed.
In ManiSkill and the real world, point clouds are further cropped to remove the irrelevant background points, and in ManiSkill, the table surface is also removed.
We apply random downsampling to 32768 points followed by voxel downsampling with a voxel size of \SI{1}{\cm}.
We sample a $\sigma_\text{max}$ for VariableJitter up to \SI{5}{\mm} for ManiSkill, \SI{2}{\mm} for RoboCasa, and \SI{1}{\mm} for the real world, which augments the training data manifold while preserving crucial geometric details preserved.
Camera observations are resized to $128 \times 128$ for ManiSkill tasks and $224 \times 224$ for RoboCasa.

\subsection{Architectures and Baselines}

\begin{figure*}[ht]
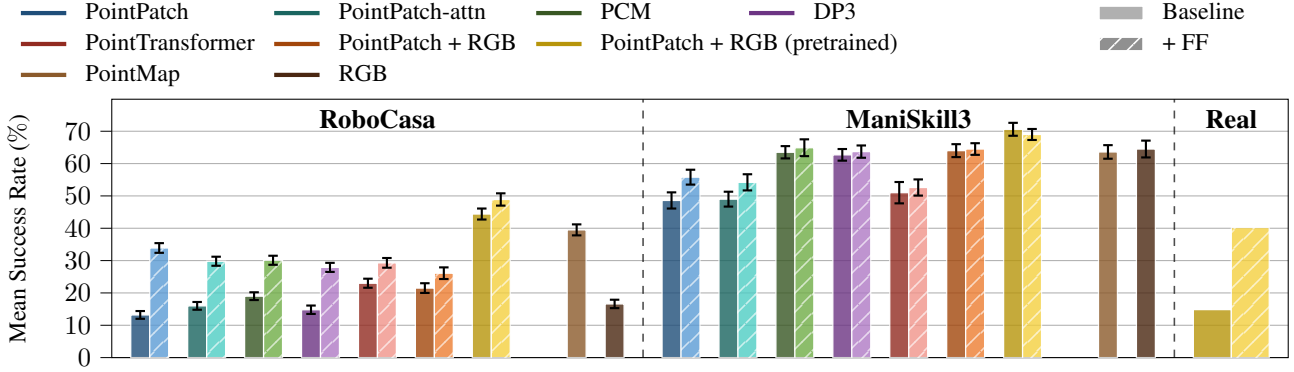

    \centering
    \input{plots/main_results_legend.tex} %
    \input{plots/main_results_bars.tex} %
    \caption{
        Mean success rate across all tasks of 3D encoders with and without Fourier features on RoboCasa (left), ManiSkill3 (middle), and the real world (right).
        Methods using Fourier Features are marked via hatched bars, and methods are displayed in the order of the legend at the top. 
        Across diverse tasks and architectures, Fourier features provide a consistent and meaningful benefit to task performance.
    }
    \label{fig:main_results}
\end{figure*}

We instantiate the denoising diffusion model as a decoder-only transformer.
The transformer receives as input a goal token, a noise level token, $H$ noisy action tokens, where $H$ is the action horizon, and a number of observation tokens that are provided by a separate architecture, such as a point cloud encoder.
A frozen CLIP RN-50 model~\citep{radford2021learning} is used to embed the goal description into a token, while the noise level is encoded as in DDPM~\citep{ho2020denoising}.
A learned token position embedding is added onto each token, except for observation tokens when the number of tokens is variable.
After the transformer layers, the final $H$ tokens are passed through a linear layer to arrive at denoised actions.

To highlight the effect of Fourier features for diffusion \gls{il}, we use the same diffusion backbone across all experiments and vary only the observation encoder across experiments. 
All architectures are instantiations of message neural networks with different aggregations and parameterizations for the learnable functions.
We evaluate three variants as representative examples of the point-patching paradigm, which differ mainly in how they aggregate across patch tokens. \textbf{PointPatch} does not aggregate the patch tokens before the denoising step~\cite{pang2022masked, pointBERT, chen2023pointgpt, zhou2024uni3d, sugar}, allowing each denoising step to attend to different parts of the observation.
\textbf{PointPatch-attn} aggregates using attention pooling~\cite{gyenes2024pointpatchrl} to produce $3$ tokens in all cases, which decreases computational cost.
Lastly, the encoder referred to as ``PointNet" in Point Cloud Matters~\citep{zhu2024point}, which we call \textbf{PCM}, uses max pooling to aggregate across point patches.
Furthermore, we apply Fourier features to existing architectures \textbf{DP3}~\cite{Ze2024DP3}, which applies max pooling across all points to create a single token, and \textbf{PointTransformer}, which uses attention to aggregate local regions iteratively.
We make only minimal, necessary changes to each architecture, and apply Fourier feature mappings to absolute and/or relative point coordinates as required by the method.
RGB images and pointmaps are tokenized using ConvNeXt V2 nano~\cite{Woo2023ConvNeXtV2}, a fully convolutional architecture.
In \textbf{PointPatch + RGB}, we tokenize RGB images and point clouds separately, and concatenate tokens from the RGB modality to the tokens from the point cloud modality.
More details on architectures are available in Appendix \ref{app:architectures}.

\subsection{Setup and Experiments.}
\textbf{Research Questions.}
Our experiments are designed to answer the following research questions:
\textbf{Q1)} Do Fourier features yield consistent benefits across point cloud encoders and benchmarks?
\textbf{Q2)} Does the benefit of Fourier feature mappings translate from simulation to real-world tasks?
\textbf{Q3)} How should these features be parameterized?
\textbf{Q4)} How task-dependent are their effects?

To answer these questions, we first explore a series of simulation benchmarks and evaluate several point cloud encoders with and without Fourier features.
Next, we train our policies on real-world data and evaluate them on a real robot.
Finally, we then conduct an extensive parameter study and spectral analysis of our policies. %

\textbf{Training and Evaluation.}
Each method is trained with $5$ random seeds, and we test performance at $3$ checkpoints.
We measure the average success rate across $50$ rollouts for RoboCasa and $100$ rollouts for ManiSkill and select the best-performing checkpoint for each seed.
We use bootstrapping and report the interquartile mean with $95\%$ confidence intervals.
See Appendix \ref{app:sim_results} for further details.

\textbf{Fourier Features.}
We use a fixed set of $L{=}16$ frequency bands with log-spaced wavelengths between $\lambda_{\max}{=}$\SI{4.0}{\m} and $\lambda_{\min}{=}$\SI{2.0}{\cm} for all experiments.
The choice of $\lambda_{\max}$ ensures both the sin and cos components of the largest band are unique within the task space, which is typical bounded to roughly $[-2, 2]^3$, while $\lambda_{\min}$ is small enough to discriminate neighboring voxels.
The resulting $3 {\times} (2L) {=} 96$ Fourier features per Cartesian point encode positions across this scale range, ranging from a global encoding at $\lambda_{\max}$ to a voxel-level encoding at $\lambda_{\min}$.
See Appendix \ref{app:hyperparameters} for additional hyperparameters.

\section{Results}

\textbf{Simulation Results.}
Figure \ref{fig:main_results} shows average success rates over all tasks for RoboCasa and ManiSkill3.
Tables~\ref{table:robocasa_results} and~\ref{table:maniskill_results} in the appendix provide detailed, per-task results for both benchmarks, respectively.
In RoboCasa, we observe that Fourier feature mappings significantly boost both aggregate success rate and success rates on a large number of individual tasks.
For example, PointPatch on CloseDrawer improves from $34\%$ to $72\%$, and TurnOffSinkFaucet improves from $28\%$ to $63\%$, while the overall average increases from $13\%$ to $34\%$.
Similarly, the OpenDrawer task improves from almost no success at all to $12\%$.
Despite using substantially different architectures, DP3 and PCM show similar trends, with both methods clearly benefitting from the use of Fourier features. 
While the relative ranking of different architectures is task-dependent, we find that the improvements from Fourier features are relatively consistent across architectures and task difficulties.
We hypothesize that these improvements come from the Fourier mappings exposing high-frequency geometric cues to the denoising model, which alleviates their spectral bias and thus allows learning sharper, more meaningful token representations.
Multimodal encoders that use RGB in conjunction with point clouds also benefit from Fourier features, despite the fact that the convolutional RGB tokenizer already has the capacity to represent high frequency functions.
This highlights the potential for integrating Fourier features into large, multi-modal models trained on internet-scale data.
On ManiSkill3 tasks, we observe only minor improvements to PointPatch and PointPatch-attn but no significant improvement to other architectures.
We hypothesize this is partly due to saturation on these relatively simple tasks.

\textbf{Real World Results.}
We evaluate our best-performing method, PointPatch + RGB, with and without Fourier features for $16$ rollouts on each real world task.
Full results are in Appendix \ref{app:real_results}.
Table \ref{tab:results_real_small} shows that policies with Fourier features achieve significantly higher scores on all tasks, with an aggregate improvement from $14.8\%$ to $40.23\%$.
The policy without Fourier demonstrates especially poor performance on the Cup-Stacking task, often because it knocks the cup over while trying to grasp it.
The least benefit is observed on the Folding task, which also requires the least geometric precision to carry out.
These results demonstrate the Fourier features do not overfit on artifacts and are robust to real world conditions such as noisy depth measurements, occlusion artifacts, and camera miscalibration.
Surprisingly, Fourier features greatly increase the performance of the PointPatch + RGB (pretrained) encoder on real world tasks, even though the performance gain on simulation tasks was more minor.
This suggests that Fourier features remain effective even on more difficult tasks and when combined in more complex architectures.
Table \ref{tab:results_real_cup_size} shows further results on the Cup-Stacking task grouped by cup diameter.
As cup size decreases, the benefit of Fourier features increases, albeit with the smallest cups, neither policy is able to solve the task consistently.
This supports our claim that Fourier features allow the encoder to extract geometric details at smaller scales.

\textbf{Qualitative Results.}
We find that policies trained on Fourier feature mappings move faster and more decisively, and more closely imitate the demonstration data.
Policies trained without Fourier features tend to hesitate before making contact with objects, or behave as if they cannot perceive the scene.
In the real world, the policies without Fourier features were often catastrophically unable to learn the data\footnote{We provide rollout videos for real world tasks on our \href{https://fourier-il.github.io/fourier-il/}{project page: https://fourier-il.github.io/fourier-il}.}.
We hypothesize that due to the spectral bias of their \gls{mlp} layers, these policies were overfitting on some global features of the observed point cloud rather than responding to the fine geometry near the end effector.
See Appendix \ref{app:qualitative_results} for a qualitative comparison of results on select RoboCasa tasks.

\section{Analysis}
\label{sec:analysis}

\begin{figure}[htb]
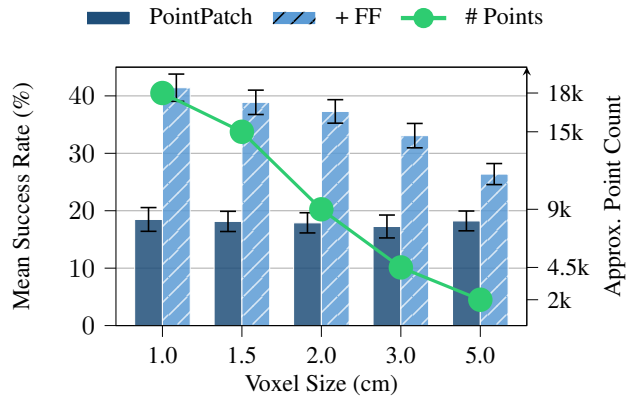

    \centering
    \input{plots/abl_size_legend} %
    \vspace{0pt}
    \input{plots/abl_size} %
    \vspace{-0.4cm}
    \caption{
        Success rates with and without Fourier features on point clouds of different sizes, achieved by voxel downsampling of the observations.
        Larger point clouds contain richer geometric detail, resulting in a larger benefit for Fourier features.
    }
    \label{fig:abl:size}
\end{figure}

\begin{figure*}[htb]
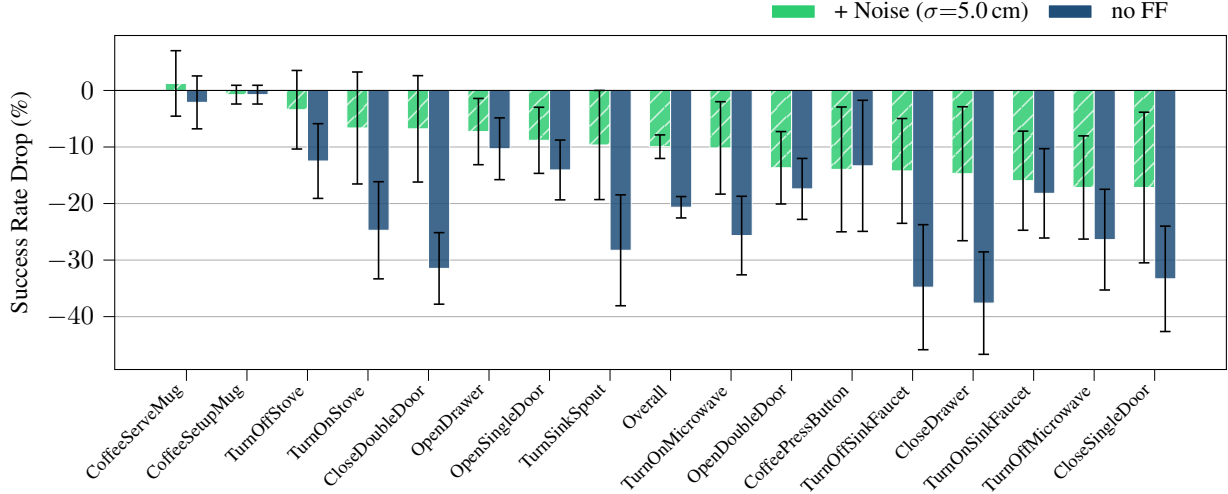

    \centering
    \input{plots/abl_filter_legend} %
    \vspace{0pt}
    \input{plots/abl_filter}
    \caption{
        Absolute drop in success rate for each RoboCasa task resulting from removing Fourier features from the PointPatch policy architecture (\textbf{no FF}) or removing fine geometric information in the observation using Gaussian jitter (\textbf{+ Noise($\boldsymbol{\sigma {=}}$\SI{5.0}{\cm})}).
        Even when high frequency information is removed, Fourier features still provide a meaningful benefit, perhaps by improving the learning dynamics of the policy.
    }
    \label{fig:abl:filter}
    \vspace{-0.2cm}
\end{figure*}

\begin{figure}[htb]
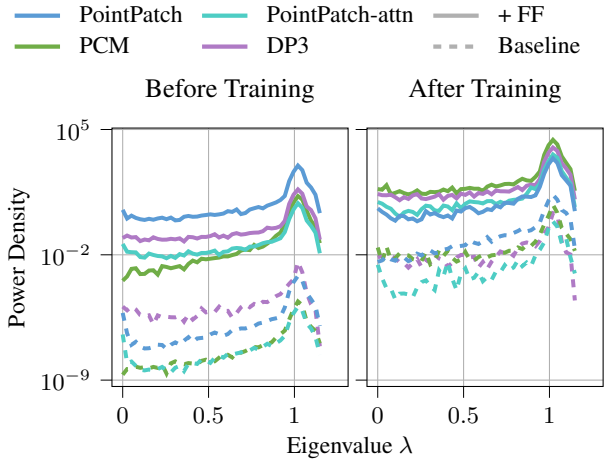

    \centering
    \input{plots/gft_spectra_legend} %
    \vspace{0pt}
    \input{plots/gft_spectra} %
    \caption{
        Graph Fourier spectra of the sensitivities of various architectures with respect to input point coordinates.
        During training, sensitivities increase by several orders of magnitude across all frequencies, and Fourier features also increase sensitivity by several more orders of magnitude relative to the baseline.
        The peak near eigenvalue of 1 indicates the orthogonal response, i.e. the isolated contribution of each point to the prediction.
    }
    \label{fig:gft_spectra}
    \vspace{-0.2cm}
\end{figure}

For additional experiments, we use a reduced set of 8 RoboCasa tasks, utilizing the Pressing Buttons, Turning Levers, and Twisting Knobs task groups.
Unless specified otherwise, we use the PointPatch encoder.
We train a single policy on all $8$ tasks, reporting the mean and $95\%$ bootstrap confidence interval over $5$ seeds for each method.

\subsection{Frequency Analysis}
\label{sec:frequencies}

\textbf{Point Cloud Size.}
Since denser point clouds contain richer geometric detail, we expect Fourier features to provide a greater benefit with larger point clouds.
To reduce point cloud sizes while retaining their overall structure, we increase the voxel size used for voxel downsampling.
Figure \ref{fig:abl:size} shows that point cloud size has a large effect on the performance of Fourier features, with the gap narrowing significantly for point clouds with $2$k points.
This demonstrates that Fourier features are more effective when point clouds contain more geometric detail to extract.
Interestingly, the baseline method is essentially unaffected by the heavy downsampling, suggesting that it does not condition on the geometric details that are removed by this process.

\textbf{Task-Dependent Fine Geometry.}
The previous result raises the question - is there any observable pattern between the characteristics of a task and the effect of Fourier features?
Is the advantage limited to tasks requiring fine manipulation?
To investigate this, we train a policy with Fourier features and add Gaussian jitter with $\sigma{=}$\SI{5.0}{\cm} to the point clouds during training and rollout.
This extreme level of jitter effectively removes any fine geometric information from the observation.
We compare this against the drop in performance to that when Fourier features are not used.

Figure \ref{fig:abl:filter} shows that the advantage of Fourier features is not limited to tasks with fine geometry.
There is only a weak correlation between the effect of removing the policy's ability to condition on geometric details (\textbf{no FF}) and removing the geometric details themselves from the observation (\textbf{+ Noise($\boldsymbol{\sigma {=}}$\SI{5.0}{\cm})}).
Furthermore, the policies with jitter still significantly outperform those without Fourier features, with an average success rate of $24\%$ vs. $13\%$ across RoboCasa tasks.
The broader benefit of Fourier features in the absence of fine geometry may be due to an improvement in learning dynamics in the policy.

\textbf{Spectral Density.} 
To show how architectural changes are reflected in the policy’s sensitivity to spatial frequencies, we adapt a method from \citet{miao2024improving}.
First, we create a ``saliency point cloud", where each point contains the gradient of the predicted actions with respect to its coordinates.
We take the \gls{gft} of this point cloud to quantify how rapidly this gradient varies between neighboring points, and visualize the spectrum of eigenvalues in Figure~\ref{fig:gft_spectra}.
We find that Fourier features increase the sensitivity of the network to high frequencies, as well as low ones. 
Since the sensitivites of all networks increase during training, Fourier features allow networks to learn the data more quickly.
This suggests that the advantage of Fourier features may be due to their increased sensitivity to mid- and low-frequency bands as well.
To further isolate this effect, Appendix~\ref{app:spectral_analysis} also shows a spectral analysis of untrained PointNets on a synthetic unit sphere, confirming that Fourier features shift the model's inductive bias toward high-frequency components ($\lambda > 1.0$) from initialization.

\subsection{Parameter Studies}

\begin{figure}[htb!]
    \centering
    \input{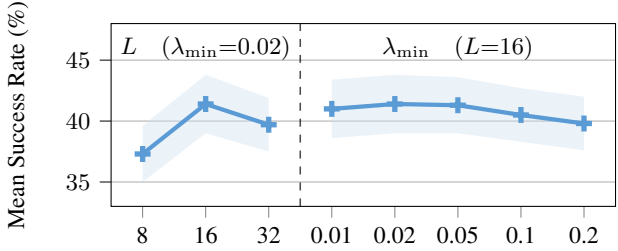}
    \caption{
        Parameter study of different Fourier feature wavelength configurations.
        Performance is robust to different numbers $L$ of log-spaced wavelengths (\textbf{left}), as well as to the minimum wavelength $\lambda_{\text{min}}$ \textbf{(right)} around our default of $\lambda_{\text{min}}{=}0.02$, $L{=}16$.
        \vspace{-0.3cm}
    }
    \label{fig:wavelength_ablations}
\end{figure}

\begin{table}[htb!]
    \centering
    \caption{
        Parameter study on jitter augmentation and Fourier feature encoding. 
        Our method uses logspaced Fourier Features (FFs) and VariableJitter.
        \textbf{Top}:
        While VariableJitter slightls improves performance, data augmentation appears not to be essential.
        \textbf{Bottom}:
        Log-spaced, axis-aligned frequencies perform better than sampling randomly from a Gaussian (RFF).
        Learning the frequencies either directly or with \gls{spe} does not show a consistent benefit.
    }
    \label{tab:abl:jitter_rffs}
    \resizebox{0.95\linewidth}{!}{%
\begin{tabular}{lc}
    \toprule
    \textbf{Configuration} & \textbf{Mean Success Rate (\%)} \\
    \midrule
    \textbf{Ours} & \textbf{41.4 $\pm$ 2.4} \\
    \midrule
    \textit{Jitter Augmentation} \\
    no FFs, no jitter      & $17.5 \pm 1.7$ \\
    no FFs, random jitter  & $17.0 \pm 1.6$ \\
    no FFs, VariableJitter & $18.5 \pm 2.1$ \\
    FFs, no jitter         & $39.9 \pm 2.3$ \\
    FFs, random jitter     & $38.9 \pm 2.2$ \\
    \midrule
    \textit{Fourier Feature Encoding} \\
    log-spaced + SPE       & $37.2 \pm 2.2$ \\
    RFF                    & $24.0 \pm 2.0$ \\
    RFF + learned          & $22.9 \pm 1.8$ \\
    RFF + SPE              & $22.5 \pm 1.8$ \\
    RFF + Cartesian        & $23.4 \pm 1.9$ \\
    \bottomrule
\end{tabular}
}

\end{table}

\textbf{Jitter.}
To investigate the robustness of Fourier features, we consider different options for the jitter data augmentation, namely no jitter, standard random jitter drawn from $\sim \mathcal{U}(-\sigma_\text{max}, \sigma_\text{max})$, and VariableJitter as described in Subsection~\ref{subsec:data_augmentation}. 
Table~\ref{tab:abl:jitter_rffs} (\textbf{top}) shows that data augmentation may not be essential when using Fourier features, although VariableJitter may provide a minor boost over random jitter.

\textbf{Wavelengths.}
We experiment with different numbers of wavelengths and a range of minimum wavelengths in Figure~\ref{fig:wavelength_ablations} (\textbf{middle}).
While $16$ wavelengths perform somewhat better than $8$ or $32$, Fourier features are robust to a wide range of minimum wavelengths.
Unlike in related work using Fourier features for \glspl{nerf}, we do not see any benefit to task-specific tuning of wavelength ranges~\cite{sun2024learning}.
Together with the above, this consistency indicates that Fourier features used in diffusion \gls{il} are relatively insensitive to hyperparameters or changes in the training setup.

\textbf{Learned and Gaussian Fourier Features.}
\citet{sun2024learning} find that optimizing the frequencies directly through gradient descent is suboptimal, and they instead propose \glsf{spe}, which applies a linear layer and a sinusoidal non-linearity instead.
Furthermore, instead of purely on-axis frequencies, \citet{fourier_features} suggest randomly sampling frequency vectors $\mathbf{v} \sim \mathcal{N}(0, \sigma^2)$ from an isotropic normal distribution, where we set $\sigma {=} 10$.
In this case, the encoding function $\gamma:\mathbb{R}^{3}\to\mathbb{R}^{2}$ applied to Cartesian point $\mathbf{p}$ is defined as $\gamma_k(\mathbf{p}) \;=\; [ \sin ( 2 \pi \mathbf{v}_k \cdot \mathbf{p} ), \, \cos ( 2 \pi \mathbf{v}_k \cdot \mathbf{p} ) ]^\textbf{T}$ for $k = 1, \dots, L$.
These Gaussian \gls{rff} can likewise be learned either by directly optimizing frequencies or with \gls{spe}.
Table~\ref{tab:abl:jitter_rffs} (\textbf{bottom}) evaluates different combinations of log-spaced and Gaussian \gls{rff} with different methods for learning frequencies.
We find that simply using a fixed encoding with log-spaced frequencies works best.

\section{Conclusion}
Neural networks are biased towards learning low-frequency functions of their inputs.
In point cloud \gls{il}, this results in policies that ignore the high-frequency information that is essential for high-precision manipulation, such as insertion tasks or grasping.
We incorporate the well-known Fourier feature mapping introduced in NeRF~\citep{mildenhall2021nerf} into a variety of point cloud-based Imitation Learning methods and test them on high-precision manipulation tasks in simulation and on a real robot.

We demonstrate that simply encoding the policy's coordinate inputs via Fourier features provides significant and consistent performance benefits.
These benefits hold across RoboCasa and ManiSkill3 tasks of varying difficulty and are consistent across different point cloud encoders.
Furthermore, performance also improves in multimodal architectures and are robust to real world noise and camera artifacts.
Our analysis confirms that Fourier features are most helpful when point clouds contain rich geometric information, and seem to improve learning dynamics even in the absence of fine geometry.
Parameter studies further show that Fourier features are robust to most hyperparameters, making them easy to use at essentially no additional cost.
We thus argue that Fourier features should be used with practically any point cloud encoder architecture rather than Cartesian point features.
Future work may investigate gradient-based learning of the optimal wavelengths or additional regularization to improve scalability.

\newpage
\section*{Acknowledgments}

The present contribution is supported by the Helmholtz Association under the joint research school “HIDSS4Health – Helmholtz Information and Data Science School for Health.
This work was supported by the European Research Council (ERC) under the European Union’s Horizon Europe programme through the project SMARTI\textsuperscript{3} (Grant Agreement No. 101171393).
The authors gratefully acknowledge the computing time provided on the high-performance computer HoreKa by the National High-Performance Computing Center at KIT (NHR@KIT). This center is jointly supported by the Federal Ministry of Education and Research and the Ministry of Science, Research and the Arts of Baden-Württemberg, as part of the National High-Performance Computing (NHR) joint funding program. HoreKa is partly funded by the German Research Foundation (DFG).
This work was supported by the Helmholtz Association's Initiative and Networking Fund on the HAICORE@KIT partition.

\section*{Impact Statement}
This paper presents work whose goal is to advance the field of machine learning.
There are many potential societal consequences of our work, none of which we feel must be specifically highlighted here.

\bibliography{bibliography}

\begin{thebibliography}{62}
\providecommand{\natexlab}[1]{#1}
\providecommand{\url}[1]{\texttt{#1}}
\expandafter\ifx\csname urlstyle\endcsname\relax
  \providecommand{\doi}[1]{doi: #1}\else
  \providecommand{\doi}{doi: \begingroup \urlstyle{rm}\Url}\fi

\bibitem[Abello et~al.(2021)Abello, Hirata, and Wang]{abello2021dissecting}
Abello, A.~A., Hirata, R., and Wang, Z.
\newblock Dissecting the high-frequency bias in convolutional neural networks.
\newblock In \emph{Proceedings of the IEEE/CVF Conference on Computer Vision and Pattern Recognition}, pp.\  863--871, 2021.

\bibitem[Agarwal et~al.(2021)Agarwal, Schwarzer, Castro, Courville, and Bellemare]{agarwal2021deep}
Agarwal, R., Schwarzer, M., Castro, P.~S., Courville, A., and Bellemare, M.~G.
\newblock Deep reinforcement learning at the edge of the statistical precipice.
\newblock \emph{Advances in Neural Information Processing Systems}, 2021.

\bibitem[Barron et~al.(2022)Barron, Mildenhall, Verbin, Srinivasan, and Hedman]{barron2022mip}
Barron, J.~T., Mildenhall, B., Verbin, D., Srinivasan, P.~P., and Hedman, P.
\newblock Mip-nerf 360: Unbounded anti-aliased neural radiance fields.
\newblock In \emph{Proceedings of the IEEE/CVF conference on computer vision and pattern recognition}, pp.\  5470--5479, 2022.

\bibitem[Black et~al.(2024)Black, Brown, Driess, Esmail, Equi, Finn, Fusai, Groom, Hausman, Ichter, et~al.]{black2024pi_0}
Black, K., Brown, N., Driess, D., Esmail, A., Equi, M., Finn, C., Fusai, N., Groom, L., Hausman, K., Ichter, B., et~al.
\newblock $\pi\_0 $: A vision-language-action flow model for general robot control.
\newblock \emph{arXiv preprint arXiv:2410.24164}, 2024.

\bibitem[Brohan et~al.(2022)Brohan, Brown, Carbajal, Chebotar, Dabis, Finn, Gopalakrishnan, Hausman, Herzog, Hsu, et~al.]{brohan2022rt}
Brohan, A., Brown, N., Carbajal, J., Chebotar, Y., Dabis, J., Finn, C., Gopalakrishnan, K., Hausman, K., Herzog, A., Hsu, J., et~al.
\newblock Rt-1: Robotics transformer for real-world control at scale.
\newblock \emph{arXiv preprint arXiv:2212.06817}, 2022.

\bibitem[Chen et~al.(2023)Chen, Wang, Yang, Yu, Yuan, and Yue]{chen2023pointgpt}
Chen, G., Wang, M., Yang, Y., Yu, K., Yuan, L., and Yue, Y.
\newblock Pointgpt: Auto-regressively generative pre-training from point clouds.
\newblock \emph{Advances in Neural Information Processing Systems}, 36:\penalty0 29667--29679, 2023.

\bibitem[Chen et~al.(2024)Chen, Garcia, Laptev, and Schmid]{sugar}
Chen, S., Garcia, R., Laptev, I., and Schmid, C.
\newblock Sugar: Pre-training 3d visual representations for robotics.
\newblock In \emph{Proceedings of the IEEE/CVF Conference on Computer Vision and Pattern Recognition (CVPR)}, pp.\  18049--18060, June 2024.

\bibitem[Chi et~al.(2023)Chi, Feng, Du, Xu, Cousineau, Burchfiel, and Song]{chi2023diffusionpolicy}
Chi, C., Feng, S., Du, Y., Xu, Z., Cousineau, E., Burchfiel, B., and Song, S.
\newblock Diffusion policy: Visuomotor policy learning via action diffusion.
\newblock In \emph{Proceedings of Robotics: Science and Systems (RSS)}, 2023.

\bibitem[Chung(1997)]{chung1997spectral}
Chung, F.~R.
\newblock \emph{Spectral Graph Theory}, volume~92.
\newblock American Mathematical Soc., 1997.

\bibitem[Donat et~al.(2025)Donat, Jia, Huang, Taranovic, Blessing, Li, Zhou, Zhang, Lioutikov, and Neumann]{donat2025towards}
Donat, A., Jia, X., Huang, X., Taranovic, A., Blessing, D., Li, G., Zhou, H., Zhang, H., Lioutikov, R., and Neumann, G.
\newblock Towards fusing point cloud and visual representations for imitation learning.
\newblock In \emph{7th Robot Learning Workshop: Towards Robots with Human-Level Abilities}, 2025.
\newblock URL \url{https://openreview.net/forum?id=5cG7ilWX1V}.

\bibitem[Gao et~al.(2023)Gao, Dai, and Zhang]{gao2023adaptive}
Gao, Z., Dai, W., and Zhang, Y.
\newblock Adaptive positional encoding for bundle-adjusting neural radiance fields.
\newblock In \emph{Proceedings of the IEEE/CVF International Conference on Computer Vision}, pp.\  3284--3294, 2023.

\bibitem[Gervet et~al.(2023)Gervet, Xian, Gkanatsios, and Fragkiadaki]{gervet2023act3d}
Gervet, T., Xian, Z., Gkanatsios, N., and Fragkiadaki, K.
\newblock Act3d: 3d feature field transformers for multi-task robotic manipulation.
\newblock \emph{arXiv preprint arXiv:2306.17817}, 2023.

\bibitem[Goyal et~al.(2023)Goyal, Xu, Guo, Blukis, Chao, and Fox]{goyal2023rvt}
Goyal, A., Xu, J., Guo, Y., Blukis, V., Chao, Y.-W., and Fox, D.
\newblock Rvt: Robotic view transformer for 3d object manipulation.
\newblock In \emph{Conference on Robot Learning}, pp.\  694--710. PMLR, 2023.

\bibitem[Gyenes et~al.(2024)Gyenes, Franke, Becker, and Neumann]{gyenes2024pointpatchrl}
Gyenes, B., Franke, N., Becker, P., and Neumann, G.
\newblock Pointpatch{RL} - masked reconstruction improves reinforcement learning on point clouds.
\newblock In \emph{8th Annual Conference on Robot Learning}, 2024.
\newblock URL \url{https://openreview.net/forum?id=3jNEz3kUSl}.

\bibitem[Gyenes et~al.(2025)Gyenes, Franke, Scheikl, Henrich, Younis, Neumann, Wagner, and Mathis-Ullrich]{kirurc}
Gyenes, B., Franke, N., Scheikl, P.~M., Henrich, P., Younis, R., Neumann, G., Wagner, M., and Mathis-Ullrich, F.
\newblock Point cloud segmentation for autonomous clip positioning in laparoscopic cholecystectomy on a phantom.
\newblock \emph{IEEE Robotics and Automation Letters}, 10\penalty0 (8):\penalty0 8522--8529, 2025.
\newblock \doi{10.1109/LRA.2025.3585357}.

\bibitem[He et~al.(2015)He, Zhang, Ren, and Sun]{he2015deepresiduallearningimage}
He, K., Zhang, X., Ren, S., and Sun, J.
\newblock Deep residual learning for image recognition, 2015.
\newblock URL \url{https://arxiv.org/abs/1512.03385}.

\bibitem[Ho et~al.(2020)Ho, Jain, and Abbeel]{ho2020denoising}
Ho, J., Jain, A., and Abbeel, P.
\newblock Denoising diffusion probabilistic models.
\newblock \emph{Advances in neural information processing systems}, 33:\penalty0 6840--6851, 2020.

\bibitem[Hornik et~al.(1989)Hornik, Stinchcombe, and White]{HORNIK1989359}
Hornik, K., Stinchcombe, M., and White, H.
\newblock Multilayer feedforward networks are universal approximators.
\newblock \emph{Neural Networks}, 2\penalty0 (5):\penalty0 359--366, 1989.
\newblock ISSN 0893-6080.
\newblock \doi{https://doi.org/10.1016/0893-6080(89)90020-8}.
\newblock URL \url{https://www.sciencedirect.com/science/article/pii/0893608089900208}.

\bibitem[Intelligence et~al.(2025)Intelligence, Black, Brown, Darpinian, Dhabalia, Driess, Esmail, Equi, Finn, Fusai, Galliker, Ghosh, Groom, Hausman, Ichter, Jakubczak, Jones, Ke, LeBlanc, Levine, Li-Bell, Mothukuri, Nair, Pertsch, Ren, Shi, Smith, Springenberg, Stachowicz, Tanner, Vuong, Walke, Walling, Wang, Yu, and Zhilinsky]{pi_0.5}
Intelligence, P., Black, K., Brown, N., Darpinian, J., Dhabalia, K., Driess, D., Esmail, A., Equi, M., Finn, C., Fusai, N., Galliker, M.~Y., Ghosh, D., Groom, L., Hausman, K., Ichter, B., Jakubczak, S., Jones, T., Ke, L., LeBlanc, D., Levine, S., Li-Bell, A., Mothukuri, M., Nair, S., Pertsch, K., Ren, A.~Z., Shi, L.~X., Smith, L., Springenberg, J.~T., Stachowicz, K., Tanner, J., Vuong, Q., Walke, H., Walling, A., Wang, H., Yu, L., and Zhilinsky, U.
\newblock $\pi_{0.5}$: a vision-language-action model with open-world generalization, 2025.
\newblock URL \url{https://arxiv.org/abs/2504.16054}.

\bibitem[Jia et~al.(2025{\natexlab{a}})Jia, Wang, Wang, Wang, Gyenes, Gospodinov, Jiang, Li, Zhou, Liao, Huang, Beck, Reuss, Lioutikov, and Neumann]{pointmappolicy}
Jia, X., Wang, Q., Wang, A., Wang, H., Gyenes, B., Gospodinov, E., Jiang, X., Li, G., Zhou, H., Liao, W., Huang, X., Beck, M., Reuss, M., Lioutikov, R., and Neumann, G.
\newblock Pointmappolicy: Structured point cloud processing for multi-modal imitation learning.
\newblock In \emph{The Thirty-ninth Annual Conference on Neural Information Processing Systems}, 2025{\natexlab{a}}.
\newblock URL \url{https://openreview.net/forum?id=ZR2mdBrhJX}.

\bibitem[Jia et~al.(2025{\natexlab{b}})Jia, Liu, Chen, Gu, Wang, Luo, Li, Wang, Wang, Zhang, and Zhang]{lift3d}
Jia, Y., Liu, J., Chen, S., Gu, C., Wang, Z., Luo, L., Li, X., Wang, P., Wang, Z., Zhang, R., and Zhang, S.
\newblock Lift3d policy: Lifting 2d foundation models for robust 3d robotic manipulation.
\newblock In \emph{Proceedings of the IEEE/CVF Conference on Computer Vision and Pattern Recognition (CVPR)}, pp.\  17347--17358, June 2025{\natexlab{b}}.

\bibitem[Jo \& Bengio(2017)Jo and Bengio]{jo2017measuring}
Jo, J. and Bengio, Y.
\newblock Measuring the tendency of cnns to learn surface statistical regularities.
\newblock \emph{arXiv preprint arXiv:1711.11561}, 2017.

\bibitem[Karras et~al.(2022)Karras, Aittala, Aila, and Laine]{karras2022elucidating}
Karras, T., Aittala, M., Aila, T., and Laine, S.
\newblock Elucidating the design space of diffusion-based generative models.
\newblock In Oh, A.~H., Agarwal, A., Belgrave, D., and Cho, K. (eds.), \emph{Advances in Neural Information Processing Systems}, 2022.
\newblock URL \url{https://openreview.net/forum?id=k7FuTOWMOc7}.

\bibitem[Ke et~al.(2025)Ke, Gkanatsios, and Fragkiadaki]{ke20243d}
Ke, T.-W., Gkanatsios, N., and Fragkiadaki, K.
\newblock 3d diffuser actor: Policy diffusion with 3d scene representations.
\newblock In Agrawal, P., Kroemer, O., and Burgard, W. (eds.), \emph{Proceedings of The 8th Conference on Robot Learning}, volume 270 of \emph{Proceedings of Machine Learning Research}, pp.\  1949--1974. PMLR, 06--09 Nov 2025.
\newblock URL \url{https://proceedings.mlr.press/v270/ke25a.html}.

\bibitem[Lai et~al.(2022)Lai, Liu, Jiang, Wang, Zhao, Liu, Qi, and Jia]{stratified_transformer}
Lai, X., Liu, J., Jiang, L., Wang, L., Zhao, H., Liu, S., Qi, X., and Jia, J.
\newblock Stratified transformer for 3d point cloud segmentation.
\newblock In \emph{Proceedings of the IEEE/CVF Conference on Computer Vision and Pattern Recognition (CVPR)}, pp.\  8500--8509, June 2022.

\bibitem[Li et~al.(2026)Li, Wen, Peng, Peng, and Zhu]{pointvla}
Li, C., Wen, J., Peng, Y., Peng, Y., and Zhu, Y.
\newblock Pointvla: Injecting the 3d world into vision-language-action models.
\newblock \emph{IEEE Robotics and Automation Letters}, 11\penalty0 (3):\penalty0 2506--2513, 2026.
\newblock \doi{10.1109/LRA.2026.3653303}.

\bibitem[Lipman et~al.(2023)Lipman, Chen, Ben-Hamu, Nickel, and Le]{lipman2023flow}
Lipman, Y., Chen, R. T.~Q., Ben-Hamu, H., Nickel, M., and Le, M.
\newblock Flow matching for generative modeling.
\newblock In \emph{The Eleventh International Conference on Learning Representations}, 2023.
\newblock URL \url{https://openreview.net/forum?id=PqvMRDCJT9t}.

\bibitem[Lippe et~al.(2023)Lippe, Veeling, Perdikaris, Turner, and Brandstetter]{lippe2023pde}
Lippe, P., Veeling, B., Perdikaris, P., Turner, R., and Brandstetter, J.
\newblock Pde-refiner: Achieving accurate long rollouts with neural pde solvers.
\newblock \emph{Advances in Neural Information Processing Systems}, 36:\penalty0 67398--67433, 2023.

\bibitem[Miao et~al.(2024)Miao, Dong, Zhang, Yu, Yang, and Gao]{miao2024improving}
Miao, Y., Dong, Y., Zhang, J., Yu, L., Yang, X., and Gao, X.-S.
\newblock Improving robustness of 3d point cloud recognition from a fourier perspective.
\newblock \emph{Advances in Neural Information Processing Systems}, 37:\penalty0 68183--68210, 2024.

\bibitem[Mildenhall et~al.(2021)Mildenhall, Srinivasan, Tancik, Barron, Ramamoorthi, and Ng]{mildenhall2021nerf}
Mildenhall, B., Srinivasan, P.~P., Tancik, M., Barron, J.~T., Ramamoorthi, R., and Ng, R.
\newblock Nerf: Representing scenes as neural radiance fields for view synthesis.
\newblock \emph{Communications of the ACM}, 65\penalty0 (1):\penalty0 99--106, 2021.

\bibitem[Nasiriany et~al.(2024)Nasiriany, Maddukuri, Zhang, Parikh, Lo, Joshi, Mandlekar, and Zhu]{robocasa2024}
Nasiriany, S., Maddukuri, A., Zhang, L., Parikh, A., Lo, A., Joshi, A., Mandlekar, A., and Zhu, Y.
\newblock Robocasa: Large-scale simulation of everyday tasks for generalist robots.
\newblock In \emph{Robotics: Science and Systems (RSS)}, 2024.

\bibitem[Pang et~al.(2022)Pang, Wang, Tay, Liu, Tian, and Yuan]{pang2022masked}
Pang, Y., Wang, W., Tay, F.~E., Liu, W., Tian, Y., and Yuan, L.
\newblock Masked autoencoders for point cloud self-supervised learning.
\newblock In \emph{European Conference on Computer Vision}, pp.\  604--621. Springer, 2022.

\bibitem[Qi et~al.(2017{\natexlab{a}})Qi, Su, Mo, and Guibas]{qi2017pointnet}
Qi, C.~R., Su, H., Mo, K., and Guibas, L.~J.
\newblock Pointnet: Deep learning on point sets for 3d classification and segmentation.
\newblock In \emph{Proceedings of the IEEE conference on computer vision and pattern recognition}, pp.\  652--660, 2017{\natexlab{a}}.

\bibitem[Qi et~al.(2017{\natexlab{b}})Qi, Yi, Su, and Guibas]{qi2017pointnetplusplus}
Qi, C.~R., Yi, L., Su, H., and Guibas, L.~J.
\newblock Pointnet++: Deep hierarchical feature learning on point sets in a metric space.
\newblock \emph{arXiv preprint arXiv:1706.02413}, 2017{\natexlab{b}}.

\bibitem[Qian et~al.(2022)Qian, Li, Peng, Mai, Hammoud, Elhoseiny, and Ghanem]{qian2022pointnext}
Qian, G., Li, Y., Peng, H., Mai, J., Hammoud, H., Elhoseiny, M., and Ghanem, B.
\newblock Pointnext: Revisiting pointnet++ with improved training and scaling strategies.
\newblock \emph{Advances in neural information processing systems}, 35:\penalty0 23192--23204, 2022.

\bibitem[Radford et~al.(2021)Radford, Kim, Hallacy, Ramesh, Goh, Agarwal, Sastry, Askell, Mishkin, Clark, et~al.]{radford2021learning}
Radford, A., Kim, J.~W., Hallacy, C., Ramesh, A., Goh, G., Agarwal, S., Sastry, G., Askell, A., Mishkin, P., Clark, J., et~al.
\newblock Learning transferable visual models from natural language supervision.
\newblock In \emph{International conference on machine learning}, pp.\  8748--8763. PmLR, 2021.

\bibitem[Rahaman et~al.(2019)Rahaman, Baratin, Arpit, Draxler, Lin, Hamprecht, Bengio, and Courville]{pmlr-v97-rahaman19a}
Rahaman, N., Baratin, A., Arpit, D., Draxler, F., Lin, M., Hamprecht, F., Bengio, Y., and Courville, A.
\newblock On the spectral bias of neural networks.
\newblock In Chaudhuri, K. and Salakhutdinov, R. (eds.), \emph{Proceedings of the 36th International Conference on Machine Learning}, volume~97 of \emph{Proceedings of Machine Learning Research}, pp.\  5301--5310. PMLR, 09--15 Jun 2019.
\newblock URL \url{https://proceedings.mlr.press/v97/rahaman19a.html}.

\bibitem[Reuss et~al.(2023)Reuss, Li, Jia, and Lioutikov]{reuss2023goal}
Reuss, M., Li, M., Jia, X., and Lioutikov, R.
\newblock Goal conditioned imitation learning using score-based diffusion policies.
\newblock In \emph{Proceedings of Robotics: Science and Systems (RSS)}, 2023.

\bibitem[Scarselli et~al.(2009)Scarselli, Gori, Tsoi, Hagenbuchner, and Monfardini]{gnn}
Scarselli, F., Gori, M., Tsoi, A.~C., Hagenbuchner, M., and Monfardini, G.
\newblock The graph neural network model.
\newblock \emph{IEEE Transactions on Neural Networks}, 20\penalty0 (1):\penalty0 61--80, 2009.
\newblock \doi{10.1109/TNN.2008.2005605}.

\bibitem[Song et~al.(2021)Song, Meng, and Ermon]{song2021denoising}
Song, J., Meng, C., and Ermon, S.
\newblock Denoising diffusion implicit models.
\newblock In \emph{ICLR}, 2021.

\bibitem[Song \& Dhariwal(2024)Song and Dhariwal]{song2024improved}
Song, Y. and Dhariwal, P.
\newblock Improved techniques for training consistency models.
\newblock In \emph{The Twelfth International Conference on Learning Representations}, 2024.
\newblock URL \url{https://openreview.net/forum?id=WNzy9bRDvG}.

\bibitem[Sun et~al.(2024)Sun, Yuan, Xu, Mai, Siddharth, Chen, and Marina]{sun2024learning}
Sun, C., Yuan, Z., Xu, K., Mai, L., Siddharth, N., Chen, S., and Marina, M.~K.
\newblock Learning high-frequency functions made easy with sinusoidal positional encoding.
\newblock \emph{arXiv preprint arXiv:2407.09370}, 2024.

\bibitem[Tancik et~al.(2020)Tancik, Srinivasan, Mildenhall, Fridovich-Keil, Raghavan, Singhal, Ramamoorthi, Barron, and Ng]{fourier_features}
Tancik, M., Srinivasan, P.~P., Mildenhall, B., Fridovich-Keil, S., Raghavan, N., Singhal, U., Ramamoorthi, R., Barron, J.~T., and Ng, R.
\newblock Fourier features let networks learn high frequency functions in low dimensional domains.
\newblock In \emph{Proceedings of the 34th International Conference on Neural Information Processing Systems}, NIPS '20, Red Hook, NY, USA, 2020. Curran Associates Inc.
\newblock ISBN 9781713829546.

\bibitem[Tao et~al.(2025)Tao, Xiang, Shukla, Qin, Hinrichsen, Yuan, Bao, Lin, Liu, Chan, Gao, Li, Mu, Xiao, Gurha, N, Choi, Chen, Huang, Calandra, Chen, Luo, and Su]{tao2025maniskill}
Tao, S., Xiang, F., Shukla, A., Qin, Y., Hinrichsen, X., Yuan, X., Bao, C., Lin, X., Liu, Y., Chan, T.-K., Gao, Y., Li, X., Mu, T., Xiao, N., Gurha, A., N, V., Choi, Y.~W., Chen, Y.-R., Huang, Z., Calandra, R., Chen, R., Luo, S., and Su, H.
\newblock Maniskill3: {GPU} parallelized robot simulation and rendering for generalizable embodied {AI}.
\newblock In \emph{7th Robot Learning Workshop: Towards Robots with Human-Level Abilities}, 2025.
\newblock URL \url{https://openreview.net/forum?id=GgTxudXaU8}.

\bibitem[Vincent(2011)]{6795935}
Vincent, P.
\newblock A connection between score matching and denoising autoencoders.
\newblock \emph{Neural Computation}, 23\penalty0 (7):\penalty0 1661--1674, 2011.
\newblock \doi{10.1162/NECO_a_00142}.

\bibitem[Von~Luxburg(2007)]{von2007tutorial}
Von~Luxburg, U.
\newblock A tutorial on spectral clustering.
\newblock \emph{Statistics and computing}, 17\penalty0 (4):\penalty0 395--416, 2007.

\bibitem[Wang et~al.(2020)Wang, Wu, Huang, and Xing]{wang2020high}
Wang, H., Wu, X., Huang, Z., and Xing, E.~P.
\newblock High-frequency component helps explain the generalization of convolutional neural networks.
\newblock In \emph{Proceedings of the IEEE/CVF conference on computer vision and pattern recognition}, pp.\  8684--8694, 2020.

\bibitem[Wang et~al.(2024)Wang, Leroy, Cabon, Chidlovskii, and Revaud]{dust3r}
Wang, S., Leroy, V., Cabon, Y., Chidlovskii, B., and Revaud, J.
\newblock Dust3r: Geometric 3d vision made easy.
\newblock In \emph{Proceedings of the IEEE/CVF Conference on Computer Vision and Pattern Recognition}, pp.\  20697--20709, 2024.

\bibitem[Wilcox et~al.(2025)Wilcox, Ghanem, Moghani, Barroso, Joffe, and Garg]{adapt3r}
Wilcox, A., Ghanem, M., Moghani, M., Barroso, P., Joffe, B., and Garg, A.
\newblock Adapt3r: Adaptive 3d scene representation for domain transfer in imitation learning.
\newblock \emph{CoRR}, abs/2503.04877, March 2025.
\newblock URL \url{https://doi.org/10.48550/arXiv.2503.04877}.

\bibitem[Woo et~al.(2023)Woo, Debnath, Hu, Chen, Liu, Kweon, and Xie]{Woo2023ConvNeXtV2}
Woo, S., Debnath, S., Hu, R., Chen, X., Liu, Z., Kweon, I.~S., and Xie, S.
\newblock Convnext v2: Co-designing and scaling convnets with masked autoencoders.
\newblock \emph{arXiv preprint arXiv:2301.00808}, 2023.

\bibitem[Wu et~al.(2025)Wu, Chen, Swamy, Brantley, and Sun]{wu2025diffusing}
Wu, R., Chen, Y., Swamy, G., Brantley, K., and Sun, W.
\newblock Diffusing states and matching scores: A new framework for imitation learning.
\newblock In \emph{The Thirteenth International Conference on Learning Representations}, 2025.
\newblock URL \url{https://openreview.net/forum?id=kWRKNDU6uN}.

\bibitem[W{\"u}rth et~al.(2026)W{\"u}rth, Freymuth, Neumann, and K{\"a}rger]{wurth2026diffusion}
W{\"u}rth, T., Freymuth, N., Neumann, G., and K{\"a}rger, L.
\newblock Diffusion-based hierarchical graph neural networks for simulating nonlinear solid mechanics.
\newblock \emph{Advances in Neural Information Processing Systems}, 39, 2026.

\bibitem[Yu et~al.(2022)Yu, Tang, Rao, Huang, Zhou, and Lu]{pointBERT}
Yu, X., Tang, L., Rao, Y., Huang, T., Zhou, J., and Lu, J.
\newblock Point-bert: Pre-training 3d point cloud transformers with masked point modeling.
\newblock In \emph{Proceedings of the IEEE/CVF conference on computer vision and pattern recognition}, pp.\  19313--19322, 2022.

\bibitem[Ze et~al.(2024)Ze, Zhang, Zhang, Hu, Wang, and Xu]{Ze2024DP3}
Ze, Y., Zhang, G., Zhang, K., Hu, C., Wang, M., and Xu, H.
\newblock 3d diffusion policy: Generalizable visuomotor policy learning via simple 3d representations.
\newblock In \emph{Proceedings of Robotics: Science and Systems (RSS)}, 2024.

\bibitem[Ze et~al.(2025)Ze, Chen, Wang, Chen, He, Yuan, Peng, and Wu]{idp3}
Ze, Y., Chen, Z., Wang, W., Chen, T., He, X., Yuan, Y., Peng, X.~B., and Wu, J.
\newblock Generalizable humanoid manipulation with 3d diffusion policies, 2025.
\newblock URL \url{https://arxiv.org/abs/2410.10803}.

\bibitem[Zelnik-Manor \& Perona(2004)Zelnik-Manor and Perona]{zelnik2004self}
Zelnik-Manor, L. and Perona, P.
\newblock Self-tuning spectral clustering.
\newblock \emph{Advances in neural information processing systems}, 17, 2004.

\bibitem[Zhao et~al.(2021)Zhao, Jiang, Jia, Torr, and Koltun]{point_transformer}
Zhao, H., Jiang, L., Jia, J., Torr, P.~H., and Koltun, V.
\newblock Point transformer.
\newblock In \emph{Proceedings of the IEEE/CVF international conference on computer vision}, pp.\  16259--16268, 2021.

\bibitem[Zhao et~al.(2023)Zhao, Kumar, Levine, and Finn]{zhao2023learning}
Zhao, T.~Z., Kumar, V., Levine, S., and Finn, C.
\newblock {Learning Fine-Grained Bimanual Manipulation with Low-Cost Hardware}.
\newblock In \emph{Proceedings of Robotics: Science and Systems}, Daegu, Republic of Korea, July 2023.
\newblock \doi{10.15607/RSS.2023.XIX.016}.

\bibitem[Zhou et~al.(2024)Zhou, Wang, Ma, Liu, Huang, and Wang]{zhou2024uni3d}
Zhou, J., Wang, J., Ma, B., Liu, Y.-S., Huang, T., and Wang, X.
\newblock Uni3d: Exploring unified 3d representation at scale.
\newblock In \emph{The Twelfth International Conference on Learning Representations}, 2024.
\newblock URL \url{https://openreview.net/forum?id=wcaE4Dfgt8}.

\bibitem[Zhu et~al.(2024)Zhu, Wang, Huang, Ye, Ouyang, and He]{zhu2024point}
Zhu, H., Wang, Y., Huang, D., Ye, W., Ouyang, W., and He, T.
\newblock Point cloud matters: Rethinking the impact of different observation spaces on robot learning.
\newblock In \emph{The Thirty-eight Conference on Neural Information Processing Systems Datasets and Benchmarks Track}, 2024.
\newblock URL \url{https://openreview.net/forum?id=zgSnSZ0Re6}.

\bibitem[Zhu et~al.(2025)Zhu, Zhu, Li, Wen, Xu, Liu, Cheng, Shen, Peng, Feng, et~al.]{zhu2025scaling}
Zhu, M., Zhu, Y., Li, J., Wen, J., Xu, Z., Liu, N., Cheng, R., Shen, C., Peng, Y., Feng, F., et~al.
\newblock Scaling diffusion policy in transformer to 1 billion parameters for robotic manipulation.
\newblock In \emph{2025 IEEE International Conference on Robotics and Automation (ICRA)}, pp.\  10838--10845. IEEE, 2025.

\bibitem[Zitkovich et~al.(2023)Zitkovich, Yu, Xu, Xu, Xiao, Xia, Wu, Wohlhart, Welker, Wahid, et~al.]{zitkovich2023rt}
Zitkovich, B., Yu, T., Xu, S., Xu, P., Xiao, T., Xia, F., Wu, J., Wohlhart, P., Welker, S., Wahid, A., et~al.
\newblock Rt-2: Vision-language-action models transfer web knowledge to robotic control.
\newblock In \emph{Conference on Robot Learning}, pp.\  2165--2183. PMLR, 2023.

\end{thebibliography}
\bibliographystyle{icml2026}

\newpage
\appendix
\onecolumn

\section{Appendix}

\subsection{Tasks}
\label{app:tasks}

\begin{figure}[htb!]
    \centering
    \includegraphics[width=0.9\linewidth]{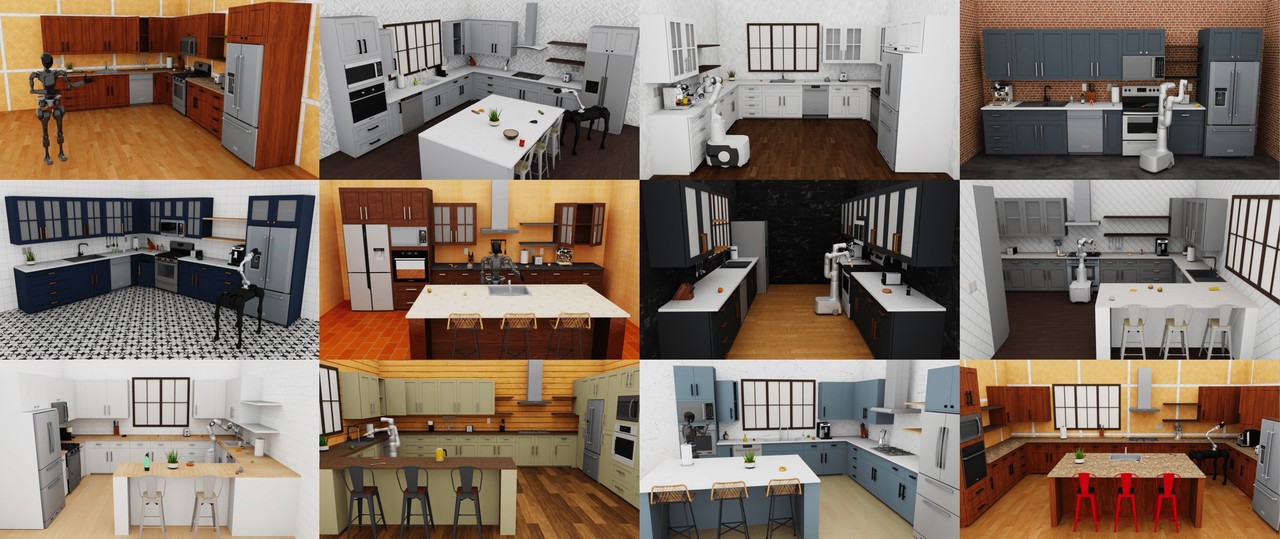}
    \caption{\textbf{Overview of RoboCasa Simulation Environments.} Example kitchen scenes and tasks illustrating the diversity of household manipulation settings provided by RoboCasa.}
    \label{fig:robocasa_benchmark}
\end{figure}

\begin{table}[htb!]
    \centering

\resizebox{0.8\linewidth}{!}{%
\renewcommand{\arraystretch}{1.3}
\setlength{\aboverulesep}{0pt}
\setlength{\belowrulesep}{0pt}
\begin{tabular}{lll}
    \toprule
    \textbf{Category} & \textbf{Task} & \textbf{Description} \\
    \midrule

    \multirow{2}{*}{Insertion}
    & CoffeeServeMug
    & Remove the mug from the holder and place it on the counter.
    \\
    & CoffeeSetupMug
    & Place the mug into the coffee machine’s mug holder.
    \\
    \midrule
    \multirow{3}{*}{Pressing Buttons}
    & CoffeePressButton
    & Press the button to pour coffee into the mug.
    \\
    & TurnOnMicrowave
    & Start the microwave by pressing the start button.
    \\
    & TurnOffMicrowave
    & Stop the microwave by pressing the stop button.
    \\
    \midrule
    \multirow{3}{*}{Turning Levers}
    & TurnOnSinkFaucet
    & Turn on the sink faucet to start water flow.
    \\
    & TurnOffSinkFaucet
    & Turn off the sink faucet to stop water flow.
    \\
    & TurnSinkSpout
    & Rotate the sink spout.
    \\
    \midrule
    \multirow{2}{*}{Twisting Knobs}
    & TurnOnStove
    & Turn on a specific stove burner by twisting its knob.
    \\
    & TurnOffStove
    & Turn off a specific stove burner by twisting its knob.
    \\
    \midrule
    \multirow{2}{*}{Open/Close Drawers}
    & OpenDrawer
    & Open a drawer.
    \\
    & CloseDrawer
    & Close a drawer.
    \\
    \midrule
    \multirow{4}{*}{Opening and Closing Doors}
    
    & OpenSingleDoor
    & Open a microwave door or a cabinet with a single door.
    \\
    & CloseSingleDoor
    & Close a microwave door or a cabinet with a single door.
    \\
    & OpenDoubleDoor
    & Open a cabinet with two opposite-facing doors.
    \\
    & CloseDoubleDoor
    & Close a cabinet with two opposite-facing doors.
    \\
    \bottomrule
    
\end{tabular}
}

    \caption{RoboCasa evaluation tasks.}
    \label{tab:robocasa_tasks}
\end{table}

\textbf{RoboCasa.}
RoboCasa~\citep{robocasa2024} is a large-scale simulation benchmark designed for training generalist robots in realistic household settings, with an emphasis on kitchen environments.
It provides 100 tasks in total: 25 atomic tasks with 50 human demonstrations each, and 75 composite tasks with automatically generated demonstrations.
The task set covers eight fundamental skills that are essential for home robotics: (1) pick-and-place, (2) door opening and closing, (3) drawer opening and closing, (4) knob turning, (5) lever manipulation, (6) button pressing, (7) insertion, and (8) navigation.
The joint action space is 7-dimensional, including end-effector translation, rotation, and gripper control.
For our experiments, we exclude the $8$ Pick and Place tasks because they are too difficult, and the NavigateKitchen task because it has an incompatible action space, making multi-task training impractical.
Table \ref{tab:robocasa_tasks} lists each task along with the goal text used for it.
Note that TurnSinkSpout, TurnOnStove, and TurnOffStove have dynamic goals that vary on each episode.

\clearpage

\begin{figure}[htb!]
    \centering
    \includegraphics[width=0.9\linewidth]{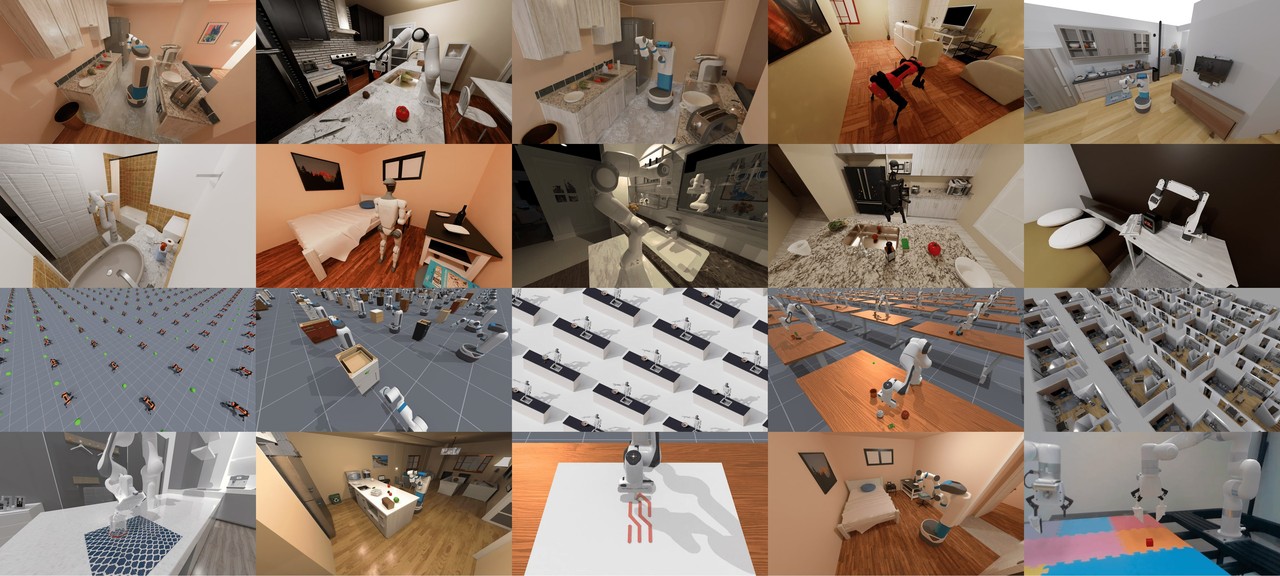}
    \caption{\textbf{Overview of ManiSkill3 Simulation Environments.} Example object-centric manipulation tasks illustrating the diversity of interactions supported by ManiSkill3.}
    \label{fig:maniskill_benchmark}
\end{figure}

\begin{table}[ht]
    \centering
    \caption{ManiSkill3 evaluation tasks.}
    \label{tab:maniskill3_tasks}
    \resizebox{0.8\linewidth}{!}{%
\renewcommand{\arraystretch}{1.3}
\setlength{\aboverulesep}{0pt}
\setlength{\belowrulesep}{0pt}
\begin{tabular}{lll}
    \toprule
    \textbf{Category} & \textbf{Task} & \textbf{Description} \\
    \midrule

    \multirow{4}{*}{Table-Top 2 Finger Gripper}
    & PullCube-v1
    & Pick up the cube and pull it to the target.
    \\
    & PushCube-v1
    & Push the cube into the target.
    \\
    & PokeCube-v1
    & Use the tool to poke the cube until it reaches the target.
    \\
    & RollBall-v1
    & Push the ball to make it roll into the target.
    \\
    \bottomrule
\end{tabular}
}

\end{table}

\textbf{ManiSkill3.}
ManiSkill3~\citep{tao2025maniskill} is a large-scale GPU-parallelized simulation benchmark designed for scalable training of embodied agents. 
It offers diverse object-centric manipulation tasks such as grasping, assembling, and tool use, with support for both imitation and reinforcement learning. 
Unlike RoboCasa, which emphasizes long-horizon household tasks in visually rich kitchen environments, ManiSkill3 provides highly parallelized simulation and rendering of physics-based interactions, enabling efficient large-scale experimentation and evaluation of manipulation policies.
A summary of all ManiSkill3 tasks can be found in Table \ref{tab:maniskill3_tasks}, each representing a distinct skill. 

\clearpage

\begin{table}[htb!]
    \centering
    \caption{Real world tasks with their goal descriptions.}
    \label{tab:tasks_real}
    \resizebox{0.5\linewidth}{!}{%
\renewcommand{\arraystretch}{1.3}
\setlength{\aboverulesep}{0pt}
\setlength{\belowrulesep}{0pt}
\begin{tabular}{lll}
    \toprule
    \textbf{Task} & \textbf{Descriptions} \\
    \midrule

    \multirow{2}{*}{Drawer}
    & put the red cube in the top drawer \\
    & put the red cube in the bottom drawer \\
    \midrule
    \multirow{2}{*}{Cup-Stacking}
    & stack the blue cup in the purple cup \\
    & stack the orange cup in the red cup \\
    & stack the red cup in the blue cup \\
    & stack the yellow cup in the orange cup \\
    \midrule
    \multirow{2}{*}{Arranging}
    & put the blue cup in the coffee machine \\
    & put the orange cup in the coffee machine \\
    & put the pink bowl in the coffee machine \\
    & put the red cup in the coffee machine \\
    \bottomrule
\end{tabular}
}

\end{table}

\textbf{Real World.}
A summary of the real world tasks along with their various goal descriptions can be found in Table \ref{tab:tasks_real}.

\clearpage

\subsection{Architectures}
\label{app:architectures}

\textbf{PointPatch.}
The PointPatch encoder~\citep{pang2022masked, pointBERT, gyenes2024pointpatchrl} divides a given point cloud into overlapping patches, tokenizes each patch and then uses a transformer for token processing.
Point features are the Cartesian coordinates relative to the patch center, and are encoded with a lightweight PointNet~\citep{qi2017pointnetplusplus} to create patch tokens.
Token position embeddings are computed by passing each centroid position through a two-layer \gls{mlp}~\citep{pang2022masked}.
We apply Fourier feature projections to the relative patch coordinates as well as the centroid positions.

\textbf{PointCloudMatters Encoder.}
We additionally evaluate the PointCloudMatters-PointNet (PCM) encoder used in \citet{zhu2024point}.
This architecture is based on the patching paradigm introduced in PointMAE~\citep{pang2022masked}, but applies a max aggregation across patches, followed by a final projection head to compute a single token for the entire point cloud.
In addition, each point is assigned its absolute Cartesian coordinates as well as its relative coordinates within the patch as features.

\textbf{DP3 Encoder.}
Unlike patch-based methods, the DP3 encoder~\citep{Ze2024DP3} creates a single token that embeds information from the entire point cloud.
Point features are passed through an \gls{mlp}, followed by a max-pooling operation to obtain order-invariant global features.
A final projection head maps the embedding to the token dimension, resulting in $\mathbf{T} \in \mathbb{R}^{1 \times D}$.
Although this architecture is simple, it is quite data efficient due to its small number of parameters.
Unlike in the authors' implementation, we do not apply \gls{fps} sampling before the encoder since this decreased performance in our experiments.

\textbf{PointTransformer Encoder.}
Inspired by the success of attention-based in methods in vision models, PointTransformer~\cite{point_transformer} applies attention in a hierarchical manner.
PointTransformer blocks augment point features using self-attention between points within a local neighborhood.
These local neighborhoods are computed with \gls{fps} and \gls{knn}.
This blocks alternate with down-sampling blocks that reduce the number of points and increase feature size.
We use the classification variety, which outputs a single token from the point cloud.
Since the node position inputs are used for \gls{fps} and \gls{knn} queries, we do not modify this input in any way.
Instead, we also pass the point positions as the node features, with a Fourier feature projection where applicable.
Because of the high compute costs of this architecture, we train each policy on 2-3 tasks instead of the typical 8.

\textbf{Pointmap Encoder.}
To compare against alternative 3D representations, we also evaluate point maps~\citep{dust3r,pointmappolicy}, which contain the same information as point clouds but are arranged in a 2D grid.
Given depth images from multiple cameras and their intrinsics and extrinsics parameters, we unproject each pixel into 3D and transform it into the world frame, resulting in a dense point map $\mathbf{X} \in \mathbb{R}^{H \times W \times 3}$ for each camera.
The resulting 3D representation can be processed directly with convolutional backbones such as ConvNeXt V2 \citep{Woo2023ConvNeXtV2} or ResNet \citep{he2015deepresiduallearningimage}.

\textbf{RGB+PointPatch Encoder.}
Due to the flexible transformer design of our denoising model, multimodal observations can be processed by concatenating tokens from parallel observation encoders.
We encode the RGB stream from each camera into a token with a ConvNeXt V2 nano (${\sim}15$M parameters)~\citep{Woo2023ConvNeXtV2} that is initialized from a pretrained checkpoint.
Point clouds are processed by the PointPatch encoder.

\clearpage

\subsection{Hyperparameters}
\label{app:hyperparameters}

\begin{table}[htb!]
\caption{Summary of select hyperparameters for simulation experiments.}
\label{tab:hyperparameters}
\centering
\begin{tabular}{lcc}
\hline
\textbf{Hyperparameter} & \textbf{ManiSkill} & \textbf{RoboCasa} \\
\hline
Training Epochs & 150 & 50 \\
Number of Attention Blocks & 4 & 4 \\
Attention Heads & 4 & 4 \\
Action Chunk Size & 10 & 20 \\
History Length & 1 & 1 \\
Embedding Dimension & 256 & 256 \\
Goal Lang Encoder & CLIP Resnet-50 & CLIP Resnet-50 \\
Attention Dropout & 0.3 & 0.3 \\
Residual Dropout & 0.1 & 0.1 \\
MLP Dropout & 0.1 & 0.1 \\
Optimizer & AdamW & AdamW \\
Betas & [0.9, 0.9] & [0.9, 0.9] \\
Learning Rate & 1e-4 & 1e-4 \\
Weight Decay & 0.05 & 0.05 \\
$\sigma_{\max}$ & 80 & 80 \\
$\sigma_{\min}$ & 0.001 & 0.001 \\
$\sigma_{t}$ & 0.5 & 0.5 \\
EMA decay & 0.995 & 0.995 \\
Time steps & Exponential & Exponential \\
Sampler & DDIM & DDIM \\
Denoising Steps & 10 & 10 \\
\hline
\end{tabular}

\end{table}

\clearpage

\begin{table*}[tb]
\centering
\caption{Average success rates on different RoboCasa tasks across $5$ seeds.
Fourier features generally lead to significant improvements for both PointPatch, DP3, and PCM architectures. 
In contrast, the convolutional PointMap struggles on these tasks, likely due to task complexity and data sparsity.
}
\label{table:robocasa_results}

\resizebox{\linewidth}{!}{%
\renewcommand{\arraystretch}{1.3}
\setlength{\aboverulesep}{0pt}
\setlength{\belowrulesep}{0pt}
\begin{tabular}{l l | >{\centering\arraybackslash}p{2.5cm} >{\centering\arraybackslash}p{2.5cm} | >{\centering\arraybackslash}p{2.5cm} >{\centering\arraybackslash}p{2.5cm} | >{\centering\arraybackslash}p{2.5cm} >{\centering\arraybackslash}p{2.5cm} | >{\centering\arraybackslash}p{2.5cm} >{\centering\arraybackslash}p{2.5cm}}
\toprule
 &  & PointPatch & + FF & PointPatch-attn & + FF & PCM & + FF & DP3 & + FF \\
Category & Task &  &  &  &  &  &  &  &  \\
\midrule
\multirow[c]{2}{*}{Insertion} & CoffeeServeMug & $1.0 \pm 2.2$ & $3.1 \pm 4.1$ & $1.8 \pm 2.6$ & $3.5 \pm 2.5$ & $2.7 \pm 2.9$ & $5.2 \pm 4.0$ & $1.0 \pm 2.2$ & $1.5 \pm 2.5$ \\
 & CoffeeSetupMug & $0.0 \pm 0.0$ & $0.7 \pm 1.7$ & $0.0 \pm 0.0$ & $0.2 \pm 1.4$ & $0.0 \pm 0.0$ & $0.0 \pm 0.0$ & $0.0 \pm 0.0$ & $0.0 \pm 0.0$ \\
\cline{1-10}
\multirow[c]{3}{*}{Pressing Buttons} & CoffeePressButton & $19.8 \pm 7.0$ & $33.1 \pm 9.3$ & $16.1 \pm 4.3$ & $24.2 \pm 5.4$ & $22.2 \pm 5.4$ & $23.4 \pm 5.4$ & $18.6 \pm 5.8$ & $20.6 \pm 5.4$ \\
 & TurnOnMicrowave & $7.8 \pm 3.4$ & $\mathbf{33.5 \pm 6.1}$ & $9.4 \pm 5.0$ & $\mathbf{32.1 \pm 5.5}$ & $21.0 \pm 5.4$ & $27.6 \pm 5.2$ & $20.0 \pm 6.0$ & $\mathbf{33.9 \pm 7.3}$ \\
 & TurnOffMicrowave & $16.6 \pm 5.0$ & $\mathbf{42.9 \pm 7.3}$ & $18.6 \pm 7.8$ & $\mathbf{38.0 \pm 5.6}$ & $22.6 \pm 5.8$ & $\mathbf{39.0 \pm 5.8}$ & $21.1 \pm 6.1$ & $\mathbf{42.7 \pm 8.9}$ \\
\cline{1-10}
\multirow[c]{3}{*}{Turning Levers} & TurnOnSinkFaucet & $16.2 \pm 4.6$ & $\mathbf{34.4 \pm 6.4}$ & $13.9 \pm 4.1$ & $\mathbf{25.7 \pm 6.7}$ & $20.8 \pm 5.2$ & $32.5 \pm 7.5$ & $18.8 \pm 5.2$ & $\mathbf{34.2 \pm 5.8}$ \\
 & TurnOffSinkFaucet & $28.0 \pm 8.4$ & $\mathbf{62.8 \pm 7.2}$ & $35.2 \pm 6.0$ & $\mathbf{54.4 \pm 5.2}$ & $52.2 \pm 7.4$ & $54.4 \pm 6.0$ & $37.2 \pm 9.6$ & $\mathbf{61.0 \pm 6.2}$ \\
 & TurnSinkSpout & $36.3 \pm 6.9$ & $\mathbf{64.6 \pm 7.0}$ & $41.2 \pm 5.6$ & $50.0 \pm 5.6$ & $56.4 \pm 5.6$ & $60.7 \pm 6.3$ & $45.8 \pm 7.4$ & $54.6 \pm 6.6$ \\
\cline{1-10}
\multirow[c]{2}{*}{Twisting Knobs} & TurnOnStove & $14.6 \pm 5.4$ & $\mathbf{39.3 \pm 6.7}$ & $24.1 \pm 6.3$ & $35.4 \pm 5.8$ & $29.8 \pm 5.0$ & $38.6 \pm 6.6$ & $21.4 \pm 5.8$ & $\mathbf{35.4 \pm 5.8}$ \\
 & TurnOffStove & $7.9 \pm 4.1$ & $\mathbf{20.4 \pm 5.2}$ & $12.2 \pm 4.6$ & $19.7 \pm 5.9$ & $15.1 \pm 4.9$ & $17.6 \pm 6.0$ & $9.8 \pm 4.2$ & $16.4 \pm 4.0$ \\
\cline{1-10}
\multirow[c]{2}{*}{Open/Close Drawers} & OpenDrawer & $1.3 \pm 1.9$ & $\mathbf{11.7 \pm 5.1}$ & $3.6 \pm 2.4$ & $8.4 \pm 4.8$ & $3.4 \pm 2.2$ & $\mathbf{11.1 \pm 5.3}$ & $3.8 \pm 3.0$ & $3.3 \pm 3.1$ \\
 & CloseDrawer & $33.9 \pm 6.5$ & $\mathbf{71.5 \pm 6.3}$ & $38.6 \pm 8.6$ & $\mathbf{62.1 \pm 8.3}$ & $23.8 \pm 5.4$ & $\mathbf{57.2 \pm 7.2}$ & $27.6 \pm 6.4$ & $\mathbf{53.2 \pm 6.0}$ \\
\cline{1-10}
\multirow[c]{4}{*}{Open/Close Doors} & OpenSingleDoor & $0.3 \pm 2.1$ & $\mathbf{14.4 \pm 4.9}$ & $3.2 \pm 3.2$ & $\mathbf{17.5 \pm 6.1}$ & $3.7 \pm 3.1$ & $11.9 \pm 5.3$ & $0.2 \pm 1.4$ & $\mathbf{11.6 \pm 4.8}$ \\
 & CloseSingleDoor & $24.8 \pm 5.6$ & $\mathbf{58.1 \pm 7.5}$ & $19.6 \pm 6.4$ & $\mathbf{57.0 \pm 7.8}$ & $13.6 \pm 6.0$ & $\mathbf{56.7 \pm 5.9}$ & $6.3 \pm 5.3$ & $\mathbf{40.5 \pm 6.5}$ \\
 & OpenDoubleDoor & $0.0 \pm 0.0$ & $\mathbf{17.4 \pm 5.4}$ & $0.0 \pm 0.0$ & $\mathbf{13.7 \pm 5.5}$ & $0.7 \pm 1.7$ & $\mathbf{9.8 \pm 4.6}$ & $0.0 \pm 0.0$ & $\mathbf{10.5 \pm 4.3}$ \\
 & CloseDoubleDoor & $1.5 \pm 2.5$ & $\mathbf{33.0 \pm 5.8}$ & $18.0 \pm 5.2$ & $\mathbf{34.0 \pm 7.2}$ & $15.6 \pm 6.0$ & $\mathbf{34.2 \pm 7.0}$ & $3.4 \pm 3.0$ & $\mathbf{26.6 \pm 5.4}$ \\
\cline{1-10}
\multicolumn{2}{c|}{\textbf{Overall}} & $13.2 \pm 1.2$ & $\mathbf{33.9 \pm 1.5}$ & $16.0 \pm 1.2$ & $\mathbf{29.8 \pm 1.4}$ & $19.0 \pm 1.2$ & $\mathbf{30.1 \pm 1.4}$ & $14.8 \pm 1.3$ & $\mathbf{27.9 \pm 1.4}$ \\
\cline{1-10}
\bottomrule
\end{tabular}
}

\vspace{0.5cm} %

\resizebox{\linewidth}{!}{%
\renewcommand{\arraystretch}{1.3}
\setlength{\aboverulesep}{0pt}
\setlength{\belowrulesep}{0pt}
\begin{tabular}{l l | >{\centering\arraybackslash}p{2.5cm} >{\centering\arraybackslash}p{2.5cm} | >{\centering\arraybackslash}p{2.5cm} >{\centering\arraybackslash}p{2.5cm} | >{\centering\arraybackslash}p{2.5cm} >{\centering\arraybackslash}p{2.5cm} | >{\centering\arraybackslash}p{2.5cm} | >{\centering\arraybackslash}p{2.5cm}}
\toprule
 &  & PointTransformer & + FF & PointPatch+RGB & + FF & PointPatch+RGB (pretrained) & + FF & PointMap & RGB \\
Category & Task &  &  &  &  &  &  &  &  \\
\midrule
\multirow[c]{2}{*}{Insertion} & CoffeeServeMug & $1.4 \pm 2.2$ & $3.2 \pm 2.8$ & $13.6 \pm 5.2$ & $14.7 \pm 6.1$ & $43.7 \pm 7.5$ & $49.1 \pm 7.7$ & $38.8 \pm 8.4$ & $3.8 \pm 3.0$ \\
 & CoffeeSetupMug & $0.0 \pm 0.0$ & $1.5 \pm 2.5$ & $0.2 \pm 1.4$ & $1.8 \pm 2.6$ & $10.2 \pm 3.8$ & $9.2 \pm 4.0$ & $5.5 \pm 2.9$ & $0.3 \pm 2.1$ \\
\cline{1-10}
\multirow[c]{3}{*}{Pressing Buttons} & CoffeePressButton & $23.4 \pm 5.4$ & $25.8 \pm 5.8$ & $40.8 \pm 8.4$ & $40.5 \pm 8.3$ & $81.1 \pm 7.3$ & $84.6 \pm 7.0$ & $43.2 \pm 8.4$ & $37.1 \pm 6.1$ \\
 & TurnOnMicrowave & $18.8 \pm 7.2$ & $\mathbf{33.5 \pm 5.7}$ & $26.0 \pm 8.4$ & $33.6 \pm 10.4$ & $43.3 \pm 7.5$ & $52.2 \pm 11.0$ & $36.2 \pm 7.8$ & $20.6 \pm 5.8$ \\
 & TurnOffMicrowave & $24.2 \pm 5.8$ & $\mathbf{37.8 \pm 6.6}$ & $20.4 \pm 5.2$ & $37.9 \pm 14.7$ & $51.0 \pm 10.2$ & $52.0 \pm 11.2$ & $46.5 \pm 6.9$ & $10.9 \pm 4.3$ \\
\cline{1-10}
\multirow[c]{3}{*}{Turning Levers} & TurnOnSinkFaucet & $26.4 \pm 5.6$ & $\mathbf{44.2 \pm 7.4}$ & $24.2 \pm 7.0$ & $26.5 \pm 6.7$ & $54.3 \pm 5.7$ & $60.0 \pm 7.2$ & $34.6 \pm 5.4$ & $16.5 \pm 6.3$ \\
 & TurnOffSinkFaucet & $55.9 \pm 6.9$ & $59.0 \pm 9.4$ & $29.4 \pm 5.4$ & $37.2 \pm 7.2$ & $44.6 \pm 6.6$ & $55.9 \pm 8.1$ & $46.2 \pm 9.0$ & $22.1 \pm 5.1$ \\
 & TurnSinkSpout & $51.6 \pm 7.2$ & $62.9 \pm 7.7$ & $29.3 \pm 9.5$ & $34.1 \pm 5.9$ & $5.0 \pm 3.0$ & $7.4 \pm 4.2$ & $43.1 \pm 8.5$ & $24.4 \pm 7.2$ \\
\cline{1-10}
\multirow[c]{2}{*}{Twisting Knobs} & TurnOnStove & $33.6 \pm 6.4$ & $37.4 \pm 5.4$ & $13.6 \pm 5.6$ & $13.6 \pm 4.8$ & $19.5 \pm 5.9$ & $22.4 \pm 5.2$ & $18.9 \pm 6.7$ & $10.1 \pm 3.5$ \\
 & TurnOffStove & $16.6 \pm 4.6$ & $20.7 \pm 4.9$ & $6.6 \pm 3.0$ & $7.7 \pm 3.1$ & $10.3 \pm 4.9$ & $13.3 \pm 4.7$ & $11.7 \pm 5.5$ & $4.9 \pm 3.1$ \\
\cline{1-10}
\multirow[c]{2}{*}{Open/Close Drawers} & OpenDrawer & $5.8 \pm 3.4$ & $\mathbf{15.6 \pm 6.0}$ & $7.2 \pm 4.0$ & $14.6 \pm 4.6$ & $28.1 \pm 6.7$ & $26.8 \pm 6.8$ & $40.0 \pm 7.2$ & $6.5 \pm 3.1$ \\
 & CloseDrawer & $57.6 \pm 8.8$ & $57.0 \pm 9.8$ & $49.8 \pm 7.4$ & $54.5 \pm 10.3$ & $76.6 \pm 5.0$ & $80.4 \pm 5.2$ & $95.1 \pm 2.7$ & $41.0 \pm 7.4$ \\
\cline{1-10}
\multirow[c]{4}{*}{Open/Close Doors} & OpenSingleDoor & $14.4 \pm 4.0$ & $13.3 \pm 5.5$ & $29.2 \pm 6.0$ & $27.4 \pm 8.6$ & $45.2 \pm 7.2$ & $49.8 \pm 5.8$ & $49.0 \pm 7.4$ & $24.7 \pm 5.9$ \\
 & CloseSingleDoor & $30.2 \pm 8.2$ & $33.3 \pm 6.3$ & $42.4 \pm 6.8$ & $52.0 \pm 6.8$ & $74.2 \pm 5.4$ & $73.2 \pm 7.2$ & $67.7 \pm 5.5$ & $40.0 \pm 8.4$ \\
 & OpenDoubleDoor & $1.9 \pm 2.9$ & $6.6 \pm 3.8$ & $1.3 \pm 1.9$ & $4.9 \pm 3.1$ & $61.9 \pm 10.3$ & $74.0 \pm 14.8$ & $21.8 \pm 7.0$ & $0.7 \pm 1.7$ \\
 & CloseDoubleDoor & $5.7 \pm 4.3$ & $\mathbf{16.7 \pm 4.5}$ & $9.3 \pm 5.1$ & $15.9 \pm 5.3$ & $60.8 \pm 7.2$ & $72.5 \pm 9.7$ & $34.0 \pm 7.6$ & $1.2 \pm 2.0$ \\
\cline{1-10}
\multicolumn{2}{c|}{\textbf{Overall}} & $23.0 \pm 1.4$ & $\mathbf{29.3 \pm 1.5}$ & $21.5 \pm 1.5$ & $\mathbf{26.1 \pm 1.8}$ & $44.4 \pm 1.7$ & $\mathbf{48.9 \pm 1.9}$ & $39.5 \pm 1.7$ & $16.6 \pm 1.3$ \\
\cline{1-10}
\bottomrule
\end{tabular}
}

\end{table*}

\begin{table*}[htb]
\caption{
Average success rates on Maniskill tasks across $5$ seeds.
Fourier features improve the performance of point-cloud based architectures, likely because they enable better differentiation of fine-grained details.
For Maniskill, PointMaps are competitive with approaches enhanced with Fourier features, presumably due to larger training datasets.
\textbf{Bold} denotes statistical significance.
}
\begin{center}

\resizebox{\linewidth}{!}{%
\renewcommand{\arraystretch}{1.3}
\setlength{\aboverulesep}{0pt}
\setlength{\belowrulesep}{0pt}
\begin{tabular}{l l | >{\centering\arraybackslash}p{2.5cm} >{\centering\arraybackslash}p{2.5cm} | >{\centering\arraybackslash}p{2.5cm} >{\centering\arraybackslash}p{2.5cm} | >{\centering\arraybackslash}p{2.5cm} >{\centering\arraybackslash}p{2.5cm} | >{\centering\arraybackslash}p{2.5cm} >{\centering\arraybackslash}p{2.5cm}}
\toprule
 &  & PointPatch & + FF & PointPatch-attn & + FF & PCM & + FF & DP3 & + FF \\
Category & Task &  &  &  &  &  &  &  &  \\
\midrule
\multirow[c]{4}{*}{Table-Top 2 Finger Gripper} & PullCube-v1 & $51.4 \pm 4.6$ & $57.1 \pm 4.3$ & $46.8 \pm 6.0$ & $52.9 \pm 5.3$ & $77.3 \pm 3.9$ & $78.8 \pm 3.8$ & $80.0 \pm 4.4$ & $78.3 \pm 3.9$ \\
 & PushCube-v1 & $69.3 \pm 5.3$ & $75.7 \pm 5.7$ & $69.3 \pm 4.5$ & $74.5 \pm 5.1$ & $78.4 \pm 3.8$ & $81.4 \pm 5.4$ & $79.3 \pm 3.1$ & $81.8 \pm 3.4$ \\
 & PokeCube-v1 & $55.4 \pm 5.4$ & $63.6 \pm 4.2$ & $56.4 \pm 4.2$ & $63.5 \pm 5.9$ & $71.0 \pm 3.6$ & $69.2 \pm 6.6$ & $68.2 \pm 4.0$ & $66.9 \pm 3.7$ \\
 & RollBall-v1 & $18.2 \pm 5.0$ & $26.8 \pm 4.2$ & $23.6 \pm 3.6$ & $25.8 \pm 4.0$ & $27.1 \pm 4.1$ & $30.0 \pm 4.6$ & $23.3 \pm 3.3$ & $27.7 \pm 4.3$ \\
\cline{1-10}
\multicolumn{2}{c|}{\textbf{Overall}} & $48.6 \pm 2.5$ & $\mathbf{55.8 \pm 2.3}$ & $49.0 \pm 2.3$ & $\mathbf{54.2 \pm 2.5}$ & $63.5 \pm 1.9$ & $64.9 \pm 2.6$ & $62.7 \pm 1.8$ & $63.7 \pm 1.9$ \\
\cline{1-10}
\bottomrule
\end{tabular}
}

\vspace{0.5cm} %

\resizebox{\linewidth}{!}{%
\renewcommand{\arraystretch}{1.3}
\setlength{\aboverulesep}{0pt}
\setlength{\belowrulesep}{0pt}
\begin{tabular}{l l | >{\centering\arraybackslash}p{2.5cm} >{\centering\arraybackslash}p{2.5cm} | >{\centering\arraybackslash}p{2.5cm} >{\centering\arraybackslash}p{2.5cm} | >{\centering\arraybackslash}p{2.5cm} >{\centering\arraybackslash}p{2.5cm} | >{\centering\arraybackslash}p{2.5cm} | >{\centering\arraybackslash}p{2.5cm}}
\toprule
 &  & PointTransformer & + FF & PointPatch+RGB & + FF & PointPatch+RGB (pretrained) & + FF & PointMap & RGB \\
Category & Task &  &  &  &  &  &  &  &  \\
\midrule
\multirow[c]{4}{*}{Table-Top 2 Finger Gripper} & PullCube-v1 & $57.5 \pm 8.5$ & $63.9 \pm 5.3$ & $75.4 \pm 4.8$ & $74.5 \pm 3.9$ & $82.7 \pm 3.9$ & $80.6 \pm 3.6$ & $71.7 \pm 4.3$ & $74.3 \pm 6.9$ \\
 & PushCube-v1 & $74.5 \pm 6.7$ & $73.9 \pm 4.1$ & $82.7 \pm 3.1$ & $81.2 \pm 3.0$ & $91.2 \pm 3.0$ & $88.6 \pm 2.6$ & $81.2 \pm 4.2$ & $79.4 \pm 4.0$ \\
 & PokeCube-v1 & $57.8 \pm 5.0$ & $58.1 \pm 6.3$ & $69.9 \pm 3.7$ & $75.3 \pm 4.3$ & $76.5 \pm 3.9$ & $74.8 \pm 3.8$ & $69.8 \pm 4.4$ & $72.9 \pm 4.9$ \\
 & RollBall-v1 & $14.1 \pm 6.9$ & $14.4 \pm 4.4$ & $27.9 \pm 4.3$ & $27.0 \pm 3.8$ & $32.0 \pm 5.4$ & $31.7 \pm 4.3$ & $31.5 \pm 4.1$ & $31.3 \pm 5.3$ \\
\cline{1-10}
\multicolumn{2}{c|}{\textbf{Overall}} & $51.0 \pm 3.3$ & $52.6 \pm 2.5$ & $64.0 \pm 2.0$ & $64.5 \pm 1.8$ & $70.6 \pm 2.0$ & $69.0 \pm 1.7$ & $63.6 \pm 2.1$ & $64.5 \pm 2.6$ \\
\cline{1-10}
\bottomrule
\end{tabular}
}

\end{center}
\label{table:maniskill_results}
\end{table*}

\subsection{Simulation Results}
\label{app:sim_results}

Tables \ref{table:robocasa_results} and \ref{table:maniskill_results} show per-task success rates for RoboCasa and Maniskill real tasks, respectively.
As in \citet{pointmappolicy}, we test policies at checkpoints after $60\%$, $80\%$, and $100\%$ of training epochs ($30$, $40$, and $50$ epochs for RoboCasa, and $90$, $120$, and $150$ epochs for ManiSkill).
We perform $50$ rollouts on RoboCasa tasks and $100$ rollouts on ManiSkill tasks, compute the average success rate and select the best-performing checkpoint for each seed.
The final success rate is the mean over $5$ random seeds.

To compute confidence bounds, we adapt the bootstrapping method from \citet{agarwal2021deep} and report the interquartile mean (ICM) and $95\%$ confidence bounds.
To simulate one experiment, we sample $5$ seeds with replacement.
For each checkpoint of each seed, we sample episodes with replacement ($50$ for RoboCasa and $100$ for ManiSkill), then select the best checkpoints and recompute the mean success.
After repeating this for $50,000$ simulated experiments, the interquartile mean is the average success after discarding the top and bottom quartiles, while the $95\%$ confidence bounds are the values that $95\%$ of experiments fall between.

\clearpage

\begin{figure}[tb]
    \centering
    \begin{minipage}[b]{0.1\linewidth}
        \centering
        \textbf{Time $\longrightarrow$}
    \end{minipage}\hfill
    \begin{minipage}[b]{0.15\linewidth}
        \centering
        \textbf{$t=1$}
    \end{minipage}\hfill
    \begin{minipage}[b]{0.15\linewidth}
        \centering
        \textbf{$t=2$}
    \end{minipage}\hfill
    \begin{minipage}[b]{0.15\linewidth}
        \centering
        \textbf{$t=3$}
    \end{minipage}\hfill
    \begin{minipage}[b]{0.15\linewidth}
        \centering
        \textbf{$t=4$}
    \end{minipage}\hfill
    \begin{minipage}[b]{0.15\linewidth}
        \centering
        \textbf{$t=5$}
    \end{minipage}
    
    \subfloat[\text{RoboCasa: \textit{CoffeePressButton}}\label{subfig: coffee}]{
        \centering
        \begin{minipage}{\linewidth}
            \begin{minipage}[b]{0.1\linewidth}
                \textbf{With FF} \\ \\
            \end{minipage}
            \begin{minipage}[b]{0.17\linewidth}
                \centering
                \includegraphics[width=\linewidth]{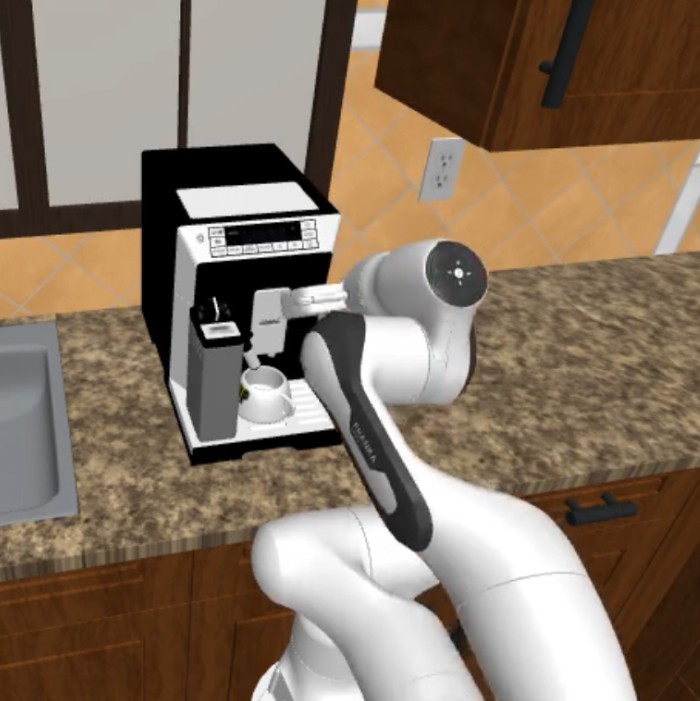}
            \end{minipage}\hfill
            \begin{minipage}[b]{0.17\linewidth}
                \centering
                \includegraphics[width=\linewidth]{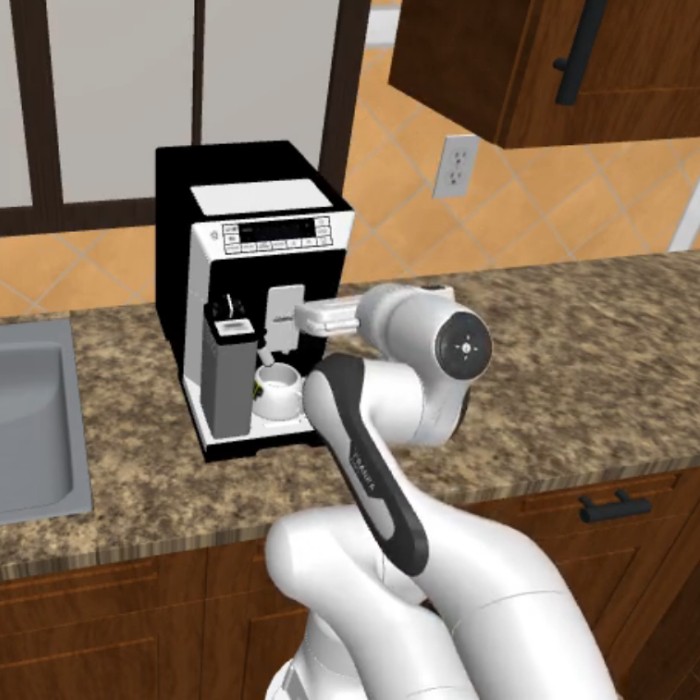}
            \end{minipage}\hfill
            \begin{minipage}[b]{0.17\linewidth}
                \centering
                \includegraphics[width=\linewidth]{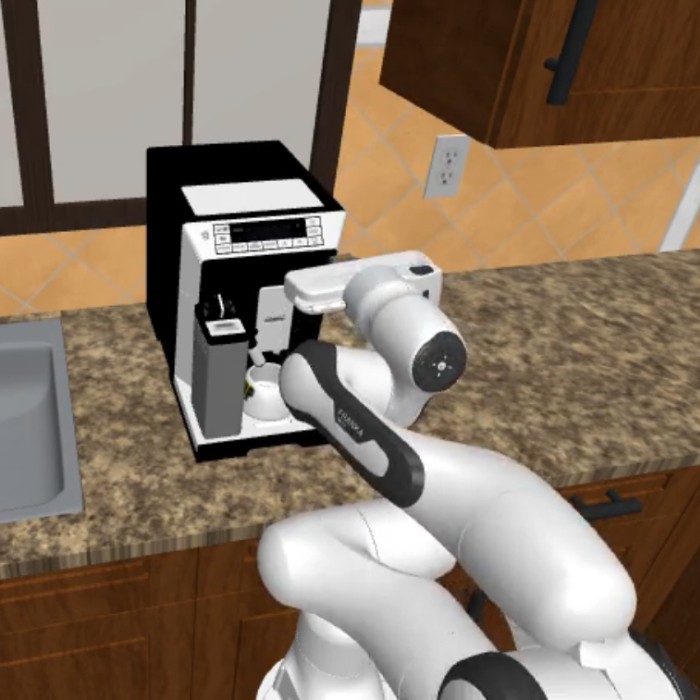}
            \end{minipage}\hfill
            \begin{minipage}[b]{0.17\linewidth}
                \centering
                \includegraphics[width=\linewidth]{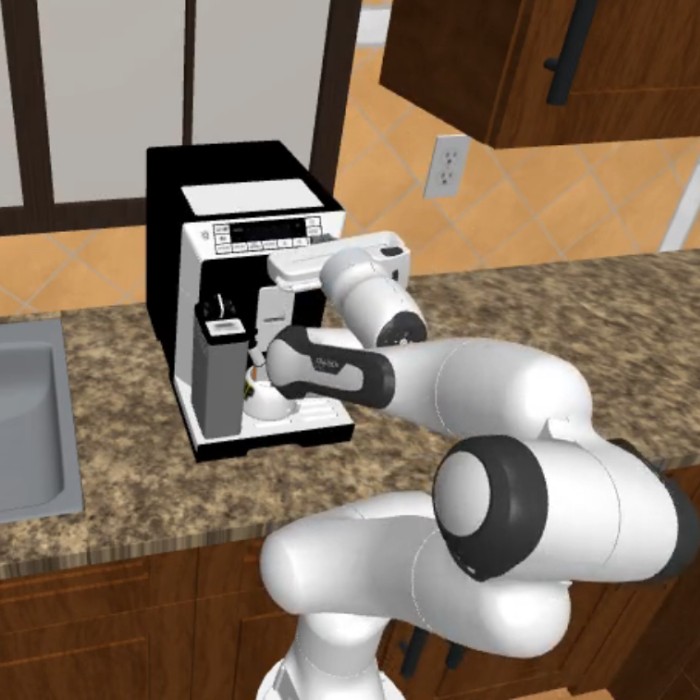}
            \end{minipage}\hfill
            \begin{minipage}[b]{0.17\linewidth}
                \centering
                \includegraphics[width=\linewidth]{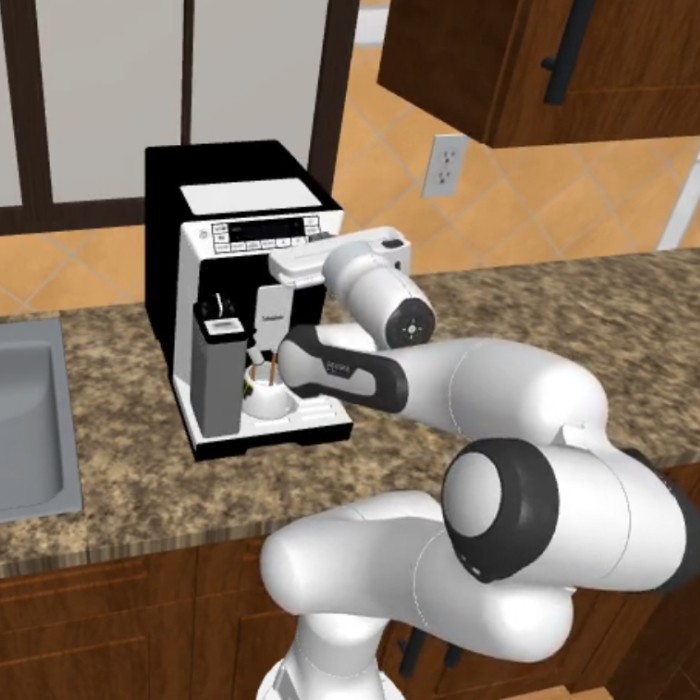}
            \end{minipage}
    
            \vspace{2pt}
            \begin{minipage}[b]{0.1\linewidth}
                \textbf{W/o FF} \\ \\
            \end{minipage}
            \begin{minipage}[b]{0.17\linewidth}
                \centering
                \includegraphics[width=\linewidth]{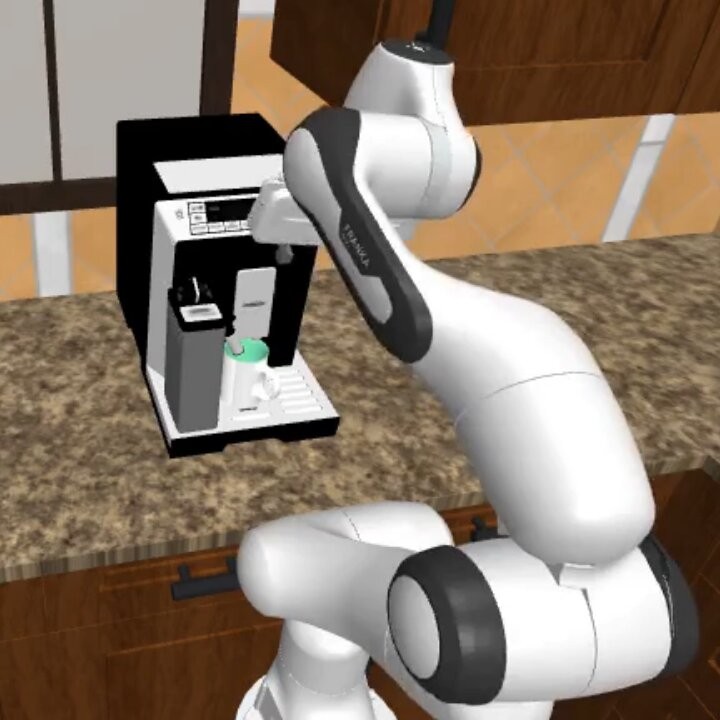}
            \end{minipage}\hfill
            \begin{minipage}[b]{0.17\linewidth}
                \centering
                \includegraphics[width=\linewidth]{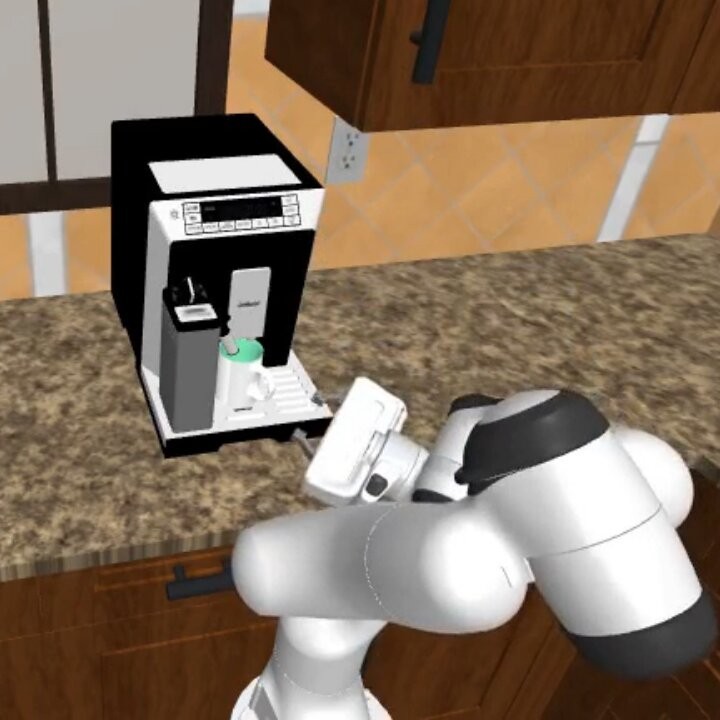}
            \end{minipage}\hfill
            \begin{minipage}[b]{0.17\linewidth}
                \centering
                \includegraphics[width=\linewidth]{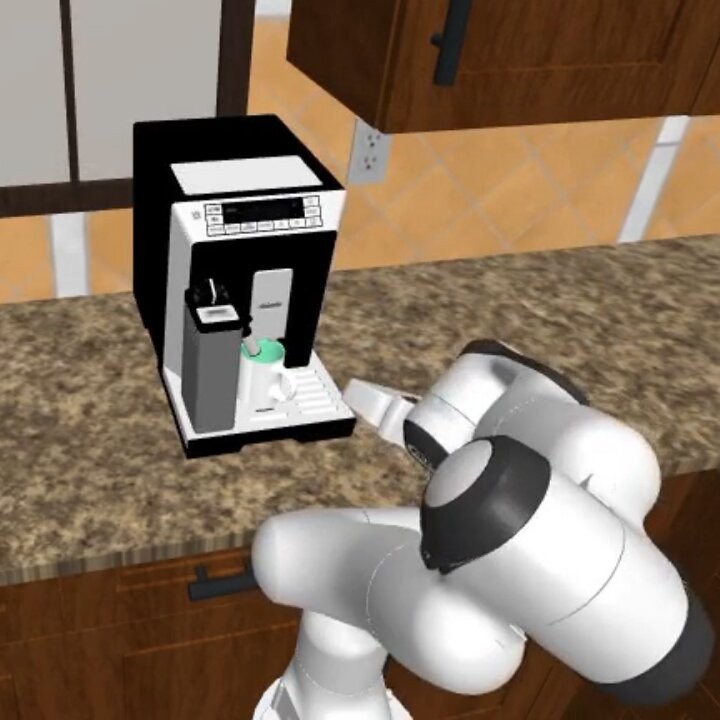}
            \end{minipage}\hfill
            \begin{minipage}[b]{0.17\linewidth}
                \centering
                \includegraphics[width=\linewidth]{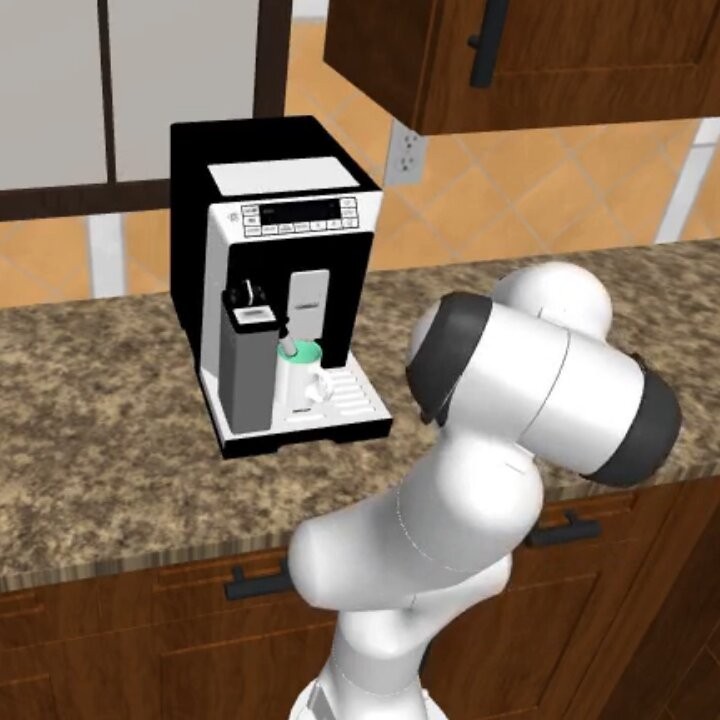}
            \end{minipage}\hfill
            \begin{minipage}[b]{0.17\linewidth}
                \centering
                \includegraphics[width=\linewidth]{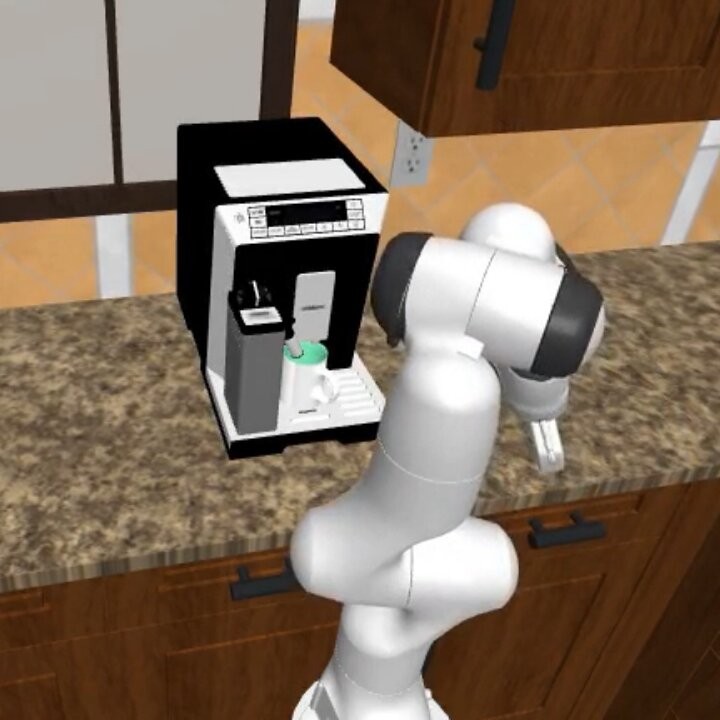}
            \end{minipage}
        \end{minipage}
    }

    \subfloat[\text{RoboCasa: \textit{TurnOnMicrowave}}\label{subfig: microwave}]{
        \centering

        \begin{minipage}{\linewidth}
            \begin{minipage}[b]{0.1\linewidth}
                \textbf{With FF} \\ \\
            \end{minipage}
            \begin{minipage}[b]{0.17\linewidth}
                \centering
                \includegraphics[width=\linewidth]{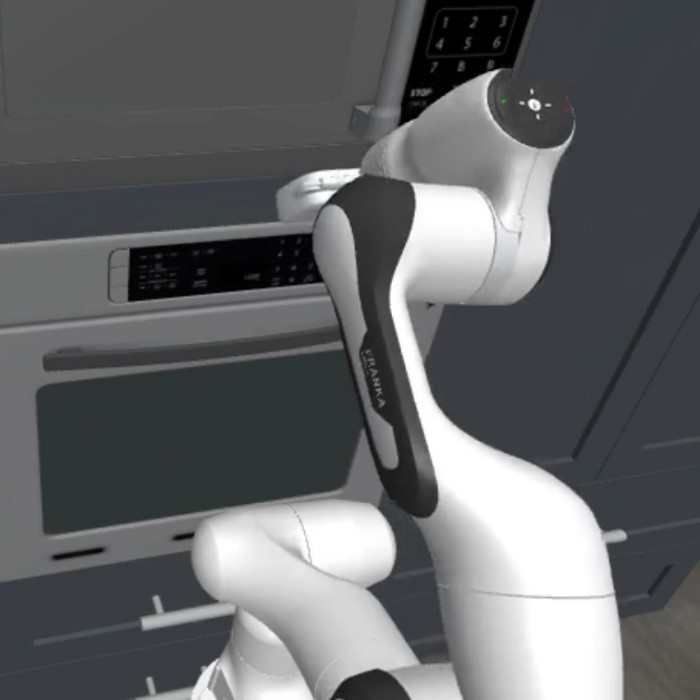}
            \end{minipage}\hfill
            \begin{minipage}[b]{0.17\linewidth}
                \centering
                \includegraphics[width=\linewidth]{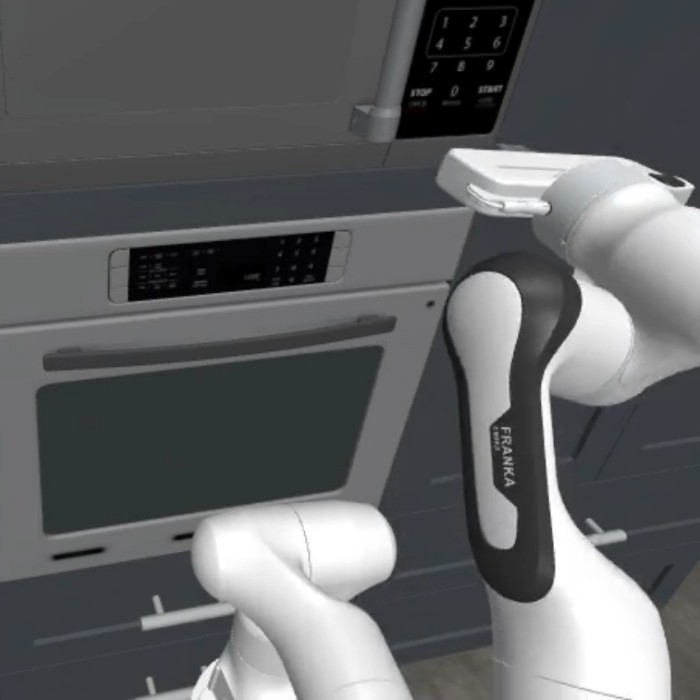}
            \end{minipage}\hfill
            \begin{minipage}[b]{0.17\linewidth}
                \centering
                \includegraphics[width=\linewidth]{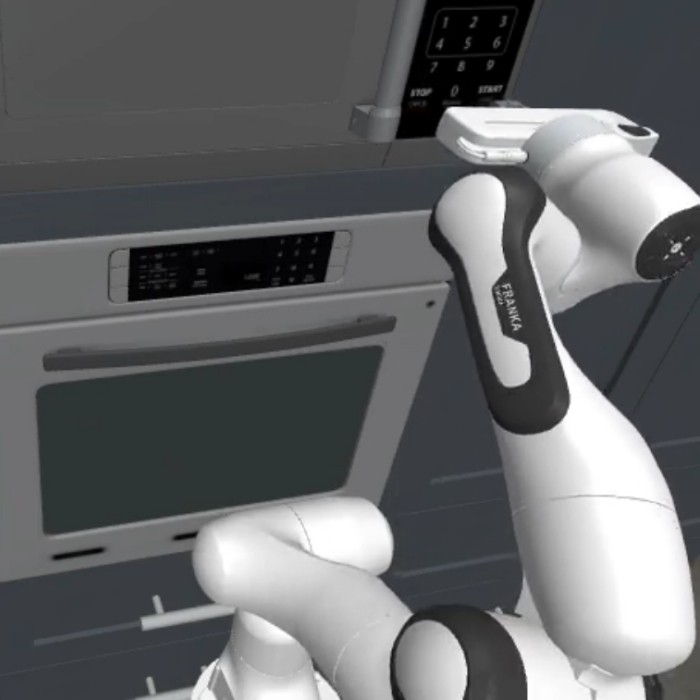}
            \end{minipage}\hfill
            \begin{minipage}[b]{0.17\linewidth}
                \centering
                \includegraphics[width=\linewidth]{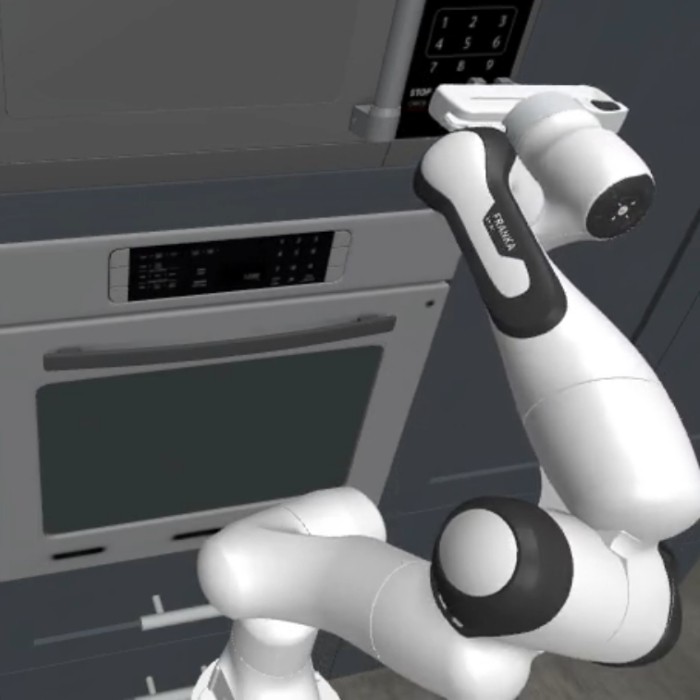}
            \end{minipage}\hfill
            \begin{minipage}[b]{0.17\linewidth}
                \centering
                \includegraphics[width=\linewidth]{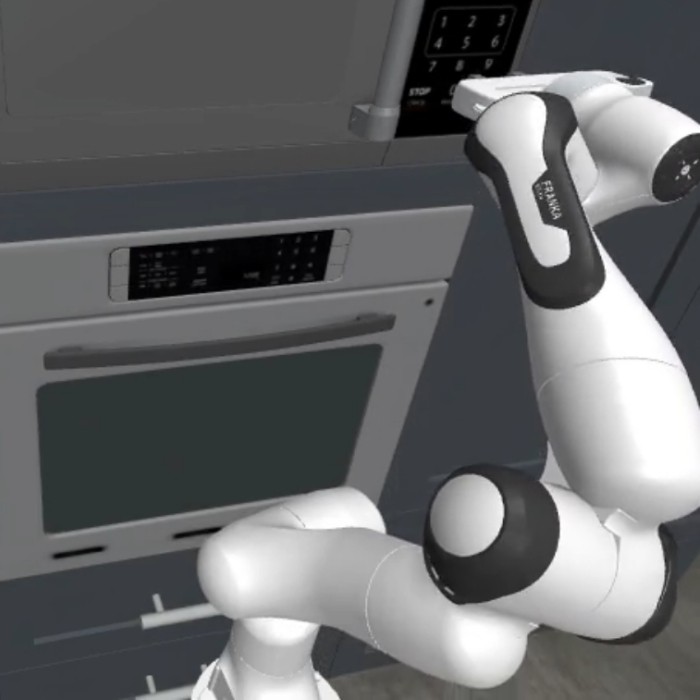}
            \end{minipage}
            
            \vspace{2pt}
            
            \begin{minipage}[b]{0.1\linewidth}
                \textbf{W/o FF} \\ \\
            \end{minipage}
            \begin{minipage}[b]{0.17\linewidth}
                \centering
                \includegraphics[width=\linewidth]{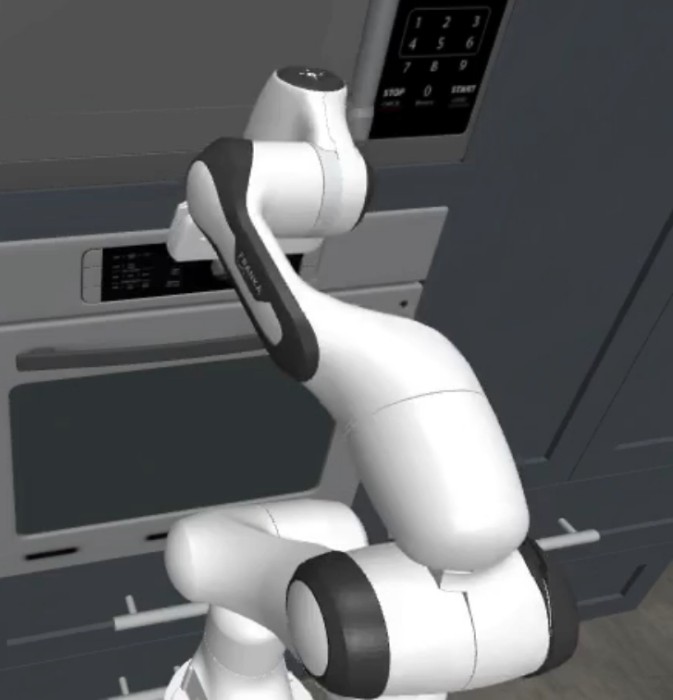}
            \end{minipage}\hfill
            \begin{minipage}[b]{0.17\linewidth}
                \centering
                \includegraphics[width=\linewidth]{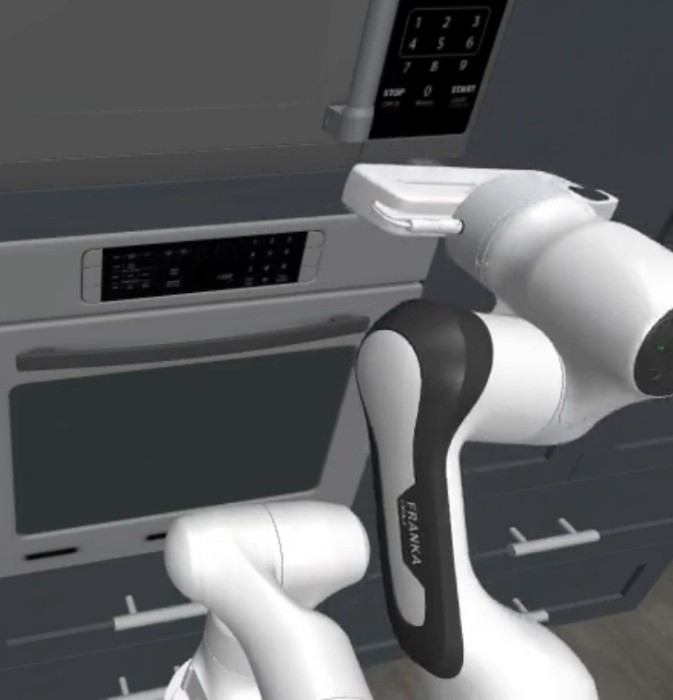}
            \end{minipage}\hfill
            \begin{minipage}[b]{0.17\linewidth}
                \centering
                \includegraphics[width=\linewidth]{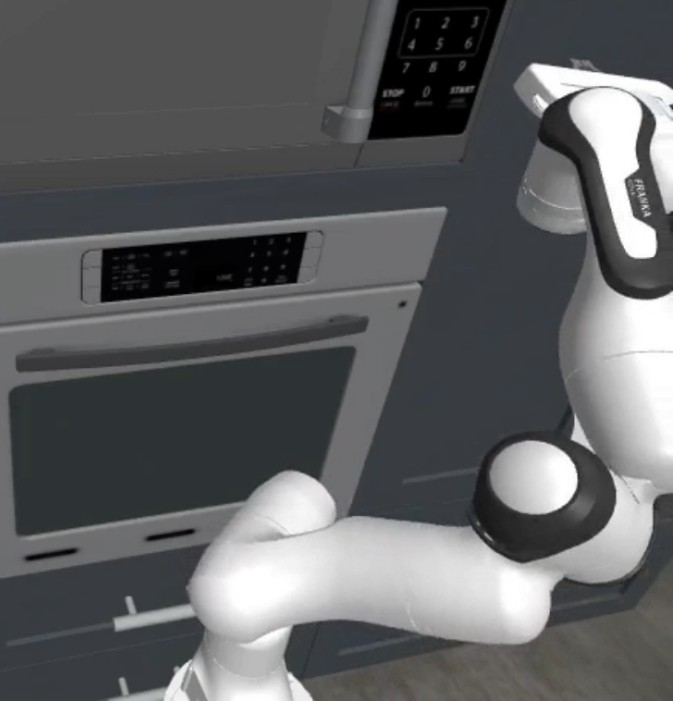}
            \end{minipage}\hfill
            \begin{minipage}[b]{0.17\linewidth}
                \centering
                \includegraphics[width=\linewidth]{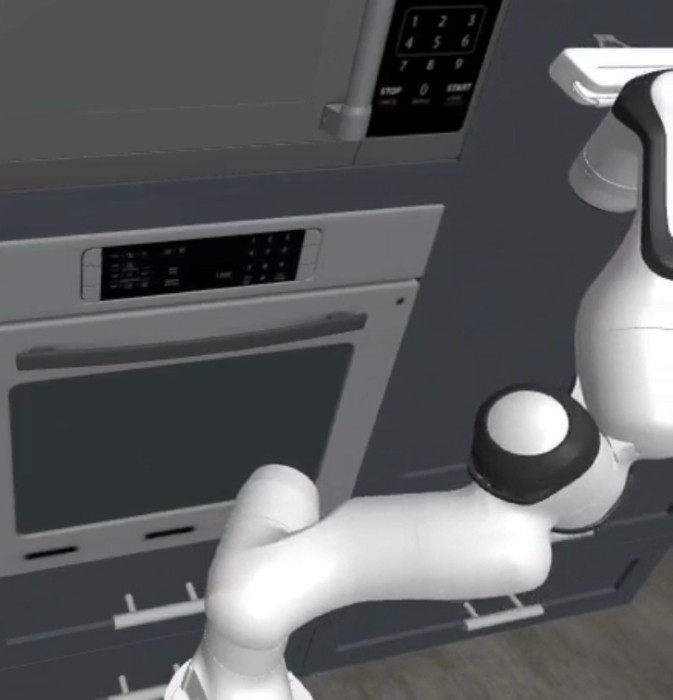}
            \end{minipage}\hfill
            \begin{minipage}[b]{0.17\linewidth}
                \centering
                \includegraphics[width=\linewidth]{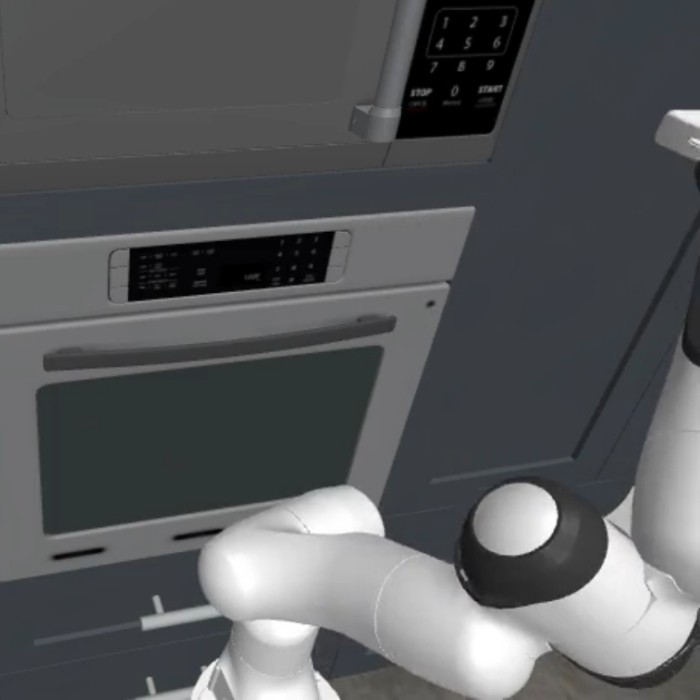}
            \end{minipage}
        \end{minipage}    
    }

    \subfloat[\text{RoboCasa: \textit{TurnSinkFaucet}}\label{subfig: sink}]{

        \begin{minipage}{\linewidth}
            \centering

            \begin{minipage}[b]{0.1\linewidth}
                \textbf{With FF} \\ \\
            \end{minipage}
            \begin{minipage}[b]{0.17\linewidth}
                \centering
                \includegraphics[width=\linewidth]{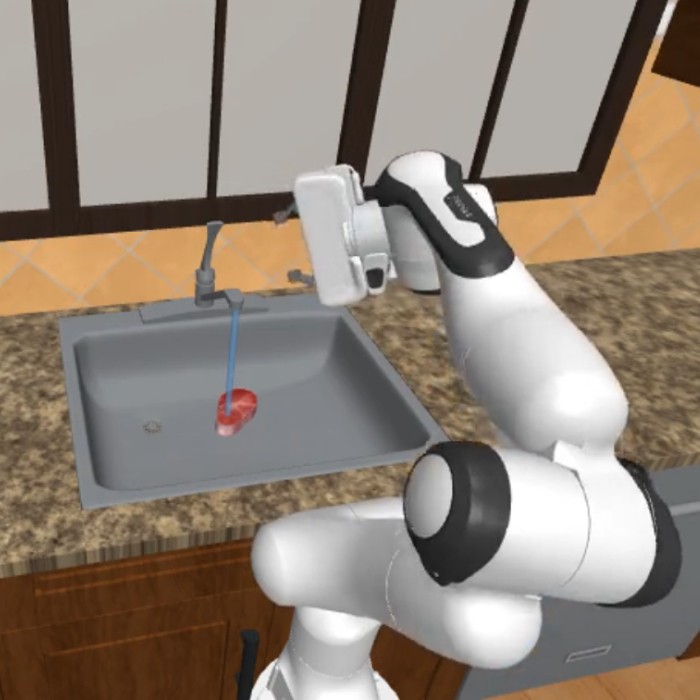}
            \end{minipage}\hfill
            \begin{minipage}[b]{0.17\linewidth}
                \centering
                \includegraphics[width=\linewidth]{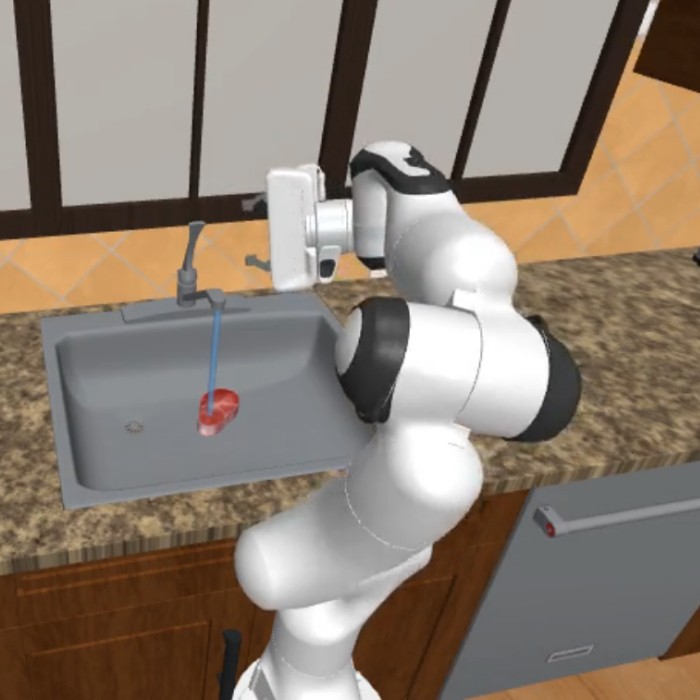}
            \end{minipage}\hfill
            \begin{minipage}[b]{0.17\linewidth}
                \centering                \includegraphics[width=\linewidth]{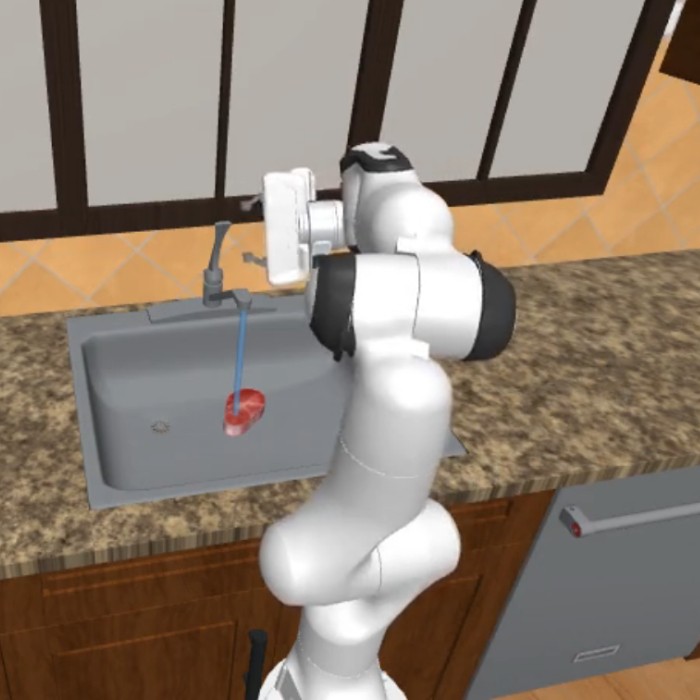}
            \end{minipage}\hfill
            \begin{minipage}[b]{0.17\linewidth}
                \centering
                \includegraphics[width=\linewidth]{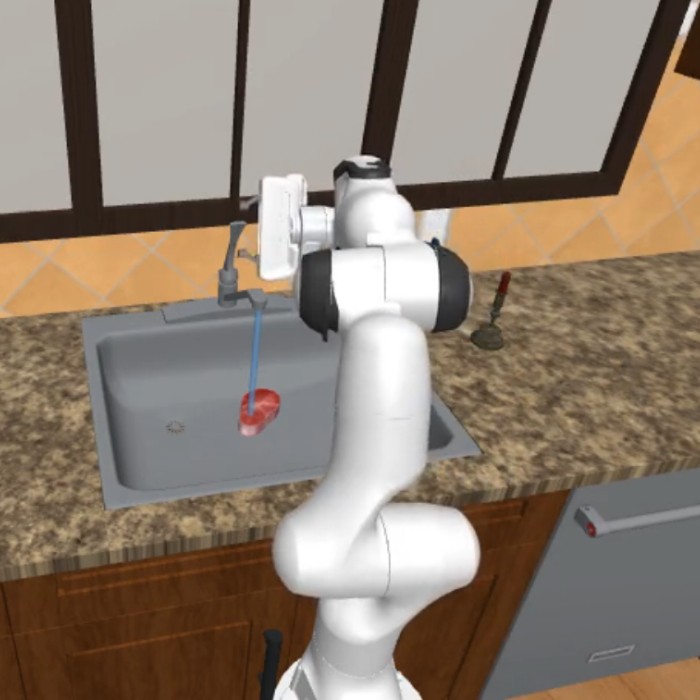}
            \end{minipage}\hfill
            \begin{minipage}[b]{0.17\linewidth}
                \centering
                \includegraphics[width=\linewidth]{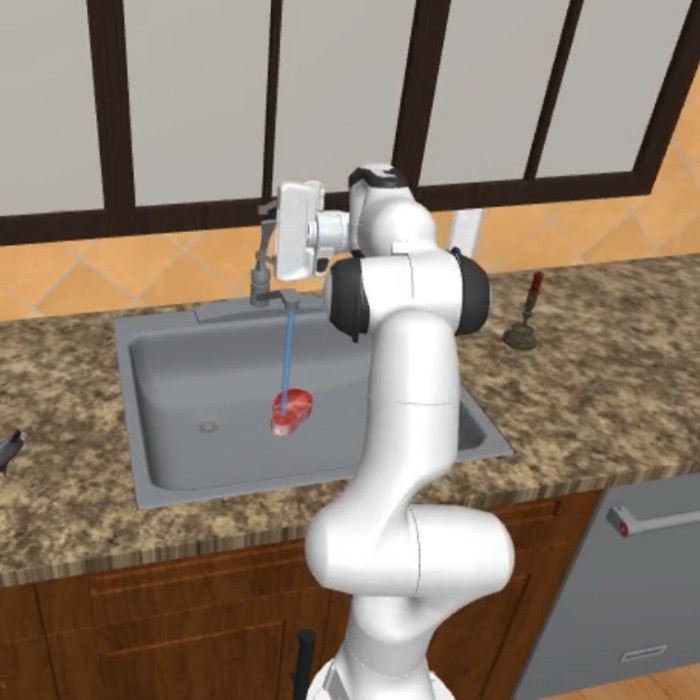}
            \end{minipage}
        
            \vspace{2pt}

            \begin{minipage}[b]{0.1\linewidth}
                \textbf{W/o FF} \\ \\
            \end{minipage}
            \begin{minipage}[b]{0.17\linewidth}
                \centering
                \includegraphics[width=\linewidth]{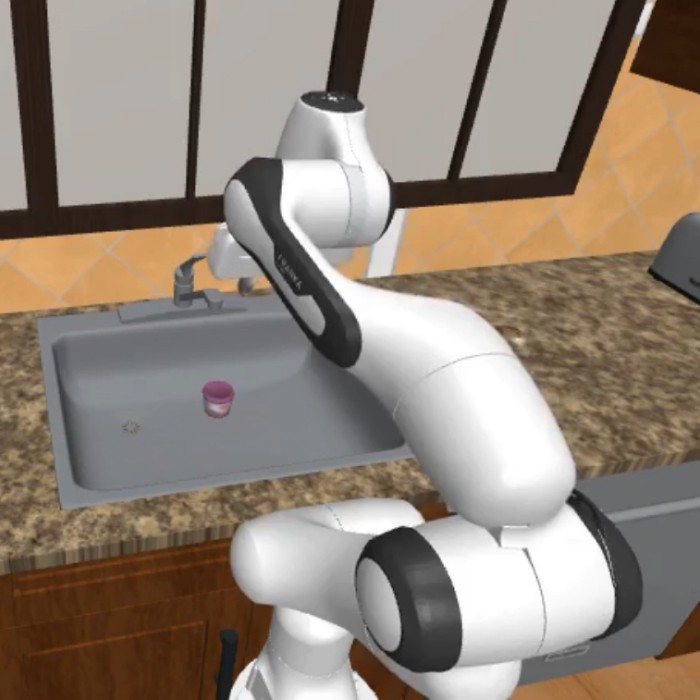}
            \end{minipage}\hfill
            \begin{minipage}[b]{0.17\linewidth}
                \centering
                \includegraphics[width=\linewidth]{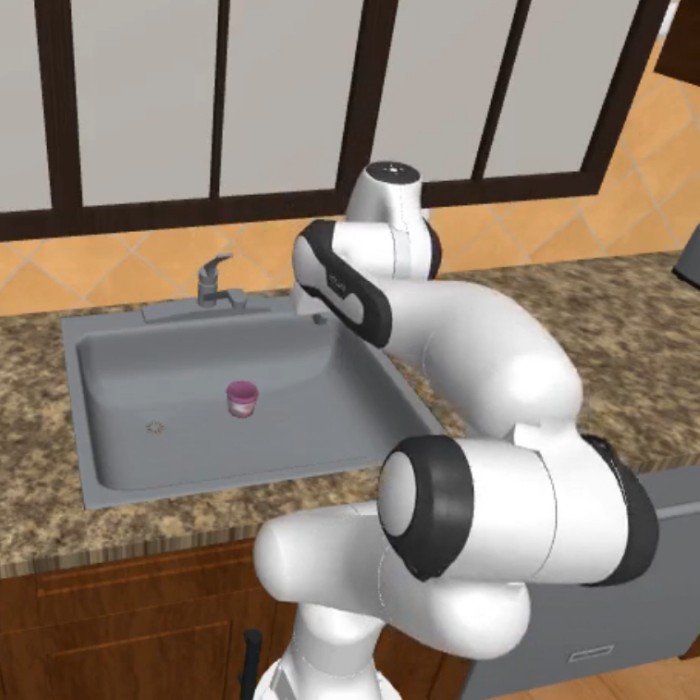}
            \end{minipage}\hfill
            \begin{minipage}[b]{0.17\linewidth}
                \centering
                \includegraphics[width=\linewidth]{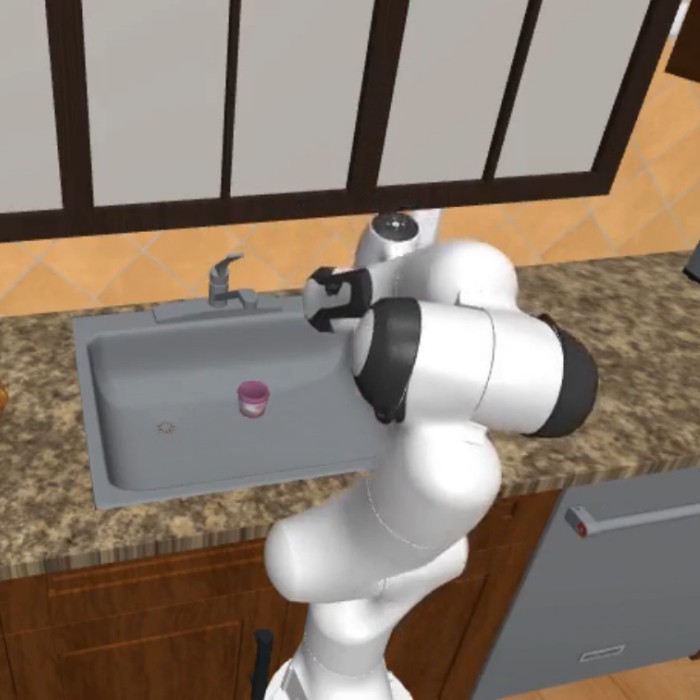}
            \end{minipage}\hfill
            \begin{minipage}[b]{0.17\linewidth}
                \centering
                \includegraphics[width=\linewidth]{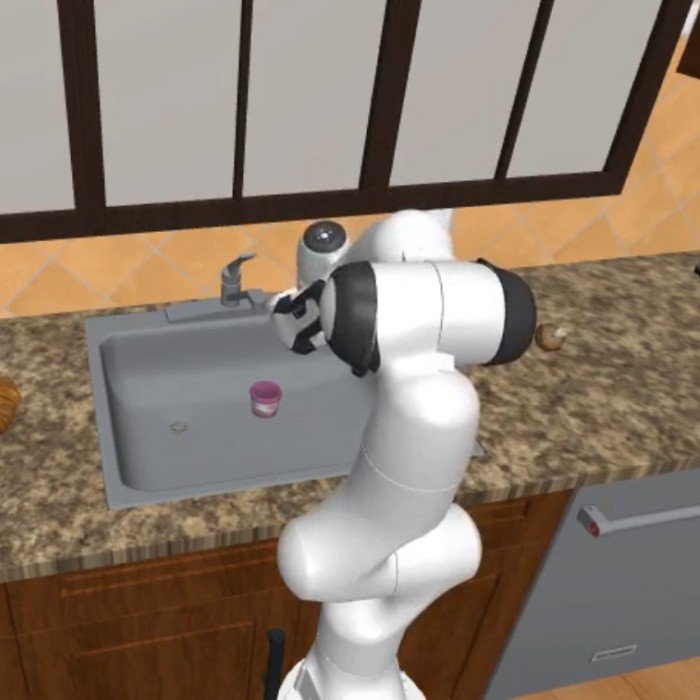}
            \end{minipage}\hfill
            \begin{minipage}[b]{0.17\linewidth}
                \centering
                \includegraphics[width=\linewidth]{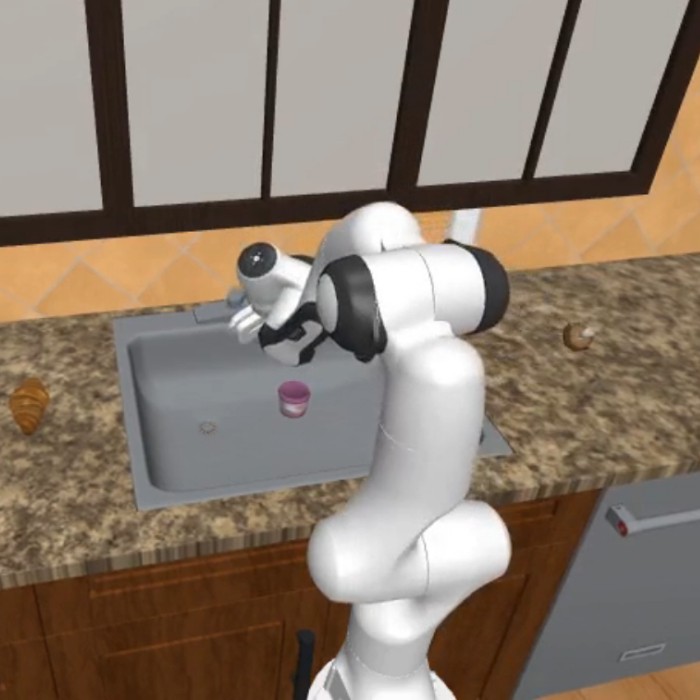}
            \end{minipage}
        \end{minipage}
    }

    \caption{
        Qualitative comparison of PointPatch + FF (upper row) and PointPatch (lower row) policies on three RoboCasa tasks.
        Policies trained without Fourier features have difficulty learning the demonstration data and carrying out complex movements with precision.
        Time proceeds from left to right in each row.
    }
    \label{fig:task_success_failure}
\end{figure}

\subsection{Qualitative Results}
\label{app:qualitative_results}

Figure \ref{fig:task_success_failure} shows representative rollouts from PointPatch+FF policies on selected RoboCasa tasks.
Overall, the agents trained with Fourier features reliably make contact with the target objects (e.g., buttons and lever handles) and completes all three tasks, whereas the agents trained without Fourier features fail to accomplish the tasks.

\clearpage

\subsection{Real World Results}
\label{app:real_results}

Table \ref{tab:results_real_small} shows per-task scores for each real world task.
In order to arrive at more fine-grained results, we report a task-dependent score rather than a simple success rate, where the maximum score for each task is given in the rightmost column of Table \ref{tab:results_real_small}.
Since policies are very sensitive to the initial scene configuration and the experiments are not blind, it is easy to introduce bias in the results.
To minimize this, we employ an alternating testing scheme, where we alternate between rollouts of our method and the baseline.
This allows the human operator to ensure a consistent scene configuration for both methods.

Table \ref{tab:results_real_cup_size} shows additional results collected on the Cup-Stacking task with cups of different sizes.
We note that these are results are distinct from those of Table \ref{tab:results_real_small}, which is why the scores of each method differ between the two tables.
However, the policy with Fourier features is still clearly better in both cases.

\begin{table}[htb!]
    \centering
    \footnotesize
    \caption{
        Average scores across $16$ rollouts on our four real-world experiments.
        Adding Fourier features results in significant improvements across all tasks, with the greatest gain seen in the Drawer task.
    }
    \label{tab:results_real_small}
    \setlength{\tabcolsep}{8pt}
    \renewcommand{\arraystretch}{1.2}
    \begin{tabular}{lcccc}
        \toprule 
        \multirow{2}{*}{\textbf{Task}} & 
        \multirow{2}{*}{\textbf{\makecell[c]{Demos\\Per Task}}} &
        \multicolumn{3}{c}{\textbf{Methods with Scores}} \\
        \cmidrule(lr){3-5}
        & & \makecell[c]{RGB + PointPatch\\(pretrained)} & + FF & Max \\
        \midrule
        Drawer       & 102 & $0.3125$ & $\mathbf{1.625}$  & $4$ \\
        Cup-Stacking & 80  & $0$      & $\mathbf{0.625}$  & $2$ \\
        Arranging    & 100 & $0.3125$ & $\mathbf{1.3125}$ & $4$ \\
        Folding      & 75  & $1.3125$ & $\mathbf{1.6875}$ & $3$ \\
        \bottomrule
    \end{tabular}
\end{table}

\begin{table}[htb!]
    \centering
    \caption{
        Average scores across 10 additional rollouts on each pair of cups used in the Cup-Stacking task.
        The gap between the baseline and the policy with Fourier features widens as cup size decreases.
        For the smallest cups, neither policy is able to solve this challenging task and the advantage disappears.
    }
    \label{tab:results_real_cup_size}
    \begin{tabular}{lccc}
        \toprule
        \multirow{2}{*}{\textbf{Cup Pair}} & 
        \multirow{2}{*}{\textbf{\makecell[c]{Diameter of Smaller/\\Bigger Cup (cm)}}} & 
        \multicolumn{2}{c}{\textbf{Methods with Scores (out of 2.0)}} \\
        \cmidrule(lr){3-4}
        & & \makecell[c]{RGB + PointPatch\\(pretrained)} & + FF \\
        \midrule
        Yellow/Orange & $7.5/8.0$ & $1.1$ & $1.4$ \\
        Blue/Purple   & $6.0/7.0$ & $0.5$ & $1.6$ \\
        Red/Blue      & $5.5/6.0$ & $0.5$ & $0.6$ \\
        Orange/Red    & $5.0/5.5$ & $0.5$ & $0.4$ \\
        \bottomrule
    \end{tabular}
\end{table}

\clearpage

\subsection{Spectral Analysis}
\label{app:spectral_analysis}
To evaluate the spectral sensitivity of Fourier features, we consider the frequency response of different trained and untrained models with and without Fourier features. 
Specifically, we consider a random trajectory from the demonstrations for CoffeePressButton in RoboCasa. 
For each time step, we construct a $k$-nearest neighbor graph from the input point cloud, and then weight the resulting edges using the Zelnik-Manor product~\citep{zelnik2004self}.
We compute the eigenvalue decomposition of the Symmetric Normalized Laplacian, of which the eigenvalues typically lie in $\lambda \in [0, 2]$ \citep{chung1997spectral}. 
We utilize a sparse neighborhood of $k=8$ to consider the local data manifold \citep{von2007tutorial}, ensuring the Laplacian remains sensitive to high-frequency geometric features. 
Given this graph and its Laplacian, we transform the input-output gradients onto the resulting graph Fourier basis and plot the model's response across the spectrum. 
Figure \ref{fig:gft_spectra} visualizes the averaged and binned spectrum. 
We find that Fourier features increase the model's sensitivity to a wide range of frequencies, including high-frequency signals with $\lambda>1.0$ that are otherwise suppressed by the low-pass bias of standard coordinate inputs.

\begin{figure}[ht]
    \centering
    \includegraphics[width=0.85\linewidth]{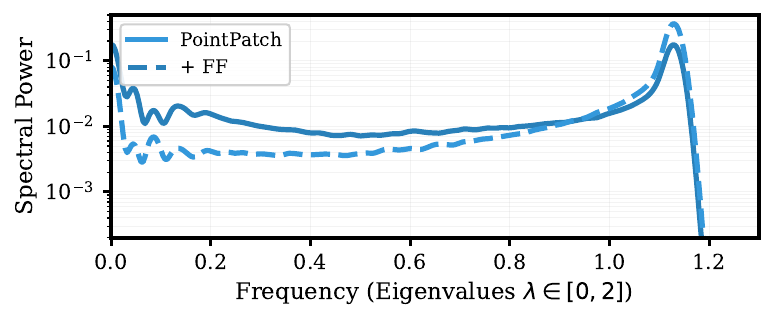}
    \caption{\textbf{Graph Spectral Sensitivity Comparison}.
    We consider a toy problem where we study the point-wise gradient of the sum of model outputs of an untrained PointNet on a point cloud of a sphere. By projecting these gradients onto the basis of a Symmetric Normalized Laplacian constructed with Zelnik-Manor local scaling, we observe that the vanilla architecture (dashed) is inherently biased toward low-frequency geometric components. In contrast, the addition of log-spaced Fourier features (solid line) amplifies the model's sensitivity to high-frequency manifold details ($\lambda > 1.0$), even before any task-specific training occurs.
    }
    \label{app_fig:fourier_spectrum}
\end{figure}

To demonstrate the increase sensitivity to high frequency signals specifically, we consider a simple untrained PointNet consisting of a three-layer MLP followed by global attention pooling and a two-layer output MLP.
The PointNet consumes $1024$ randomly sampled points sampled from a unit ball, and we determine the gradient of the sum of all node's scalar output w.r.t.\ all input coordinates and average them to obtain a sensitivity of the output w.r.t.\ the graph input.
We repeat this setup for $100$ randomly sampled point clouds.
This input-output saliency is then projected onto the graph Fourier basis of the symmetric normalized Laplacian using Zelnik-Manor local scaling \citep{zelnik2004self}.
Through this decomposition of the saliency into the eigenvectors, we split the total gradient into directional derivatives corresponding to different wavelengths.
Due to the use of the Symmetric Normalized Laplacian, the frequency domain is normalized to $(0, 2)$ where the normalized frequency $1$ distinguishes between low and high frequencies. 
Plotting the eigenvalue strength of the saliency for a model with and without Fourier features in Figure~\ref{app_fig:fourier_spectrum}, we can observe a decrease in sensitivity to low frequency changes and an increase to higher frequencies. 
Since the saliency map equates to the linear term of a first-order Taylor approximation of the model network, this confirms that adding Fourier features counteracts the spectral bias of the used MLPs.

\end{document}